%% file: main.tex
\documentclass[10pt,twocolumn,letterpaper]{article}

\usepackage[pagenumbers]{cvpr} 

\usepackage{array}
\usepackage{multirow}
\usepackage{longtable}
\usepackage{siunitx}

\usepackage{pifont}
\newcommand{\cmark}{\ding{51}}%
\newcommand{\xmark}{\ding{55}}%

\usepackage{xspace}
\newcommand{\paper}{Foundry\xspace}
\newcommand{\papershort}{Foundry\xspace}
\newcommand{\methodname}{Foundry\xspace}

\DeclareMathOperator*{\argmax}{arg\,max}

\definecolor{cvprblue}{rgb}{0.21,0.49,0.74}
\usepackage{hyperref}
\hypersetup{pagebackref,breaklinks,colorlinks,allcolors=cvprblue}
\usepackage{cuted}
\usepackage{listings}
\usepackage{xcolor}

\definecolor{codegray}{gray}{0.95}
\definecolor{commentgreen}{RGB}{0,128,0}
\definecolor{keywordblue}{RGB}{0,0,180}
\def\paperID{45259} 
\def\confName{CVPR}
\def\confYear{2026}

\title{Foundry: Distilling 3D Foundation Models for the Edge}

\author{Guillaume Letellier$^1$ \and Siddharth Srivastava$^2$ \and Frederic Jurie$^1$ \and Gaurav Sharma$^3$\\
$^1$ GREYC, Normandy University, Unicaen, ENSICAEN, UMR CNRS 6072, F-14000 Caen, France\\$^2$ IIT Delhi, $^3$ IIT Kanpur\\
{\tt\small \{firstname.lastname\}@unicaen.fr}\\
}

\begin{document}
\maketitle

\begin{abstract}
Foundation models pre-trained with self-supervised learning (SSL) on large-scale datasets have become powerful general-purpose feature extractors. However, their immense size and computational cost make them prohibitive for deployment on edge devices such as robots and AR/VR headsets. Existing compression techniques like standard knowledge distillation create efficient `specialist' models but sacrifice the crucial, downstream-agnostic generality that makes foundation models so valuable.



In this paper, we introduce Foundation Model Distillation (FMD), a new paradigm for compressing large SSL models into compact, efficient, and faithful proxies that retain their general-purpose representational power. We present \methodname, the first implementation of FMD for 3D point clouds. Our approach, \methodname, trains a student to learn a compressed set of SuperTokens that reconstruct the teacher’s token-level representations, capturing a compact basis of its latent space. A single distilled model maintains strong transferability across diverse downstream tasks—classification, part segmentation, and few-shot scenarios—approaching full foundation-model performance while using significantly fewer tokens and FLOPs, making such models more practical for deployment on resource-constrained hardware. Project page: \url{https://guigui14460.github.io/foundry.github.io/}.

\end{abstract}

\section{Introduction}
\label{sec:intro}

The machine learning landscape is increasingly dominated by large-scale foundation models \cite{devlin2019bert,brown2020language,he2022masked}. Pre-trained on vast, unlabeled datasets using self-supervised learning (SSL), these models serve as powerful, general-purpose feature extractors that can be adapted to a wide array of downstream tasks. This paradigm has shown immense success in 3D vision, with models trained on datasets of synthetic objects \cite{wu20153d,yi2017large}, real-world scans \cite{uy2019revisiting,wu2023omniobject3d}, and large-scale scenes \cite{armeni20163d,dai2017scannet,sun2020scalability} becoming the backbone for tasks in robotics, autonomous driving, and AR/VR.

However, the very scale that makes these models powerful also creates a significant deployment bottleneck. With hundreds of millions of parameters and quadratic attention complexity, models like PointTransformer \cite{zhao2021point} cannot be executed on resource-constrained devices. As we show, even a modern GPU can fail to process a moderately sized point cloud of 300k points as demonstrated in \cref{sec:experiments:cost}, let alone the million-point scenes common in real-world applications. This computational barrier prevents the power of 3D foundation models from reaching the devices where they are often needed most.

Existing compression techniques fall short of this goal because they trade generality for efficiency. The dominant approach, knowledge distillation (KD) \cite{hinton2015distilling}, typically trains a student to mimic the teacher's logits on a specific task, creating an efficient `specialist'. While recent works have explored distilling feature embeddings from large vision-language models like CLIP \cite{yang_2024_cvpr}, these methods focus on preserving a specific cross-modal alignment capability via direct feature mimicry. This still results in a student specialized for tasks like zero-shot classification, rather than a truly general-purpose unimodal backbone. A fundamental gap remains for a method that can distill the entire \textit{representational manifold} of a unimodal SSL model.

To bridge this gap, we explore a distillation paradigm we call \emph{Foundation Model Distillation (FMD)}. The goal of FMD is not to create a task-specific specialist, but to forge a compact, portable, and efficient proxy of the original foundation model that retains its core identity as a general-purpose feature extractor. We introduce \emph{\methodname}, the first framework to realize FMD for 3D point cloud Transformers. At the heart of \methodname is the concept of learnable \emph{SuperTokens}. We train a lightweight student wrapper to compress the teacher's dense set of token embeddings into a small, fixed-size set of SuperTokens, and then reconstruct the original embeddings from this compressed representation. This compress-and-reconstruct objective forces the student to learn a highly efficient basis for the teacher's latent space, capturing its salient semantic and geometric features in a compact form.

The result is a standalone student model that acts as a miniature foundation model. It can be cheaply fine-tuned for numerous downstream tasks without ever needing the original teacher again. Our contributions are:
\begin{itemize}
    \item We propose Foundation Model Distillation (FMD), a new paradigm for creating compact, general-purpose proxies of large SSL models by distilling their entire representation space.
    \item We introduce \methodname, the first FMD framework for 3D point clouds. Its core novelty is a \emph{compress-and-reconstruct} objective that forces a student to learn a compact set of SuperTokens as an efficient basis for the teacher's latent space.
    \item We demonstrate that our distillation strategy is superior to both direct feature mimicry and task-specific KD, creating a student model with significantly better transferability and data efficiency.
    \item Our approach yields a portable and efficient 3D foundation model proxy, making large-scale 3D perception practical for resource-constrained GPUs.
\end{itemize}

\section{Related Work}
\label{sec:relatedwork}
We review prior work on data preprocessing for efficiency and methods for compressing large Transformer models.

\subsection{Efficient Data Preprocessing}
Traditionally, to reduce the number of tokens fed to a Transformer, images can be downsampled to a lower resolution, videos can be processed with temporal frame subsampling and spatial resizing, and 3D data can be downsampled using grid sampling or Farthest Point Sampling (FPS)~\cite{eldar1997farthest}. Later, manually designed methods used saliency maps \cite{DBLP:journals/mti/RuiuMG24} for 3D point clouds or quadtree-based spatial splitting and merging in STEP \cite{DBLP:conf/visigrapp/ProustPSH25} to reduce the input before tokenizing. STTM \cite{DBLP:journals/corr/abs-2507-07990} applies this idea directly to tokenized data. Other techniques are more specific to 3D point cloud compression and exploit geometric properties \cite{DBLP:conf/icassp/NguyenQVD21,DBLP:journals/corr/abs-2207-12554,DBLP:journals/corr/abs-2402-07243,DBLP:journals/corr/abs-2503-12382,DBLP:journals/corr/abs-2408-10543}. Finally, hand-designed algorithms based on graph clustering \cite{raguet2019parallel} can reduce input size by creating superpoints \cite{DBLP:conf/iccv/0002RL23}. Our work differs fundamentally from these approaches: rather than pre-sampling points, we learn a compression of the dense feature space itself.

\subsection{Model Compression for Transformers}

\noindent\textbf{Token Merging and Pruning.}
A popular line of work focuses on reducing the number of tokens processed `online' within a Transformer to accelerate inference. Methods like ToMe \cite{DBLP:conf/iclr/BolyaFDZFH23} and PiToMe \cite{DBLP:conf/nips/TranNNNLXSZNN24} progressively merge similar tokens between layers during a single forward pass. Other techniques, such as PatchMerger \cite{DBLP:journals/corr/abs-2202-12015}, learn a codebook to merge tokens at specific blocks within the network. For 3D point clouds, 3DLST \cite{lu20243d} also proposes a learnable supertoken module for a supervised segmentation task. While these methods share procedural similarities with our work, their fundamental goal is different. They are designed as architectural modifications to accelerate an existing model for a specific, pre-defined task. In contrast, our work uses a token compression mechanism not for online acceleration, but as the core of an \textit{offline distillation framework} whose sole purpose is to create a new, standalone, and general-purpose student model. Our focus is on distilling a transferable representation space, not on optimizing a single forward pass.

\noindent\textbf{Distillation of Foundation Models.}
Knowledge Distillation (KD) \cite{hinton2015distilling} is a powerful technique for model compression. When applied to foundation models, the specific knowledge being transferred is critical. We identify two main categories:

\textit{1. Distillation of Task-Specific Capabilities.} Most prior work distills a specific capability of a large model. In Vision-Language Models (VLMs), this is often the cross-modal alignment. For instance, TinyCLIP \cite{wu_2023_iccv} and other works \cite{pei_2023_cvpr} distill the contrastive loss or image-text affinity scores to create a smaller model that is good at zero-shot retrieval or classification. PromptKD \cite{li_2024_cvpr_prompt} focuses on domain-specific distillation. While effective, these methods produce `specialist' students that inherit one key function of the teacher, not its full, general-purpose representational power.

\textit{2. Distillation via Feature Mimicry.} A more direct approach is to force a student's feature embeddings to match the teacher's. CLIP-KD \cite{yang_2024_cvpr} provides a comprehensive study of this, using an L2 loss to align the student's and teacher's feature outputs. This is a form of direct \emph{feature mimicry}. While this transfers more general information than logit-based KD, it is still fundamentally different from our approach.

Our work, \methodname, establishes a third category: \emph{Representation Space Distillation}. We do not use direct mimicry. Instead, we propose a \emph{compress-and-reconstruct} objective. By forcing the student to learn a compressed representation (the SuperTokens) that can decompress back to the teacher's full set of feature embeddings, we compel it to learn a robust and efficient basis for the teacher's entire latent space. This information bottleneck objective is fundamentally different and, as we show experimentally, results in a more versatile and transferable student model, creating a true, unimodal foundation model proxy.

\section{Approach}
\label{sec:method}

The core of \methodname is a \emph{compress-and-reconstruct} framework designed to implement our Foundation Model Distillation (FMD) paradigm. The objective is to train a lightweight student model to learn a compact, yet powerful, representation that can faithfully approximate the rich latent space of a large, pre-trained teacher model. We first detail this overarching framework and its learning objective, then describe the novel components that enable it. An overview is provided in \cref{fig:overview}.

\subsection{The FMD Framework: Compressing and Reconstructing the Latent Space}

\begin{figure*}
  \centering
  \includegraphics[width=\linewidth]{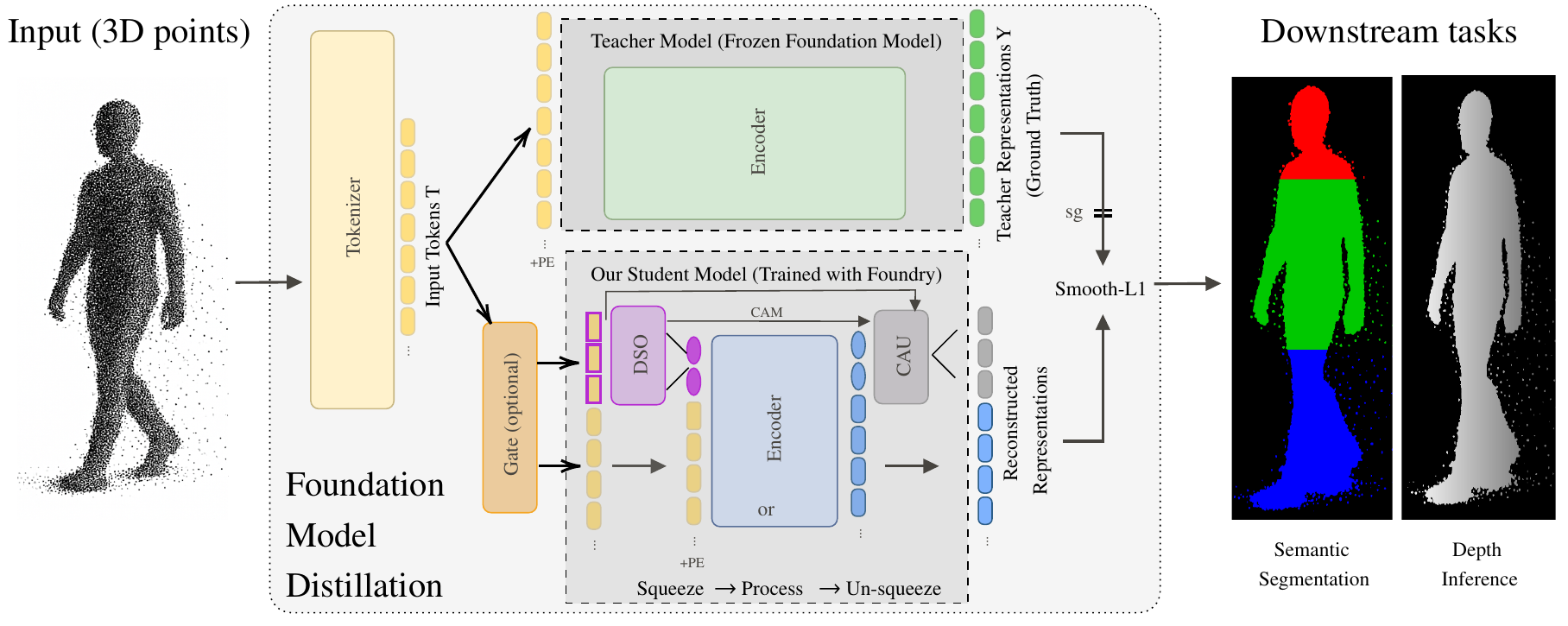}
  \caption{\textbf{Overview of the \methodname Framework.} Our distillation strategy follows a compress-and-reconstruct pipeline. The student model uses a Dynamic Supertoken Optimization (DSO) module to compress the input tokens into a small set of learnable SuperTokens. After processing by a lightweight encoder, a Cross-Attention Upsampling (CAU) module reconstructs an approximation of the teacher's latent space. The entire student is trained to minimize the reconstruction error, forcing the SuperTokens to become a powerful, compact basis for the teacher's representations.}
  \label{fig:overview}
\end{figure*}

Our distillation process consists of three conceptual steps, as illustrated in \cref{fig:overview}.

\noindent\textbf{1. Teacher Forward Pass (The Ground Truth).}
First, we establish the `ground truth` representation we aim to distill. For a given input point cloud $\mathbb{P}$, we follow the standard tokenization process of Point-BERT \cite{yu2022point} to produce a set of $c$ initial token embeddings $\mathbf{T} \in \mathbb{R}^{c\times d}$, augmented with positional encodings. These tokens are processed by the powerful, pre-trained teacher encoder, yielding a set of rich output embeddings $\mathbf{Y} \in \mathbb{R}^{c \times d}$. This set $\mathbf{Y}$ represents the target latent space that we want our student to be able to model.

\noindent\textbf{2. Student Forward Pass (Compress and Reconstruct).}
The student model's goal is to reproduce $\mathbf{Y}$ without the computational cost of the teacher. It takes the same initial tokens $\mathbf{T}$ and performs a two-stage process:
\begin{itemize}
    \item \emph{Compress:} The initial tokens $\mathbf{T}$ are fed into our \emph{Dynamic Supertoken Optimization (DSO)} module. This module compresses the information from all $c$ tokens into a much smaller, learnable set of $s$ \textit{SuperTokens} ($\mathbf{S} \in \mathbb{R}^{s \times d}$), where $s \ll c$. These SuperTokens form the compressed basis for the teacher's latent space.
    \item \emph{Reconstruct:} The compressed SuperTokens $\mathbf{S}$ are then processed by a lightweight student encoder. The updated SuperTokens, along with the original input tokens $\mathbf{T}$, are passed to our \emph{Cross-Attention Upsampling (CAU)} module. This module uses the compressed representation to reconstruct the teacher's full latent space, producing the final student output $\mathbf{\hat{Y}} \in \mathbb{R}^{c \times d}$.
\end{itemize}

\noindent\textbf{3. The Distillation Objective.}
The entire student model is trained end-to-end with a simple, direct objective: the reconstructed output $\mathbf{\hat{Y}}$ must match the teacher's ground truth output $\mathbf{Y}$. We use a Smooth L1 loss\footnote{We also experimented with MSE and obtained similar results when fine-tuning, but L1 is more robust to outliers, so it became our default choice.} for this reconstruction objective:
\begin{equation}
\mathcal{L}_{\text{distillation}} = \text{SmoothL1}(\mathbf{\hat{Y}}, \mathbf{Y}).
\label{eq:distillation-loss}
\end{equation}
This loss function directly enforces our FMD goal. By successfully minimizing this reconstruction error, the student is forced to learn SuperTokens that are an efficient and information-rich basis for the teacher's general-purpose representation space.

\subsection{Learning the Compressed Basis: Dynamic Supertoken Optimization (DSO)}
\label{sec:method:dso}

The compression stage is achieved by our DSO module, inspired by 3DLST \cite{lu20243d}, which is responsible for learning the SuperTokens.

\noindent\textbf{SuperTokens as a Learnable Basis.}
The set of SuperTokens, $\mathbf{S}=\{s_i\}_{i=1}^s \in \mathbb{R}^{s \times d}$, is a randomly initialized, learnable parameter of our model with a truncated normal distribution. Conceptually, it acts as a small, shared memory or a set of basis vectors that, over the course of distillation, will learn to represent the most salient semantic and geometric concepts found across the entire training dataset.

\noindent\textbf{Information Aggregation via Cross-Attention.}

To compress the input tokens $\mathbf{T}$ into the SuperTokens $\mathbf{S}$, we use a cross-attention mechanism \cite{vaswani2017attention}. The SuperTokens act as queries, ``attending'' to the input tokens (keys and values) to gather the most relevant information. This is formulated as:
\begin{equation}
\mathbf{Q} = \mathbf{S}\mathbf{W}_Q, \mathbf{K} = \mathbf{T}\mathbf{W}_K, \mathbf{V} = \mathbf{T}\mathbf{W}_V,
\label{eq:attention}
\end{equation}
where $W_{Q,K,V}$ are learnable projection matrices. We then compute a hard assignment\footnote{Using the Gumbel-Softmax \cite{DBLP:conf/iclr/JangGP17} with a straight-through \cite{DBLP:journals/corr/BengioLC13} estimator} by identifying the most similar SuperToken for each of the $c$ input tokens. This produces a one-hot \textbf{Cross-Attention Map (CAM)} of size $c \times s$, where each row has a single '1' indicating the chosen SuperToken:
\begin{equation}
\text{CAM}_{j,i} = 
\begin{cases} 
1 & \text{if } i = \underset{k}{\argmax}\left(\frac{\mathbf{q}_k \cdot \mathbf{k}_j}{\sqrt{d}}\right) \\ 
0 & \text{otherwise} 
\end{cases}
,
\label{eq:cam}
\end{equation}
for each input token $j \in \{1, ..., c\}$ and SuperToken $i \in \{1, ..., s\}$. This map is then used to perform a grouped average, updating each SuperToken with the mean of the value vectors of all input tokens assigned to it:
\begin{equation}
\mathbf{S}_{\text{updated}} = \frac{\text{CAM}^T\mathbf{V}}{\text{sum}(\text{CAM}^T, \text{axis=1})}.
\label{eq:cam-proj}
\end{equation}
Here, $\text{CAM}^T \mathbf{V}$ (a matrix of size $s \times d$) sums the value vectors for each group, and the division by the token counts (a vector of size $s$) normalizes each sum to produce the mean. This updated set of SuperTokens is then passed to the student encoder. Notably, we perform this semantic grouping before incorporating positional embeddings, ensuring the SuperTokens learn to represent features based on content rather than location. The positional embeddings are merged in the same way separately and added to the SuperTokens before the student encoder.

\subsection{Reconstructing the Latent Space: Cross-Attention Upsampling (CAU)}
\label{sec:method:cau}

After the lightweight student encoder has processed and refined the SuperTokens, the reconstruction stage is handled by the CAU module, adapted from 3DLST.

\noindent\textbf{Upsampling via Information Look-up.}
To compute the distillation loss, we must return from the compressed $s$-token space to the original $c$-token space. The CAU module achieves this by allowing each original token location to `look up' the relevant refined information from the updated SuperTokens. We reuse the CAM computed during the DSO step (\cref{eq:cam}) as a routing mechanism. The reconstruction is formulated as:
\begin{equation}
\mathbf{\hat{Y}} = \text{MLP}(\mathbf{T} + \text{CAM} \cdot \mathbf{S}_{\text{encoder\_out}}) \in \mathbb{R}^{c \times d}.
\label{eq:cau}
\end{equation}
A residual connection with the initial tokens $\mathbf{T}$ is crucial, as it reinjects local, high-frequency details that may have been lost during the compression, ensuring a high-fidelity reconstruction.

\subsection{Budget-Aware Inference via Gated Compression}
\label{sec:method:gate}
While our core framework creates a statically efficient model, we can extend it with a gating mechanism to enable dynamic, on-the-fly budget control at inference time.

\noindent\textbf{Gate Mechanism.}
We add a simple 2-layer MLP gate that takes the input tokens $\mathbf{T}$ and predicts a fusion probability $\pi_i$ for each token. At inference time, only tokens with $\pi_i > r$ (where $r$ is a user-defined threshold) are compressed by the DSO module. The remaining tokens bypass the compression and are processed alongside the SuperTokens in the student encoder. This allows a user to trade accuracy for performance by simply adjusting the threshold $r$.

\noindent\textbf{Regularization Objective.}
To train the gate to make meaningful selections, we add a regularization term to the distillation loss, which encourages the gate to activate:
\begin{equation}
\mathcal{L}_{\text{gate}} = -\lambda_{\text{gate}}\sum_{i=1}^{c}\pi_i.
\label{eq:gate-loss}
\end{equation}
The final objective function when the gate is enabled is $\mathcal{L} = \mathcal{L}_{\text{distillation}} + \mathcal{L}_{\text{gate}}$.

\subsection{Complexity Analysis}
We denote by $c$ the number of input tokens, by $s$ the maximum number of SuperTokens that can be selected (with $s\ll c$), by $d$ the embedding dimension, by $l$ the number of layers of the Transformer encoder, and by $r$ the ratio of selected tokens when the gate is used. \Cref{tab:complexity} shows complexity of each encoder and wrapper. When $l$ is very large, the computational overhead induced by the wrapper becomes marginal.
\begin{table*}
    \centering
    \resizebox{\textwidth}{!}{
        \begin{tabular}{l|ccc}
            \toprule
             Method & Encoder complexity & Wrapper complexity & Total \\
            \midrule
             Transformer & $\mathcal{O}(lc^2d+lcd^2)$ & - & $\mathcal{O}(lc^2d+lcd^2)$ \\
             \textbf{\methodname (Ours)} & $\mathcal{O}(ls^2d+lsd^2)$ & $\mathcal{O}(2scd + cd^2)$ & $\mathcal{O}(scd + ls^2d+lsd^2 + cd^2)$ \\
             \textbf{\methodname-Gate (Ours)} & $\mathcal{O}(l(min(s, c(1-r))+cr)^2d+l(min(s, c(1-r))+cr)d^2)$ & $\mathcal{O}(cd + 2srcd + cd^2)$ & $\mathcal{O}(srcd + l(min(s, c(1-r))+cr)^2d+l(min(s, c(1-r))+cr)d^2 + cd^2)$ \\
             \bottomrule
        \end{tabular}
    }
    \caption{\textbf{Complexity analysis}. For a small $s$, we see \methodname needs less computation to perform a forward pass compared to a regular Transformer. \methodname-Gate takes another parameter $r$ which can be fixed by the user or set dynamically by the gate itself. In the worst case $r=1$ (and $s=0$ because no tokens are selected for merging), we match Transformer complexity and best, complexity of \methodname.}
    \label{tab:complexity}
\end{table*}

\section{Experimental Results}
\label{sec:results}

We present experiments designed to validate our Foundation Model Distillation paradigm and the effectiveness of our \methodname framework. Our evaluation is structured to answer three key questions in order of importance:
\begin{enumerate}
    \item \textbf{Is the FMD paradigm effective?} We first demonstrate that distilling a general-purpose student is superior to creating task-specific specialists, especially for transfer learning and data efficiency (\cref{sec:validate_fmd}).
    \item \textbf{Is our SuperToken mechanism novel and necessary?} We then show that our compress-and-reconstruct objective with learnable SuperTokens outperforms simpler compression baselines (\cref{sec:validate_supertokens}).
    \item \textbf{How does \methodname perform?} Finally, we benchmark our validated method against other token compression techniques and analyze its performance, budget-awareness, and computational cost (\cref{sec:benchmarking} and beyond).
\end{enumerate}
All experiments were performed on a ViT-S teacher (12 layers, 384 embedding dimensions) pre-trained with Point-JEPA \cite{saito2025point}. Results with ViT-T are specified.

\subsection{Validating the Foundation Model Distillation Paradigm}
\label{sec:validate_fmd}
The core claim of our work is that creating a general-purpose student is more valuable than distilling specialists for each task. We validate this by comparing our single \methodname student against specialist students trained with traditional, task-specific knowledge distillation.

\noindent\textbf{Experimental Setup.}
We compare two approaches:
\begin{itemize}
    \item \textbf{FMD-Student (Ours):} A single \methodname student with 16 supertokens, distilled once on the general-purpose ShapeNet55 dataset \cite{chang2015shapenet}. This is our "generalist" model.
    \item \textbf{Specialist-Students:} Two students with an identical architecture to ours. The first (\textit{Specialist-SEG}) is trained via traditional KD to mimic the teacher's outputs on the ShapeNetPart \cite{yi2016scalable} segmentation task. The second (\textit{Specialist-CLS}) is trained similarly on the ShapeNet55 classification task. Both use only the KL divergence loss \cite{kullback1951information} for aligning the student and teacher output distributions without using ground truth labels.
\end{itemize}

\noindent\textbf{Generalist vs. Specialist on Downstream Tasks.}
We evaluate both native-task performance and transferability. As shown in \cref{tab:fmd_vs_specialist}, our generalist FMD-Student is highly competitive with the specialist models on their own area. For example, it achieves 89.95\% accuracy on ShapeNet55 classification, close to the specialist trained for that task. However, the true advantage of the FMD paradigm is revealed in transfer learning. When the specialist students are distilled, their performance collapses (e.g., the classification specialist's accuracy drops by nearly 15\%). In contrast, our single generalist student demonstrates robust and stable performance across both tasks, proving it has successfully inherited the teacher's general-purpose representational power.

\begin{table}[ht!]
    \centering
    \resizebox{\linewidth}{!}{
    \begin{tabular}{l|cc}
        \toprule
        Method & SN55 (Acc. $\uparrow$) & SNP ($\text{mIoU}_C$/$\text{mIoU}_I$ $\uparrow$) \\
        \midrule
        Teacher (Point-JEPA) & 90.54 & 83.91/85.73 \\
        \midrule
        \textbf{\methodname (Ours)} & \textbf{89.95} & \textbf{81.85/84.77} \\
        \midrule
        Specialist-CLS & 75.09\textsuperscript{*} & - \\
        Specialist-SEG & - & 61.88/65.72\textsuperscript{*} \\
        \bottomrule
    \end{tabular}
    }
    \caption{\textbf{Generalist vs. Specialist Distillation.} Our single FMD-Student, distilled once on ShapeNet55, maintains high performance when fine-tuned on both classification and segmentation. In contrast, small specialist students (marked with *), when distilled on the outputs of their respective specialist teachers to compress inputs to 16 supertokens, fail to perform well even on their native task, underscoring their limited effectiveness.}
    \label{tab:fmd_vs_specialist}
\end{table}

\noindent\textbf{Few-Shot Generalization.}
The robustness of a general-purpose model is most evident in low-data regimes. We fine-tune our distilled student on few-shot subsets of ModelNet40 following the Sharma and Kaul protocol \cite{sharma2020self}. As shown in \cref{tab:few_shot}, our student retains a remarkable degree of the teacher's few-shot learning capability, even with extreme compression down to a single SuperToken. For example, it achieves 91.8\% top-1 accuracy in a 10-shot setting, demonstrating that the distilled representations provide a strong semantic prior for learning from scarce data.

\begin{table}[ht!]
\centering
\resizebox{\columnwidth}{!}{
    \begin{tabular}{lcccc}
        \toprule        
        & \multicolumn{2}{c}{5-way} & \multicolumn{2}{c}{10-way}\\
        \cmidrule(lr){2-3} \cmidrule(lr){4-5}
        Method & 10-shot & 20-shot & 10-shot & 20-shot \\
        \midrule
        Point-JEPA & 96.1 \footnotesize{$\pm$ 3.45} & 98.2 \footnotesize{$\pm$ 1.32} & 93.4 \footnotesize{$\pm$ 4.23} & 95.3 \footnotesize{$\pm$ 3.35} \\
        \textbf{\paper (Ours) ($s=16$)} & \textbf{92.5 \footnotesize{$\pm$ 4.35}} & \textbf{95.2 \footnotesize{$\pm$ 2.82}} & \textbf{89.8 \footnotesize{$\pm$ 4.93}} & \textbf{92.2 \footnotesize{$\pm$ 4.96}} \\
        \textbf{\paper (Ours) ($s=1$)} & \textbf{91.8 \footnotesize{$\pm$ 4.02}} & \textbf{95.1 \footnotesize{$\pm$ 3.18}} & \textbf{88.3 \footnotesize{$\pm$ 6.56}} & \textbf{91.5 \footnotesize{$\pm$ 4.79}} \\
        \bottomrule
    \end{tabular}
}
    \caption{\textbf{Few-shot transfer performance on ModelNet40.} Even with a single SuperToken, the gateless student achieves 91.8\% (10-shot) and 95.1\% (20-shot) accuracy, approaching the Point-JEPA teacher.
    }
    \label{tab:few_shot}
\end{table}

\subsection{The Importance of the SuperToken Mechanism}
\label{sec:validate_supertokens}
Having established the value of the FMD paradigm, we now prove that our specific compress-and-reconstruct mechanism with learnable SuperTokens is the key to its success. We compare \methodname against two strong baselines designed to create a compressed student via simpler means.

\noindent\textbf{Baselines.}
\begin{itemize}
    \item \textbf{KMeans-Student:} Replaces our learnable DSO/CAU modules with a static K-Means \cite{DBLP:journals/tit/Lloyd82} clustering algorithm. We first compute K-Means centroids on the tokenizer's output embeddings and then fine-tune a student to target task. This tests the value of end-to-end learning.
    \item \textbf{FPS-Student (New Baseline):} Replaces our mechanism with naive pre-sampling. We use Farthest Point Sampling (FPS) \cite{DBLP:journals/tip/EldarLPZ97} to select a sparse subset of input tokens \textit{before} they are processed, and distill a student on this sparse representation. This tests if a learned compression is better than a simple geometric sampling.
\end{itemize}

As shown in \cref{tab:supertokens_vs_baselines}, our full \methodname model significantly outperforms both baselines. Substituting the learnable DSO with static K-Means causes a massive accuracy drop of over 13\% on ShapeNet55. This proves that learning the SuperToken basis vectors \textit{jointly} with the distillation process is critical. Furthermore, our method also surpasses the FPS-Student, demonstrating that our learned semantic compression captures richer information than a simple geometric pre-sampling of the input space.

\begin{table}[h]
    \centering
\resizebox{\columnwidth}{!}{
    \begin{tabular}{l|c}
        \toprule
        Method / Token Grouping Strategy & ShapeNet55 (Acc. $\uparrow$) \\
        \midrule
        \textbf{\methodname (Ours, learnable DSO and CAU)} & \textbf{89.68} \\
        \midrule
        KMeans-Student (static clustering) & 76.08 \\
        FPS-Student (pre-sampling) & 87.56 \\
        \bottomrule
    \end{tabular}
    }
    \caption{\textbf{Effect of the token-grouping strategy.} Our learnable DSO mechanism significantly outperforms baselines using static K-Means clustering or naive Farthest Point Sampling. This validates that our end-to-end, attention-based aggregation is essential for creating a high-fidelity student. All model encoders are completely frozen during distillations and fine-tunings using 16 supertokens, prototypes or centers.}
    \label{tab:supertokens_vs_baselines}
\end{table}

\subsection{Benchmark Performance and Analysis}
\label{sec:benchmarking}

\noindent\textbf{Overall gateless performance.}
After validating our core claims, we now benchmark the performance of our gateless \methodname variant against the teacher and other token compression methods across a wide range of datasets. Six classification datasets (ModelNet40~\cite{wu20153d} (MN40), OmniObject3D~\cite{deitke2023objaverse} (OO3D), the three ScanObjectNN~\cite{uy2019revisiting} (SONN) splits, and ShapeNet55~\cite{chang2015shapenet} (SN55)) and one part segmentation (ShapeNetPart~\cite{yi2016scalable} (SNP)) are benchmarked.

As shown in \cref{tab:finetunings-v1}, despite compressing the latent representation to a small basis of $s \ll c$ tokens, \methodname maintains accuracy within 1-2\% of the full teacher model on most datasets, mainly on synthetics and within a few points on challenging real-world datasets. While acknowledging that averaging different metrics (Accuracy and mIoUs) is not strictly rigorous, the `Avg' column provides a useful high-level indicator of overall performance trends. This summary view suggests that our method consistently outperforms naive baselines like random sampling and exhibits more stable cross-task performance than methods like ToMe and PiToMe. For instance, with just $s=4$ SuperTokens, the student achieves 91.25\% accuracy on ModelNet40 and  84.67/84.81~mIoU on ShapeNetPart. This indicates that our learned SuperTokens capture a truly reusable and task-agnostic latent basis. Unlike inference-based methods, we see that our FMD method can be combined via compress-and-reconstruct objective with other methods, such as ToMe or PatchMerger, coming very close to baseline performance while adapting the model to process increasingly compressed information.

\begin{figure}
    \centering
    \includegraphics[width=\linewidth]{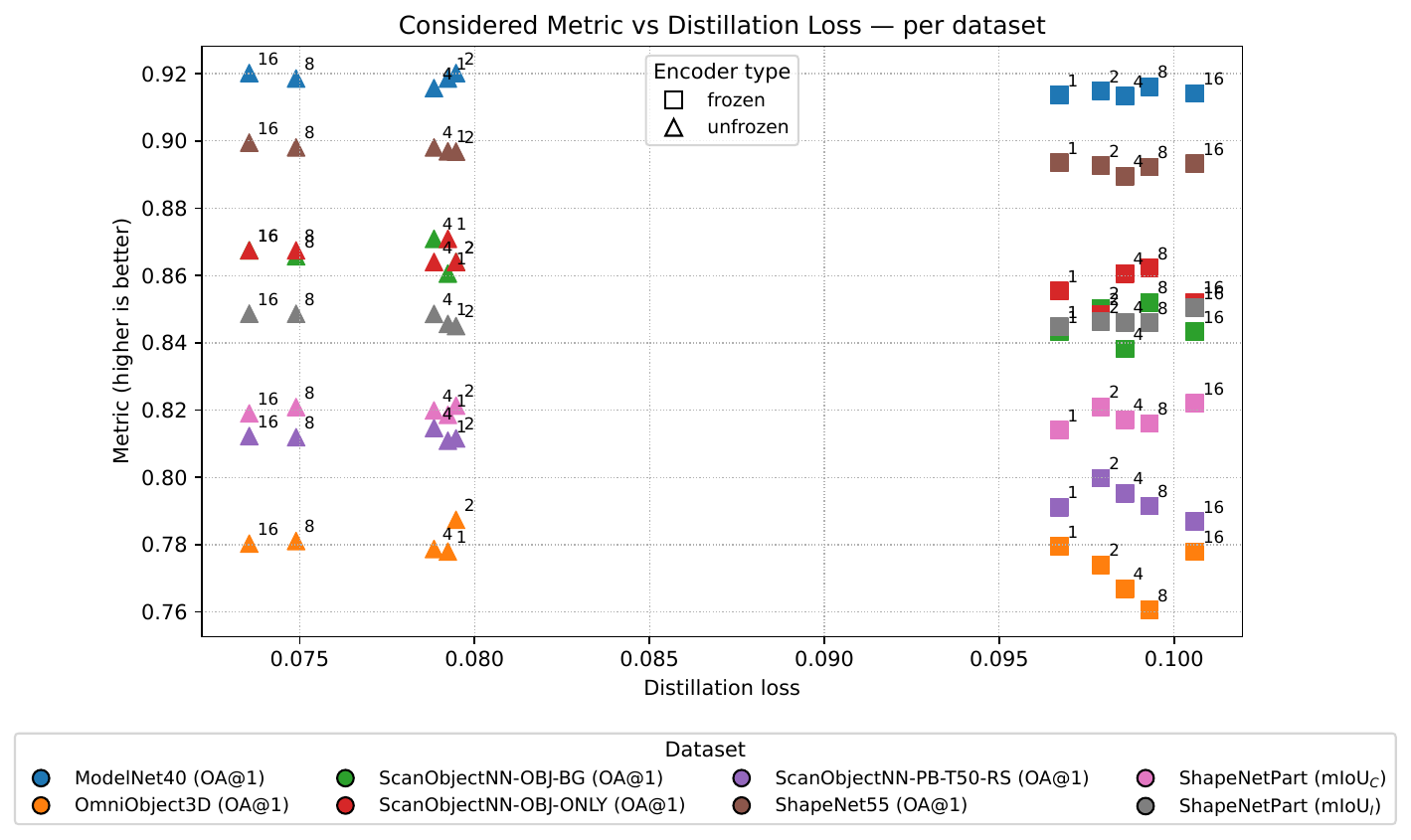}
    \caption{
    \textbf{Relationship between distillation loss and fine-tuning accuracy.}
    Each point corresponds to a distilled model with a different number of SuperTokens~($s$) across multiple datasets.
    A clear inverse correlation is observed, with diminishing improvements beyond $s{\leq}4$,
    indicating that reconstruction fidelity saturates once the student sufficiently spans the teacher’s latent space.
  }
    \label{fig:v1:metric-vs-distillation-loss}
\end{figure}

\noindent\textbf{Distillation loss as a proxy for transferability.}
\cref{fig:v1:metric-vs-distillation-loss} illustrates the relationship between distillation loss and fine-tuning performance. A clear inverse correlation is observed, with diminishing gains beyond $s=4$, indicating that once the compressed basis sufficiently spans the teacher’s latent space, further capacity increases yield marginal benefit.  
This suggests that the model has adapted to processing compressed tokens rather than the original, uncompressed tokens.

\begin{table*}[ht!]
\resizebox{\linewidth}{!}{
\begin{tabular}{l|c|c|cccccccc}
\toprule
\# supertokens & Frozen & Avg & MN40 & OO3D & SONN-\texttt{OBJ-BG} & SONN-\texttt{OBJ-ONLY} & SONN-\texttt{PB-T50-RS} & SN55 & \multicolumn{2}{c}{SNP} \\
or tokens ($c$) & student &&&&&&&& $\text{mIoU}_C$ & $\text{mIoU}_I$ \\
\midrule
baseline ($c=64$) & - & 87.72 & 93.02 \footnotesize{$\pm$ 0.14} & 82.25 \footnotesize{$\pm$ 0.37} & 91.84 \footnotesize{$\pm$ 0.47} & 88.38 \footnotesize{$\pm$ 0.41} & 86.05 \footnotesize{$\pm$ 0.42} & 90.54 \footnotesize{$\pm$ 0.14} & 83.91 \footnotesize{$\pm$ 0.25} & 85.73 \footnotesize{$\pm$ 0.15} \\

\midrule
\multicolumn{11}{c}{\textit{\textbf{Foundry (Ours)}}} \\
$s=16$ & \cmark & 84.33 & 91.41 \footnotesize{$\pm$ 0.32} & 77.30 \footnotesize{$\pm$ 0.61} & 84.74 \footnotesize{$\pm$ 0.43} & 84.97 \footnotesize{$\pm$ 0.72} & 79.93 \footnotesize{$\pm$ 1.10} & 89.56 \footnotesize{$\pm$ 0.21} & 81.88 \footnotesize{$\pm$ 0.47} & 84.82 \footnotesize{$\pm$ 0.32} \\
$s=16$ & \xmark & 84.87 & 91.75 \footnotesize{$\pm$ 0.25} & 77.80 \footnotesize{$\pm$ 0.21} & 86.23 \footnotesize{$\pm$ 0.46} & 86.29 \footnotesize{$\pm$ 0.43} & 80.36 \footnotesize{$\pm$ 0.90} & 89.87 \footnotesize{$\pm$ 0.09} & 81.87 \footnotesize{$\pm$ 0.04} & 84.82 \footnotesize{$\pm$ 0.07} \\
$s=16$ (ViT-T) & \xmark & 83.75 & 91.49 \footnotesize{$\pm$ 0.42} & 75.98 \footnotesize{$\pm$ 0.24} & 83.42 \footnotesize{$\pm$ 0.87} & 84.80 \footnotesize{$\pm$ 0.98} & 79.38 \footnotesize{$\pm$ 0.17} & 88.67 \footnotesize{$\pm$ 0.33} & 81.68 \footnotesize{$\pm$ 0.24} & 84.56 \footnotesize{$\pm$ 0.10} \\
$s=8$ & \cmark & 84.38 & 91.71 \footnotesize{$\pm$ 0.10} & 76.48 \footnotesize{$\pm$ 0.36} & 85.60 \footnotesize{$\pm$ 0.55} & 85.14 \footnotesize{$\pm$ 0.95} & 80.28 \footnotesize{$\pm$ 1.00} & 89.42 \footnotesize{$\pm$ 0.17} & 81.84 \footnotesize{$\pm$ 0.34} & 84.57 \footnotesize{$\pm$ 0.05} \\
$s=8$ & \xmark & 84.76 & 91.63 \footnotesize{$\pm$ 0.20} & 77.32 \footnotesize{$\pm$ 0.72} & 86.23 \footnotesize{$\pm$ 0.34} & 86.06 \footnotesize{$\pm$ 1.05} & 80.46 \footnotesize{$\pm$ 0.73} & 89.61 \footnotesize{$\pm$ 0.24} & 81.95 \footnotesize{$\pm$ 0.19} & 84.85 \footnotesize{$\pm$ 0.01} \\
$s=4$ & \cmark & 84.25 & 91.25 \footnotesize{$\pm$ 0.14} & 76.96 \footnotesize{$\pm$ 0.28} & 84.45 \footnotesize{$\pm$ 0.60} & 85.94 \footnotesize{$\pm$ 1.04} & 79.61 \footnotesize{$\pm$ 0.14} & 89.26 \footnotesize{$\pm$ 0.27} & 81.84 \footnotesize{$\pm$ 0.18} & 84.67 \footnotesize{$\pm$ 0.08} \\
$s=4$ & \xmark & 84.64 & 91.48 \footnotesize{$\pm$ 0.08} & 76.69 \footnotesize{$\pm$ 1.02} & 86.00 \footnotesize{$\pm$ 1.22} & 85.89 \footnotesize{$\pm$ 0.52} & 80.70 \footnotesize{$\pm$ 0.93} & 89.74 \footnotesize{$\pm$ 0.06} & 81.78 \footnotesize{$\pm$ 0.30} & 84.81 \footnotesize{$\pm$ 0.07} \\
$s=2$ & \cmark & 84.17 & 91.52 \footnotesize{$\pm$ 0.53} & 77.09 \footnotesize{$\pm$ 0.32} & 84.51 \footnotesize{$\pm$ 0.62} & 84.51 \footnotesize{$\pm$ 1.08} & 79.98 \footnotesize{$\pm$ 0.14} & 89.34 \footnotesize{$\pm$ 0.14} & 81.84 \footnotesize{$\pm$ 0.36} & 84.55 \footnotesize{$\pm$ 0.11} \\
$s=2$ & \xmark & 84.87 & 91.76 \footnotesize{$\pm$ 0.22} & 78.32 \footnotesize{$\pm$ 0.66} & 85.83 \footnotesize{$\pm$ 0.53} & 86.12 \footnotesize{$\pm$ 0.50} & 80.78 \footnotesize{$\pm$ 0.33} & 89.66 \footnotesize{$\pm$ 0.04} & 81.98 \footnotesize{$\pm$ 0.21} & 84.47 \footnotesize{$\pm$ 0.04} \\
$s=1$ & \cmark & 84.17 & 91.60 \footnotesize{$\pm$ 0.30} & 77.19 \footnotesize{$\pm$ 0.67} & 84.80 \footnotesize{$\pm$ 0.53} & 85.03 \footnotesize{$\pm$ 0.75} & 79.47 \footnotesize{$\pm$ 0.33} & 89.29 \footnotesize{$\pm$ 0.13} & 81.46 \footnotesize{$\pm$ 0.08} & 84.54 \footnotesize{$\pm$ 0.10} \\
$s=1$ & \xmark & 84.58 & 91.72 \footnotesize{$\pm$ 0.14} & 77.72 \footnotesize{$\pm$ 0.08} & 84.57 \footnotesize{$\pm$ 1.30} & 86.06 \footnotesize{$\pm$ 0.96} & 80.65 \footnotesize{$\pm$ 0.38} & 89.65 \footnotesize{$\pm$ 0.06} & 81.76 \footnotesize{$\pm$ 0.14} & 84.50 \footnotesize{$\pm$ 0.09} \\
\midrule
\multicolumn{11}{c}{\textit{Random sampling - inference from baseline}} \\
$c=16$ & - & 75.20 & 90.57 \footnotesize{$\pm$ 0.41} & 65.19 \footnotesize{$\pm$ 0.81} & 78.95 \footnotesize{$\pm$ 1.19} & 72.70 \footnotesize{$\pm$ 1.14} & 69.74 \footnotesize{$\pm$ 0.91} & 87.80 \footnotesize{$\pm$ 0.22} & 66.96 \footnotesize{$\pm$ 0.71} & 69.71 \footnotesize{$\pm$ 0.16} \\
$c=1$ & - & 21.46 & 18.33 \footnotesize{$\pm$ 0.66} & 03.82 \footnotesize{$\pm$ 0.32} & 23.22 \footnotesize{$\pm$ 1.55} & 21.31 \footnotesize{$\pm$ 1.35} & 19.90 \footnotesize{$\pm$ 0.68} & 15.54 \footnotesize{$\pm$ 0.23} & 33.70 \footnotesize{$\pm$ 0.30} & 35.89 \footnotesize{$\pm$ 0.16} \\

\midrule
\multicolumn{11}{c}{\textit{Updating group size to reduce number of generated tokens - inference from baseline}} \\
$c=16$ & - & 41.66 & 63.45 \footnotesize{$\pm$ 0.47} & 13.29 \footnotesize{$\pm$ 0.42} & 11.76 \footnotesize{$\pm$ 0.16} & 22.24 \footnotesize{$\pm$ 0.29} & 12.74 \footnotesize{$\pm$ 0.17} & 59.99 \footnotesize{$\pm$ 0.19} & 72.98 \footnotesize{$\pm$ 0.21} & 76.84 \footnotesize{$\pm$ 0.08} \\
$c=1$ & - & 15.78 & 03.88 \footnotesize{$\pm$ 0.19} & 00.47 \footnotesize{$\pm$ 0.13} & 09.24 \footnotesize{$\pm$ 0.63} & 15.85 \footnotesize{$\pm$ 0.82} & 09.37 \footnotesize{$\pm$ 0.00} & 04.25 \footnotesize{$\pm$ 0.09} & 40.30 \footnotesize{$\pm$ 0.44} & 42.90 \footnotesize{$\pm$ 0.34} \\

\midrule
\multicolumn{11}{c}{\textit{ToMe~\cite{DBLP:conf/iclr/BolyaFDZFH23} - inference from baseline (no additional training needed)}} \\
$s=16$ & - & 73.80 & 89.10 \footnotesize{$\pm$ 0.23} & 50.51 \footnotesize{$\pm$ 0.39} & 82.19 \footnotesize{$\pm$ 0.56} & 84.22 \footnotesize{$\pm$ 0.62} & 73.87 \footnotesize{$\pm$ 0.50} & 84.89 \footnotesize{$\pm$ 0.24} & 63.09 \footnotesize{$\pm$ 0.40} & 62.51 \footnotesize{$\pm$ 0.08} \\
$s=8$ & - & 66.79 & 84.02 \footnotesize{$\pm$ 0.38} & 33.54 \footnotesize{$\pm$ 0.78} & 76.87 \footnotesize{$\pm$ 1.30} & 79.88 \footnotesize{$\pm$ 0.96} & 67.32 \footnotesize{$\pm$ 0.62} & 79.68 \footnotesize{$\pm$ 0.26} & 56.81 \footnotesize{$\pm$ 0.43} & 56.23 \footnotesize{$\pm$ 0.14} \\
$s=4$ & - & 59.88 & 74.92 \footnotesize{$\pm$ 0.65} & 20.91 \footnotesize{$\pm$ 0.59} & 71.57 \footnotesize{$\pm$ 1.05} & 74.29 \footnotesize{$\pm$ 0.54} & 61.52 \footnotesize{$\pm$ 0.49} & 72.05 \footnotesize{$\pm$ 0.30} & 51.59 \footnotesize{$\pm$ 0.52} & 52.19 \footnotesize{$\pm$ 0.12} \\
$s=2$ & - & 53.19 & 62.76 \footnotesize{$\pm$ 0.44} & 11.21 \footnotesize{$\pm$ 0.36} & 67.76 \footnotesize{$\pm$ 1.03} & 68.49 \footnotesize{$\pm$ 0.84} & 56.30 \footnotesize{$\pm$ 0.62} & 61.66 \footnotesize{$\pm$ 0.30} & 47.47 \footnotesize{$\pm$ 0.45} & 49.88 \footnotesize{$\pm$ 0.12} \\
$s=1$ & - & 46.15 & 48.35 \footnotesize{$\pm$ 0.31} & 07.20 \footnotesize{$\pm$ 0.20} & 61.20 \footnotesize{$\pm$ 0.89} & 62.87 \footnotesize{$\pm$ 0.95} & 51.69 \footnotesize{$\pm$ 0.46} & 42.15 \footnotesize{$\pm$ 0.18} & 45.36 \footnotesize{$\pm$ 0.19} & 50.37 \footnotesize{$\pm$ 0.09} \\

\midrule
\multicolumn{11}{c}{\textit{ToMe~\cite{DBLP:conf/iclr/BolyaFDZFH23} - Trained using \textbf{FMD (Ours)} + finetuning}} \\
$s=16$ & \xmark & 86.52 & 93.07 & 80.95 & 90.02 & 87.44 & 83.66 & 89.84 & 82.41 & 84.76 \\
$s=1$  & \xmark & 85.87 & 92.14 & 80.08 & 89.85 & 86.40 & 83.00 & 89.84 & 81.14 & 84.50 \\
\midrule
\multicolumn{11}{c}{\textit{PiToMe~\cite{DBLP:conf/nips/TranNNNLXSZNN24} - Trained using \textbf{FMD (Ours)} + finetuning}} \\
$s=16$ & \xmark & 86.51 & 92.83 & 81.97 & 89.33 & 87.95 & 83.66 & 89.88 & 81.77 & 84.65 \\
$s=1$  & \xmark & 86.14 & 92.55 & 80.95 & 90.19 & 86.23 & 82.89 & 89.76 & 81.84 & 84.74 \\
\midrule
\multicolumn{11}{c}{\textit{PatchMerger~\cite{DBLP:journals/corr/abs-2202-12015} - Trained using \textbf{FMD (Ours)} + finetuning}} \\
$s=16$ & \xmark & 87.59 & 92.83 & 82.76 & 91.39 & 88.99 & 85.53 & 90.11 & 83.59 & 85.55 \\
$s=1$  & \xmark & 87.14 & 92.18 & 82.13 & 90.71 & 88.12 & 85.05 & 89.71 & 83.64 & 85.54 \\
\bottomrule
\end{tabular}
}
\caption{
\textbf{Performance of the gateless \paper student across benchmarks.}
Each student is distilled from a frozen foundation teacher using the compress--reconstruct objective.
Despite compressing to $s\!\ll\!c$ latent tokens, the distilled models remain within 1--2\,\% of teacher accuracy on most classification and segmentation tasks, and within a few points on challenging real-world datasets.
We vary the number of SuperTokens $s \in \{1, 2, 4, 8, 16\}$ and report mean~$\pm$~standard deviation over multiple runs. For \paper, random token sampling and changing token group size, $s$ and $c$ correspond to the number of tokens at the input and output of the student encoder. For ToMe, PiToMe, and PatchMerger, it is the number of tokens we seek to obtain at the output. Complete per-dataset results and variance analyses are reported in the supplementary material.
}
\label{tab:finetunings-v1}
\end{table*}

\noindent\textbf{Selective compression.}
We analyze the dynamic, budget-aware variant \paper-Gate, which integrates the gating mechanism introduced in \cref{sec:method:gate}. The gate regularizer $\lambda_{\text{gate}}$ controls the proportion of tokens that bypass compression, allowing on-the-fly trade-offs between computational cost and representational fidelity. We compute the Spearman correlation between the final fine-tuning performances and the used regularization value. We observe an anti-correlation, for both frozen and unfrozen settings. On ModelNet40, the correlation values are -0.45 and -0.13 respectively for frozen and unfrozen distillation. When frozen, anti-correlation is more important because the model is not updated during distillation. Complete results are available in supplementary material.
\Cref{fig:v3-inference-controllability} shows  that by reducing the number of tokens passed to the core encoder, the performance degradation compared to the baseline model trained with $r=0.5$ is acceptable. However, when $r$ is increasing (less tokens are fused), performance degrades more significantly (see supplementary material). This suggests that the model has adapted to process compressed tokens rather than on the original, uncompressed tokens.
\begin{figure}
    \centering
    \includegraphics[width=\linewidth]{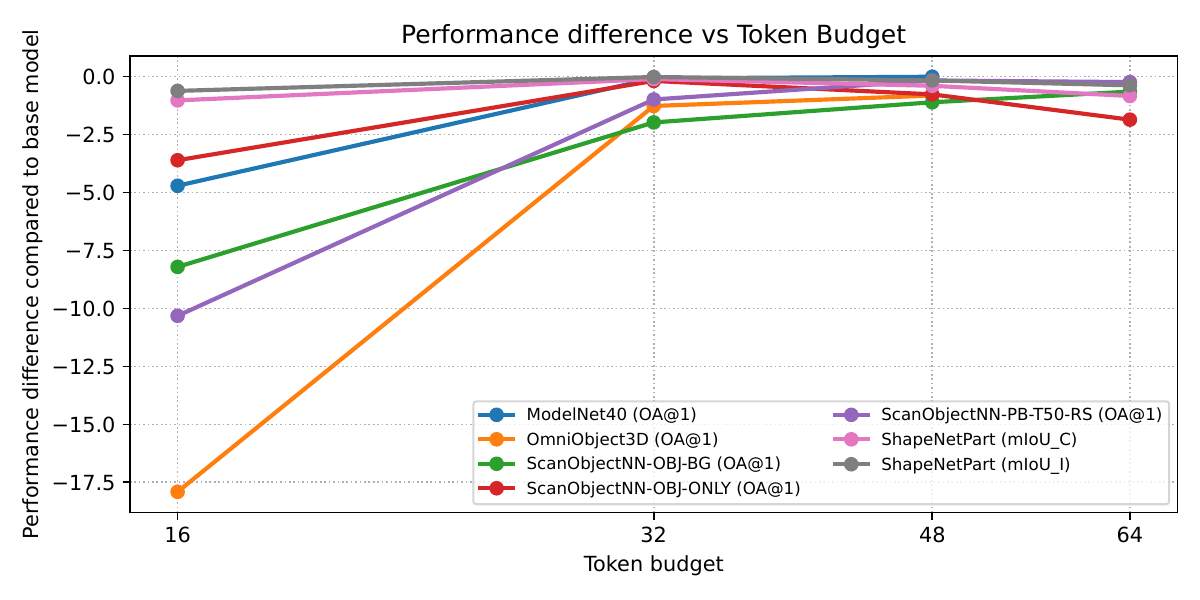}
    \caption{\textbf{Finetuning results vs token budget for
\paper-Gate}. We use the best backbone on each dataset
with $\lambda_{\text{gate}}=10^{-11}$ and no weight decay on the gate module. The token budget represents the maximum number of tokens we accept as input to the core encoder ($r$ is dynamic and changes for each input example). We report top-1 overall accuracy for classification tasks and mIoU per class and per instance for part segmentation tasks. All metrics are computed on the validation set for each dataset.}
    \label{fig:v3-inference-controllability}
\end{figure}

\subsection{Empirical Computational Cost}
\label{sec:experiments:cost}
We evaluate runtime, memory, and computational efficiency on an NVIDIA RTX~A3000 6GB Laptop GPU.  
\cref{tab:empirical-computational-cost} reports results for object-level inference
, and \cref{tab:indoor-scene-computation-simulation} reports large-scale scene inference ($2^{18}$ points). Reported experiments include tokenization and encoding steps.

\noindent\textbf{Small-scale inference.}
Relative to the baseline’s 478\,GFLOPs, \paper reduces forward-pass cost to \textbf{178\,G} ($s=16$), \textbf{156\,G} ($s=8$), and \textbf{137\,G} ($s=1$), while maintaining comparable accuracy.  
Average latency decreases from 0.09~s to 0.05–0.06~s without memory overhead.
The ViT-T variant further reduces computation to \textbf{144\,GFLOPs} at similar accuracy, demonstrating the scalability of the framework to smaller backbones.

\noindent\textbf{Large-scale scenes.}
In the indoor-scene simulation (\cref{tab:indoor-scene-computation-simulation}), both the baseline and ToMe exceed the 6\,GB VRAM limit and fail to execute, whereas \paper completes inference with a \textbf{4.0\,GB} footprint and \textbf{4.6\,s} forward time (throughput $\approx 0.22$\,obj/s).  
This confirms that the proposed distillation enables practical deployment on edge or mobile-class GPUs, where full foundation models are otherwise infeasible.

\noindent\textbf{Dynamic compute control.}
As shown in the lower blocks of \cref{tab:empirical-computational-cost}, \paper-Gate allows specifying a fixed token budget or directly adjusting the fusion ratio~$r$ at inference.  
Varying $r$ between 0.1 and 0.9 modulates latency predictably while incurring data-sensitive degradation in downstream accuracy, enabling real-time compute adaptation under resource constraints.

\begin{table}[ht!]
\resizebox{\linewidth}{!}{
\begin{tabular}{l|cc|cccccc}
\toprule
& \# supertokens & Token & Params & MACs & FLOPs & GPU Mem & Avg runtime & Throughput \\
Method & & budget & (M) & (G) & (G) & (MiB) & (s) & (obj/s) \\
\midrule
baseline & - & - & 21.84 & 238.624 & 478.128 & 1143.0 & 0.09 & 747.11 \\
ToMe \cite{DBLP:conf/iclr/BolyaFDZFH23} ($r_{\text{tome}}=16$) & - & - & 21.84 & 115.412 & 231.319 & 1143.0 & 0.06 & 1091.42 \\
ToMe \cite{DBLP:conf/iclr/BolyaFDZFH23} ($r_{\text{tome}}=8$) & - & - & 21.84 & 97.8966 & 196.233 & 1143.0 & 0.05 & 1174.07 \\
ToMe \cite{DBLP:conf/iclr/BolyaFDZFH23} ($r_{\text{tome}}=4$) & - & - & 21.84 & 90.0449 & 180.505 & 1143.0 & 0.05 & 1213.10 \\
ToMe \cite{DBLP:conf/iclr/BolyaFDZFH23} ($r_{\text{tome}}=2$) & - & - & 21.84 & 86.572 & 173.548 & 1143.0 & 0.05 & 1219.81 \\
ToMe \cite{DBLP:conf/iclr/BolyaFDZFH23} ($r_{\text{tome}}=1$) & - & - & 21.84 & 85.0621 & 170.523 & 1143.0 & 0.05 & 1217.40 \\

\midrule
\multicolumn{9}{c}{\textit{\textbf{Foundry (Ours)}}} \\
\paper & 16 & - & 22.73 & 88.9903 & 178.394 & 1135.5 & 0.06 & 1160.97 \\
\paper (ViT-T) & 16 & - & 6.33 & 71.6618 & 143.698 & 1073.0 & 0.05 & 1335.26 \\
\paper & 8 & - & 22.73 & 78.042 & 156.464 & 1135.5 & 0.05 & 1271.58 \\
\paper & 4 & - & 22.73 & 72.5678 & 145.498 & 1135.5 & 0.05 & 1309.76 \\
\paper & 2 & - & 22.73 & 69.8307 & 140.016 & 1135.5 & 0.05 & 1349.58 \\
\paper & 1 & - & 22.73 & 68.4622 & 137.274 & 1135.5 & 0.05 & 1365.92 \\
\paper-Gate & 16 & - & 22.78 & 157.593 & 315.812 & 1135.7 & 0.11 & 591.60 \\

\midrule
\multicolumn{9}{c}{\textit{\textbf{Foundry (Ours) - Inference token-budgeted}}} \\
\paper-Gate & 16 & 16 & 22.78 & 87.8332 & 176.076 & 1135.7 & 0.07 & 864.02 \\
\paper-Gate & 16 & 32 & 22.78 & 107.651 & 215.774 & 1135.7 & 0.08 & 756.92 \\
\paper-Gate & 16 & 48 & 22.78 & 126.073 & 252.674 & 1135.7 & 0.09 & 679.59 \\

\midrule
\multicolumn{9}{c}{\textit{\textbf{Foundry (Ours) - change fusion ratio $r$}}} \\
\paper-Gate (r=0.1) & 16 & - & 22.78 & 87.8332 & 176.076 & 1135.7 & 0.07 & 881.45 \\
\paper-Gate (r=0.2) & 16 & - & 22.78 & 89.1921 & 178.798 & 1135.7 & 0.07 & 875.99 \\
\paper-Gate (r=0.3) & 16 & - & 22.78 & 159.254 & 319.139 & 1135.7 & 0.09 & 751.83 \\
\paper-Gate (r=0.4) & 16 & - & 22.78 & 161.028 & 322.694 & 1135.7 & 0.11 & 588.67 \\
\paper-Gate (r=0.5) & 16 & - & 22.78 & 157.593 & 315.812 & 1135.7 & 0.11 & 593.70 \\
\paper-Gate (r=0.6) & 16 & - & 22.78 & 152.006 & 304.621 & 1135.7 & 0.10 & 623.29 \\
\paper-Gate (r=0.7) & 16 & - & 22.78 & 152.006 & 304.621 & 1135.7 & 0.10 & 623.66 \\
\paper-Gate (r=0.8) & 16 & - & 22.78 & 152.006 & 304.621 & 1135.7 & 0.10 & 620.73 \\
\paper-Gate (r=0.9) & 16 & - & 22.78 & 152.006 & 304.621 & 1135.7 & 0.10 & 623.58 \\

\bottomrule
\end{tabular}
}
\caption{
\textbf{Runtime and memory comparison on 64$\!\times\!$1024-point inputs.}
\paper reduces FLOPs from 478\,G to 178-137\,G while maintaining comparable accuracy.
Wall-clock latency decreases from 0.09\,s to 0.05–0.06\,s with reduced used memory. We also provide inference values on CPU in the supplementary material.
}
\label{tab:empirical-computational-cost}
\end{table}
\begin{table}[ht!]
\resizebox{\linewidth}{!}{
\begin{tabular}{l|cc|cccccc}
\toprule
& \# supertokens & Token & Params & MACs & FLOPs & GPU Mem & Avg runtime & Throughput \\
Method & & budget & (M) & (G) & (G) & (MiB) & (s) & (obj/s) \\
\midrule
baseline & - & - & 21.84 & \multicolumn{5}{c}{\textit{OOM}} \\
ToMe \cite{DBLP:conf/iclr/BolyaFDZFH23} & - & - & 21.84 & \multicolumn{5}{c}{\textit{OOM}} \\

\midrule
\multicolumn{9}{c}{\textit{\textbf{Foundry (Ours)}}} \\
\paper & 16 & - & 22.73 & 268.719 & 538.821 & 4034.1 & 4.59 & 0.22 \\
\paper (ViT-T) & 16 & - & 6.33 & 264.837 & 531.037 & 3971.5 & 4.57 & 0.22 \\
\paper & 8 & - & 22.73 & 268.547 & 538.476 & 4034.1 & 4.57 & 0.22 \\
\paper & 4 & - & 22.73 & 268.461 & 538.304 & 4034.1 & 4.58 & 0.22 \\
\paper & 2 & - & 22.73 & 268.418 & 538.218 & 4034.1 & 4.56 & 0.22 \\
\paper & 1 & - & 22.73 & 268.396 & 538.174 & 4034.1 & 4.60 & 0.22 \\

\midrule
\multicolumn{9}{c}{\textit{\textbf{Foundry (Ours) - Inference token-budgeted}}} \\
\paper-Gate & 16 & 16 & 22.78 & 269.526 & 540.438 & 4034.3 & 4.57 & 0.22 \\
\paper-Gate & 16 & 32 & 22.78 & 269.836 & 541.058 & 4034.3 & 4.57 & 0.22 \\
\paper-Gate & 16 & 64 & 22.78 & 270.497 & 542.382 & 4034.3 & 4.57 & 0.22 \\
\paper-Gate & 16 & 128 & 22.78 & 271.818 & 545.028 & 4034.3 & 4.57 & 0.22 \\
\paper-Gate & 16 & 256 & 22.78 & 274.46 & 550.321 & 4034.3 & 4.57 & 0.22 \\
\paper-Gate & 16 & 512 & 22.78 & 279.745 & 560.907 & 4034.3 & 4.60 & 0.22 \\
\paper-Gate & 16 & 1024 & 22.78 & 290.315 & 582.08 & 4034.3 & 4.59 & 0.22 \\

\bottomrule
\end{tabular}
}
\caption{
\textbf{Scalability of \paper on large indoor scenes ($2^{18}$ points).}
The baseline and ToMe exceed 6\,GB VRAM and fail to execute,
whereas \paper completes inference with a 4.0\,GB footprint and $\sim$4.6\,s forward time.
This demonstrates memory-efficient scalability suitable for edge deployment.
}
\label{tab:indoor-scene-computation-simulation}
\end{table}

\subsection{Ablation Study}
\label{sec:experiments:ablation}
We conduct architectural variations as ablations to isolate the contributions of key architectural components. Other key points are discussed in \cref{sec:validate_fmd,sec:validate_supertokens}.
\cref{tab:ablations} compares several architectural alternatives.  
Replacing the hard $\argmax$ assignment with soft attention slightly increases the reconstruction loss (0.0736 $\rightarrow$ 0.0787).  
Introducing QKV biases or CAU layer normalization yields similarly marginal increases in the loss, justifying our simpler, more streamlined architectural design.  
Unfreezing the student during distillation, however, reduces the loss from 0.1006 to 0.0736. This clearly shows that allowing the student to learn enables it to process compressed information more effectively and demonstrates once again that the compressive learning objective that has been put in place is sufficient.
\begin{table}[ht!]
\centering
\resizebox{\linewidth}{!}{
\begin{tabular}{l|c|c}
\toprule
Ablations & Frozen student & Distillation loss \\
\midrule
\paper & \cmark & 0.1006 \\
\paper & \xmark & 0.0736 \\
\midrule
$\argmax$ $\rightarrow$ softmax & \xmark & 0.0787 \\
Bias on QKV in DSO & \xmark & 0.0763 \\
Output layer norm at CAU & \xmark & 0.0748 \\
\bottomrule
\end{tabular}
}
\caption{
\textbf{Architectural ablations of the distillation framework.}
Soft assignment, QKV biases, and CAU layer normalization each slightly increase the reconstruction loss. Unfreezing the student during distillation further reduces the loss (0.1006$\!\rightarrow\!$0.0736), indicating that most benefits arise from the compressive learning objective itself and that giving the model the ability to update its weights helps it better process compressed tokens.
}
\label{tab:ablations}
\end{table}

\section{Conclusion}
\label{sec:conclusion}
This paper introduced Foundation Model Distillation (FMD), a new paradigm for making large SSL models practical for the edge. We presented Foundry, the first FMD framework, which successfully distills a general-purpose 3D foundation model into a compact proxy that is a fraction of the size. Our core contribution, a compress-and-reconstruct objective based on learnable SuperTokens, creates a student that retains the teacher's transferability and data efficiency—a feat that task-specific distillation fails to achieve. By producing a portable, standalone, and efficient model, Foundry paves the way for deploying powerful 3D perception on resource-constrained systems, from autonomous robots to AR headsets. Our results indicate that it is possible to distill a foundation model into a compact backbone that preserves its transferability across classification, segmentation, and few-shot tasks, while reducing computational cost. We focus on a single 3D self-supervised teacher (Point-JEPA) and leave extending \methodname to other 3D foundation models and modalities for future work; it remains to be seen how well our compress–reconstruct design transfers in those settings.

\paragraph{Acknowledgement.}{
This work was supported by the Agence Nationale pour la Recherche (ANR) under award number ANR-19-CHIA-0017. This work was performed using HPC resources from GENCI–IDRIS (Grant 2024-AD011014730R1).
}

\bibliographystyle{ieeenat_fullname}
\bibliography{main}

\clearpage 
\appendix

\onecolumn
\begin{center}
  \Large\textbf{\thetitle}\\
  \vspace{0.5em}Supplementary Material \\
  \vspace{1.0em}
\end{center}


\section{Datasets and metrics}
\label{sec:datasets}

This section briefly reports information for each used dataset in this paper.

\subsection{Datasets}
\paragraph{ShapeNet55 \cite{chang2015shapenet} (SN55).}{
The dataset contains synthetic 3D meshes from 55 categories, split into 41,952 shapes for training and 10,518 for validation and testing. Below is the list of shape names: \textit{airplane, trash bin, bag, basket, bathtub, bed, bench, birdhouse, bookshelf, bottle, bowl, bus, cabinet, camera, can, cap, car, cellphone, chair, clock, keyboard, dishwasher, display, earphone, faucet, file cabinet, guitar, helmet, jar, knife, lamp, laptop, loudspeaker, mailbox, microphone, microwaves, motorbike, mug, piano, pillow, pistol, flowerpot, printer, remote, rifle, rocket, skateboard, sofa, stove, table, telephone, tower, train, watercraft, washer}.
}
\paragraph{ModelNet40 \cite{wu20153d} (MN40).}{
The dataset consists of synthetic 3D meshes from 40 distinct classes, with 9,840 samples for training and 2,468 for validation and testing. The shape names are: \textit{airplane, bathtub, bed, bench, bookshelf, bottle, bowl, car, chair, cone, cup, curtain, desk, door, dresser, flower pot, glass box, guitar, keyboard, lamp, laptop, mantel, monitor, night stand, person, piano, plant, radio, range hood, sink, sofa, stairs, stool, table, tent, toilet, tv stand, vase, wardrobe, xbox}.
}
\paragraph{ScanObjectNN \cite{uy2019revisiting} (SONN).}{
ScanObjectNN comprises 2,304 training examples and 581 validation/test samples from 15 semantic classes. The samples are real-scanned objects divided into three splits: (i) \texttt{OBJ-BG} includes both object and background, (ii) \texttt{OBJ-ONLY} contains only the object, and \texttt{PB-T50-RS} objects and backgrounds perturbed samples with challenging random augmentations to increase classification complexity. Below is the list of shape names: \textit{bag, bin, box, cabinet, chair, desk, display, door, shelf, table, bed, pillow, sink, sofa, toilet}.
}
\paragraph{OmniObject3D \cite{wu2023omniobject3d} (OO3D).}{
The dataset is divided into training and validation/testing sets with a 80/20 split per class. Typically, for a class with 100 elements, the last 20 are used for validation/testing, and the rest for training. If a class has 2 or fewer elements, at least one is used for testing and one for training. The final split counts are 4,641 for training, and 1,270 for validation/testing. The dataset contains 216 categories, including: \textit{anise, antique, apple, asparagus, backpack, ball, bamboo shoots, banana, battery, beauty blender, bed, belt, biscuit, book, bottle, bowl, box, boxed beverage, bread, broad bean, broccoli, broccolini, brush, brussels sprout, bucket noodle, bumbag, bun, burrito, cabinet, cake, calculator, candle, candy, canned beverage, carrot, cauliflower, chair, cheese, cherry, chess, chicken leg, chili, china, chinese chess, chocolate, clock, coconut, conch, corn, cucumber, cup, dice, dinosaur, dish, doll, donut, drawing, drum, dumbbell, dumpling, durian, dustbin, earplug, egg, egg tart, eraser, facial cream, fan, fig, fire extinguisher, flash light, flower pot, flute, fork, frisbee, garage kit, garlic, ginger, glasses, glasses case, gloves, green bean cake, guitar, hair dryer, hairpin, hamburger, hami melon, hammer, hand cream, handbag, hat, haw thorn, hazelnut, helmet, hot dog, house, insole, kennel, kettle, keyboard, kite, kiwifruit, knife, laptop, laundry detergent, lemon, light, lipstick, litchi, longan, loquat, lotus root, magnet, mango, mangosteen, medicine bottle, microwaveoven, monitor, mooncake, mouse, mushroom, nipple, onion, orange, ornament, ornaments, oyster, package, pad, pan, pancake, pastry, peach, peanut, pear, pen, picnic basket, pie, pillow, pineapple, pinecone, pistachio, pitaya, pizza, plant, plug, pomegranate, popcorn, potato, potato chips, power strip, projector, puff, pumpkin, razor, red jujube, red wine glass, remote control, rice cake, ricecooker, rubik cube, sandwich, sausage, scissor, shampoo, shaomai, shoe, shrimp, skateboard, soap, sofa, spanner, speaker, squash, starfish, steamed bun, stool, strawberry, suitcase, sushi, sweet potato, sword bean, table, table tennis bat, tape measure, teapot, teddy bear, thermos, thimble, timer, tissue, tomato, tooth brush, tooth paste, toothpick box, toy animals, toy boat, toy bus, toy car, toy motorcycle, toy plane, toy plant, toy train, toy truck, tvstand, umbrella, vase, waffle, wallet, walnut, watch, water chestnut, watermelon, whistle, yam, zongzi}.
}
\paragraph{ShapeNetPart \cite{yi2016scalable} (SNP).}{
This part segmentation dataset contains 16 object types and 50 part categories, comprising 16,881 synthetic 3D shapes. The data is split into a training set of 13,998 shapes and a validation/testing set of 2,874 shapes.
}
\paragraph{Objaverse \cite{deitke2023objaverse}.}{
This unannotated dataset contains 661,575 objects in the train set and 3000 in the validation set.
PointLLM \cite{DBLP:conf/eccv/XuWWCPL24} provides annotations via description on each object but it was not utilized in our framework.
}

\subsection{Metrics}
\paragraph{Classification task.}{We use top-1 overall accuracy which can be sometimes abbreviated as \texttt{OA@1}.}
\paragraph{Part segmentation task.}{We use mean intersection over union abbreviated as \texttt{mIoU} for classes and instances. We denote them respectively by $\text{mIoU}_C$ and $\text{mIoU}_I$. When unspecified, $\text{mIoU}_C$ is used as \texttt{mIoU}.}

\section{Implementation details}
We give some details about the implementation of each tested method.

\subsection{Pre-trainings}
\paragraph{Point-JEPA \cite{saito2025point}.}{
We pre-train a Transformer \cite{vaswani2017attention} encoder using the Point-JEPA method. The size of the encoder has the same hyper-parameters as ViT-S \cite{DBLP:conf/iclr/DosovitskiyB0WZ21}. We use the exact same hyper-parameters and architecture for pre-training except floating-point precision (\texttt{float32} vs \texttt{float16}), and we apply truncated normal as weight initialization function \cite{devlin2019bert} with rescaling w.r.t. layer depth (GPT-2 layer-wise weight rescaling). Results can diverge a bit from the initial paper due to these additions. This model is called the \textit{teacher} or \textit{baseline}.
}

\subsection{Distillations}
\paragraph{\paper.}{
\Cref{fig:overview} in the main paper illustrates the core idea. We use the architecture described before with our pre-training. As said in the main paper, for the DSO block we use the Gumbel-Softmax \cite{DBLP:conf/iclr/JangGP17} with a straight-through \cite{DBLP:journals/corr/BengioLC13} estimator for differentiability. In contrast to 3DLST \cite{lu20243d}, we do not double the output channel of CAU for recovering the teacher token dimension.
}
\paragraph{\paper-Gate.}{
The Gate is a 2-layer MLP with a fixed hidden dimension of 128. The final gate activation is a sigmoid function to produce a probability for each token. When distilling or fine-tuning, no gradients from the target task loss are passed to the gate.
}
\paragraph{Specialists.}{
For the specialists presented in \cref{tab:fmd_vs_specialist} in main paper, we use fine-tuned models (pre-trained with Point-JEPA) on two datasets, ShapeNet55 (Specialist-CLS) and ShapeNetPart (Specialist-SEG). For each, we have a classical Transformer. We apply distillation only with KL divergence \cite{kullback1951information} without ground truth attachment to only use the teacher model as we present. We distill on their respective datasets to obtain a set of supertokens for compressing input tokens. We freeze the core encoder, tokenizer, and positional embedder but allow the heads to be updated during the process.
}
\paragraph{FPS-Student.}{
The idea is to distill an existing backbone with a fixed number of centers for tokenizing the input point cloud to another fixed number of centers with its particular tokenizer. We apply the same distillation strategy as \paper while changing the architecture so that the new student may be compatible with distillation. Instead of using one single grouping function with $c=64$ and $k=32$, we use a second tokenizer with another grouping function with $c_2=16$ and $k=2\cdot p/c_2$ with $p$ being the number of the initial point cloud (we set 2 for adding overlap between groups). Note that the data augmentation is shared for the two branches as well as the first $c$ sampled centers up to $c_2$ which are sampled by the Farthest Point Sampling (FPS) \cite{DBLP:journals/tip/EldarLPZ97} algorithm. As the number of output tokens differs between teacher and student, we use a feature upsampler \cite{DBLP:conf/nips/QiYSG17} to recover the last tokens of $c$ by interpolating existing $c_2$ to $c$.
}
\paragraph{ToMe \cite{DBLP:conf/iclr/BolyaFDZFH23} and PiToMe \cite{DBLP:conf/nips/TranNNNLXSZNN24}.}{
We use our strategy to distill both methods. For obtaining 16 output tokens, we set $r_{\text{tome}}=[32,16]$ which after the first attention block, removes 32 tokens and at the second, removes another 16. Same for PiToMe but using token ratio instead of a fixed number of tokens to end with. The ten last layers process only 16 tokens. When we have 1 single output token, $r_{\text{tome}}=[32,16,8,4,2,1]$. We track the source to be able to place information of merged tokens. For merged tokens, we put information of the corresponding remaining token. We also set proportional attention to \texttt{True} to add a bias in the attention matrix based on already fused tokens. We track and add proportional bias during distillation and fine-tunings.
}
\paragraph{PatchMerger \cite{DBLP:journals/corr/abs-2202-12015}.}{
Once again, we use our strategy to distill the encoder with PatchMerger. For obtaining 16 output tokens, we use the default hyper-parameter by placing the module between the 6th and 7th Transformer layers. We track the source to be able to place information of merged tokens. For merged tokens, we put information of the corresponding remaining token. We track when distilling and fine-tuning.
}

\subsection{Fine-tunings}
\paragraph{Classification.}{
We use the architecture and method defined in Point-JEPA for all models except KMeans-Student. For this method, we tokenize all training samples and use them for creating $s$ prototypes with K-Means \cite{DBLP:journals/tit/Lloyd82} clustering algorithm. After, we encode these prototypes into a frozen pre-trained backbone. Finally, we retrieve all tokens using the assignation matrix and we pool tokens by concatenating \texttt{mean+max} as Point2Vec \cite{zeid2023point2vec} does for probing and classifying to obtain one single token to use as input for our head.
}
\paragraph{Part segmentation.}{
We use the architecture and method defined in Point-JEPA.
}
\paragraph{Few-Shot Classification.}{
We use the architecture and method defined in Point-JEPA.
}
\paragraph{Linear Probing.}{
We use the architecture and method defined in Point-JEPA.
}

\subsection{Inferences}
\paragraph{Random Sampling.}{
We use fine-tuned models and change at inference-time the number of tokens given to the encoder. We tokenize the input point cloud and obtain $c$ centers. We randomly subsample $c$ to $c_{\text{rs}}$ which we set beforehand to match the number of supertokens $s$ we want to match.
}
\paragraph{Larger group size.}{
We use fine-tuned models and change at inference-time the $c$ and $k$ values. We reduce $c$, and $k$ is obtained via the following calculation $k=2\cdot p/c$.
}
\paragraph{\paper-Gate.}{
For a fixed $r$ value and dynamic setting with token-budget, we use fine-tuned models and we change $r$ at inference-time without making any additional training. Specifically, when we budget the number of tokens to pass to the encoder, we select a specific $r$ value to have $c-s$ tokens to merge into $s$ supertokens maximum. We provide the code in \cref{lst:budget-aware-gate}.
\begin{lstlisting}[language=Python, caption={\textbf{Implementation of the token-budgeted \paper-Gate}. We use PyTorch as the deep learning framework. For quick prototyping, we implement only for one single input sample.}, label={lst:budget-aware-gate}, captionpos=b]
class BudgetAwareMLPTokenSelector(MLPTokenSelector):
    def __init__(self, token_budget: int, num_supertokens: int, embed_dim: int, hidden_dim: int = 128, act: Callable = nn.GELU()) -> None:
        super().__init__(embed_dim, hidden_dim, act)
        self.token_budget = token_budget
        self.num_supertokens = num_supertokens
    
    def _gate_forward(self, x: torch.Tensor) -> torch.Tensor:
        B, T, _ = x.shape

        h = self.act(self.fc1(x))
        logits = self.fc2(h)  # (B, T, 1)
        probs = torch.sigmoid(logits).view(B, T, 1)
        return probs

    def _select_ratio(self, probs: torch.Tensor):
        B, T, _ = probs.shape

        if T <= self.token_budget:
            return 1.0

        sorted_vals, _ = probs[:, :, 0].sort(dim=1, descending=True)

        num_to_select = T - self.token_budget + self.num_supertokens
        if num_to_select == T:
            fusion_ratio = probs.new_full((B,), -torch.finfo(probs.dtype).eps)
        else:
            fusion_ratio = sorted_vals[:, num_to_select+1]  # we put +1 because in forward we use > (if)

        return fusion_ratio.reshape(B, 1, 1).repeat(1, T, 1)

    def forward(self, x: torch.Tensor, inference: bool = False, **kwargs) -> MLPTokenSelectorOutput:
        _, T, _ = x.shape

        probs = self._gate_forward(x)
        fusion_ratio = self._select_ratio(probs)

        selection_hard = (probs > fusion_ratio).float()
        if inference: # non differentiable
            selection = selection_hard
        else:  # straight-through method
            selection = probs + (selection_hard - probs).detach()
        
        mask = (selection > fusion_ratio).squeeze(-1)  # (B, T), boolean
        seqlens_selected = mask.sum(dim=1)

        final_num_tokens = (T - seqlens_selected + self.num_supertokens).max().item()
        if final_num_tokens > self.token_budget:
            print("more tokens used than token budget, diff:", final_num_tokens - self.token_budget)

        outputs = MLPTokenSelectorOutput(
            weighted_tokens=x * selection + x * (1 - selection),
            selected_tokens=x[mask],
            selected_seqlens=seqlens_selected,
            unselected_tokens=x[~mask],
            unselected_seqlens=T - seqlens_selected,
            selected_indices=torch.nonzero(mask.flatten(), as_tuple=False).flatten(),
            unselected_indices=torch.nonzero((~mask).flatten(), as_tuple=False).flatten(),
            probs=probs,
            selection=selection,
            mask=mask,
        )
        
        return outputs
    
    @staticmethod
    def from_mlp_token_selector(gate: MLPTokenSelector, token_budget: int, num_supertokens: int) -> "BudgetAwareMLPTokenSelector":
        new_gate = BudgetAwareMLPTokenSelector(
            token_budget,
            num_supertokens,
            embed_dim=gate.embed_dim,
            hidden_dim=gate.hidden_dim,
            act=gate.act,
        )
        new_gate.load_state_dict(gate.state_dict())

        if gate.training:
            new_gate.train()
        
        return new_gate
\end{lstlisting}
}
\paragraph{ToMe.}{
We use fine-tuned models and change at inference-time the number of tokens at the encoder output. We choose $r_{\text{tome}}$ accordingly to the desired final number of output tokens. We track source tokens only when resolving the part segmentation task.
}


\section{Theoretical FLOPs Computation}
\label{sec:flops}
This section presents the theoretical FLOPs computation for our modules and final models. All expressed formulas are for 32-bit floating point operations. We denote $S,N,D,nh,hd,d,L$ for the maximum number of supertokens, input number of tokens, embedding dimension, number of heads, head dimension, gate hidden dimension, and number of Transformer layers respectively.

\subsection{Transformer}
\begin{equation}
\text{MLP}_{\text{FLOPs}}=2\times (2\times N\times 4\times D^2)=16ND^2
\end{equation}
\begin{equation}
\text{SA}_{\text{FLOPs}}= 6ND^2 + 2N^2D + 2N^2D + 2ND^2 = 4N^2D+8ND^2
\end{equation}
\begin{align}
\text{Transformer}_{\text{FLOPs}}=& L\times (\text{SA}_{\text{FLOPs}} + \text{MLP}_{\text{FLOPs}}) = L \times (4N^2D+8ND^2 + 16ND^2) \\
=& L \times (4N^2D+24ND^2) = 4LND(N + 6D)
\end{align}

\subsection{\paper and \paper-Gate modules}
\paragraph{Dynamic Supertoken Optimization \cite{lu20243d}.}{
\begin{align}
\text{DSO}_{\text{FLOPs}}=&(2SD^2+4ND^2)+2SND+2nhShdD+2SD^2\\
=&4SD^2+4ND^2+4SND=4D^2(S+N)+4SND
\end{align}
}
\paragraph{Cross-Attention-guided Upsampling \cite{lu20243d}.}{
\begin{align}
\text{CAU}_{\text{FLOPs}}=&nhSN + 2nhSN + nhSN + 2nhSNhd + \text{MLP}_{\text{FLOPs}} \\
=&4nhSN+2SND+16ND^2=2SN(2nh+D)+16ND^2
\end{align}
}
\paragraph{Gate.}{
\begin{align}
\text{Gate}_{\text{FLOPs}}= 2NDd+2NdD = 4NDd
\end{align}
}

\subsection{\paper and \paper-Gate architectures}{
\begin{align}
\text{\papershort}_{\text{FLOPs}} =& \text{Transformer}_{\text{FLOPs}}(N=S) + \nonumber\\
&\text{DSO}_{\text{FLOPs}} + \text{CAU}_{\text{FLOPs}}\\
=&[4LSD(S+6D)] + (4SD^2+4ND^2+4SND) + (4nhSN+2SND+16ND^2) \\
=& 4SD^2+20ND^2+6SND+4nhSN+4LSD(S+6D) \\
=& 4D^2(S+5N)+2SN(3D+2nh)+4LSD(S+6D)
\end{align}
Let $R=\left\lfloor (1-r)N \right\rfloor, r\in [0,1]$ be the number of fused tokens by the gate and $U=N-R$ the number of unfused tokens.
\begin{align}
\text{\papershort-Gate}_{\text{FLOPs}}(R)=& \text{Gate}_{\text{FLOPs}} + \text{DSO}_{\text{FLOPs}}(N=R) + \text{Transformer}_{\text{FLOPs}}(N=S+U) +\text{CAU}_{\text{FLOPs}}(N=R) \\
=& (4NDd) + (4SD^2+4RD^2+4SRD) + [4LD(S+U)(S+U+6D)] +\nonumber\\
&(4nhSR+2SRD+16RD^2) \\
=& 4NDd+4SD^2+20RD^2+ 6SRD+4nhSR+ [4LD(S+U)(S+U+6D)] \\
=& 4D^2(S+5R)+ (2SR)(3D+2nh)+4NDd + [4LD(S+U)(S+U+6D)]
\end{align}
}

\section{Additional Results}
\label{sec:additional-results}

\subsection{Additional ablations}
\paragraph{Pre-training with supertokens.}{
For completeness, we provide results on pre-training a Transformer with Point-JEPA method directly using supertokens. The DSO/CAU modules are considered as part of the encoder. \Cref{tab:linear-probing-v2} shows linear probing results at the end of the pre-training.
\begin{table}[ht!]
\centering
\resizebox{0.65\linewidth}{!}{
\begin{tabular}{l|ccc}
\toprule
Method & MN40 (OA@1) & OO3D (OA@1) & SONN-\texttt{OBJ-BG} (OA@1) \\
\midrule
Point-JEPA & 93.01 \footnotesize{$\pm$ 0.18} & 77.39 \footnotesize{$\pm$ 0.52} & 83.24 \footnotesize{$\pm$ 0.66} \\
Point-JEPA + DSO+CAU & 78.12 & 30.63 & 51.29 \\
\bottomrule
\end{tabular}
}
\caption{\textbf{Linear probing results of SuperTokens-Point-JEPA.} We only used datasets for classification task. Mean $\pm$ std over 10 runs for baseline.}
\label{tab:linear-probing-v2}
\end{table}
We can clearly see that with the same hyper-parameters we set for SuperTokens, the differences in performance are very large. We believe that the tasks of compressing information via supertokens and extracting as much information as possible via the self-supervision method lead to a conflict that limits the final capacity of the pre-trained model with this architecture. This demonstrates that using a pre-trained model on all tokens and then performing a distillation step remains the simplest and, above all, the best solution.
}

\paragraph{MSE vs Smooth-L1.}{
\Cref{fig:v1:mse-vs-smoothl1-loss} illustrates the impact of the distillation loss choice during model fine-tunings. During distillations and fine-tunings, the backbone is kept frozen. We observe a similar trend with similar values. No loss is recommended.
\begin{figure}
    \centering
    \includegraphics[width=0.5\linewidth]{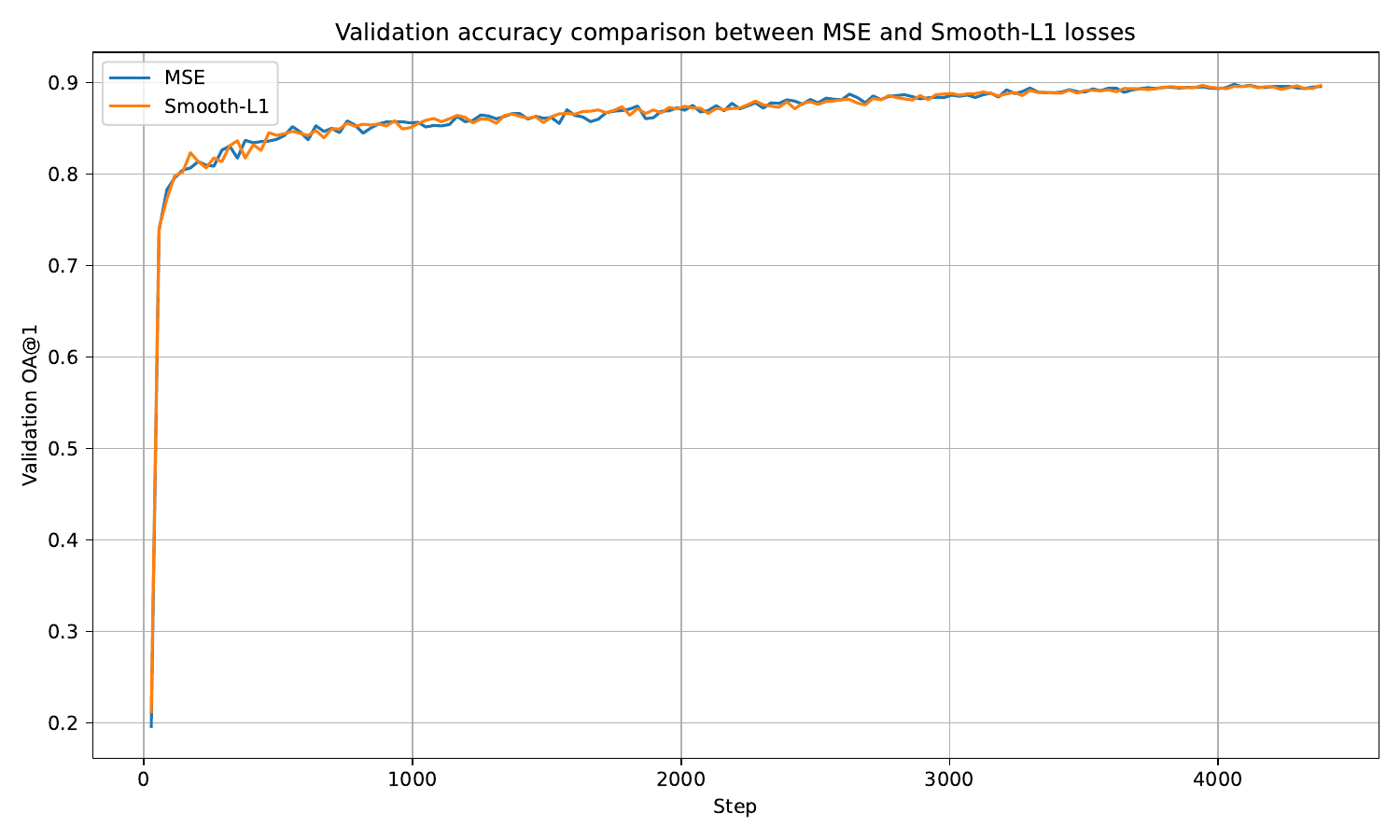}
    \caption{\textbf{Comparison on distillation loss.} We use \paper with a loss update from Smooth-L1 to MSE. To show differences, we show validation top-1 accuracy curves during fine-tuning on ShapeNet55 after distillation using the specified loss.}
    \label{fig:v1:mse-vs-smoothl1-loss}
\end{figure}
}

\subsection{Detailed fine-tuning results}
\paragraph{\paper.}{
\Cref{tab:finetunings-v1-detailed} describes detailed results when fine-tuning the distilled backbone using different numbers of supertokens. The maximum number of epochs for classification and part segmentation tasks are 150 and 300 respectively. In addition, we traditionally unfreeze the backbone respectively at 100 and 0 epoch.
\input{tables/finetuning-v1-complete}
}

\paragraph{\paper-Gate.}{
Similarly, we have \cref{tab:finetunings-v3-detailed} which details results when using the gate module. For simplifying the results visualization, \cref{fig:v3:finetuning-metric-vs-gate-reg-frozen-unfreeze} plots the metric values of fine-tuned models for each dataset at each gate regularization value.
\input{tables/finetuning-v3-complete}
\begin{figure}[ht]
    \centering
    \begin{subfigure}{0.3\textwidth}
        \centering
        \includegraphics[width=\textwidth]{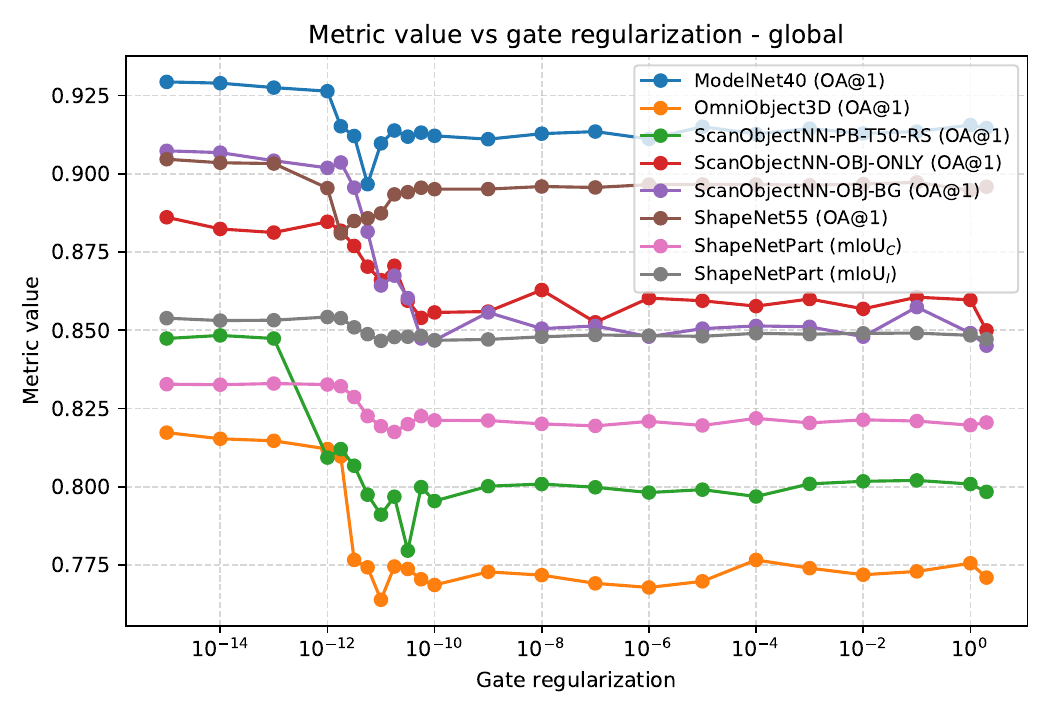}
        \caption{All distillation type}
    \end{subfigure}
    \hfill
    \begin{subfigure}{0.3\textwidth}
        \centering
        \includegraphics[width=\textwidth]{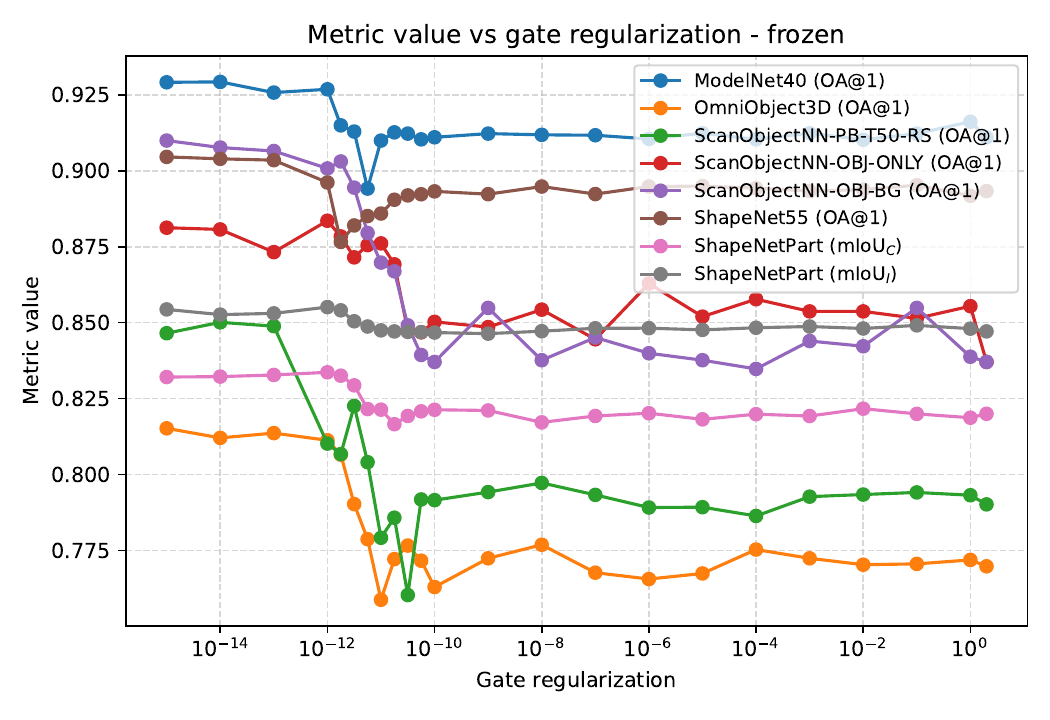}
        \caption{Frozen encoder during distillation}
    \end{subfigure}
    \hfill
    \begin{subfigure}{0.3\textwidth}
        \centering
        \includegraphics[width=\textwidth]{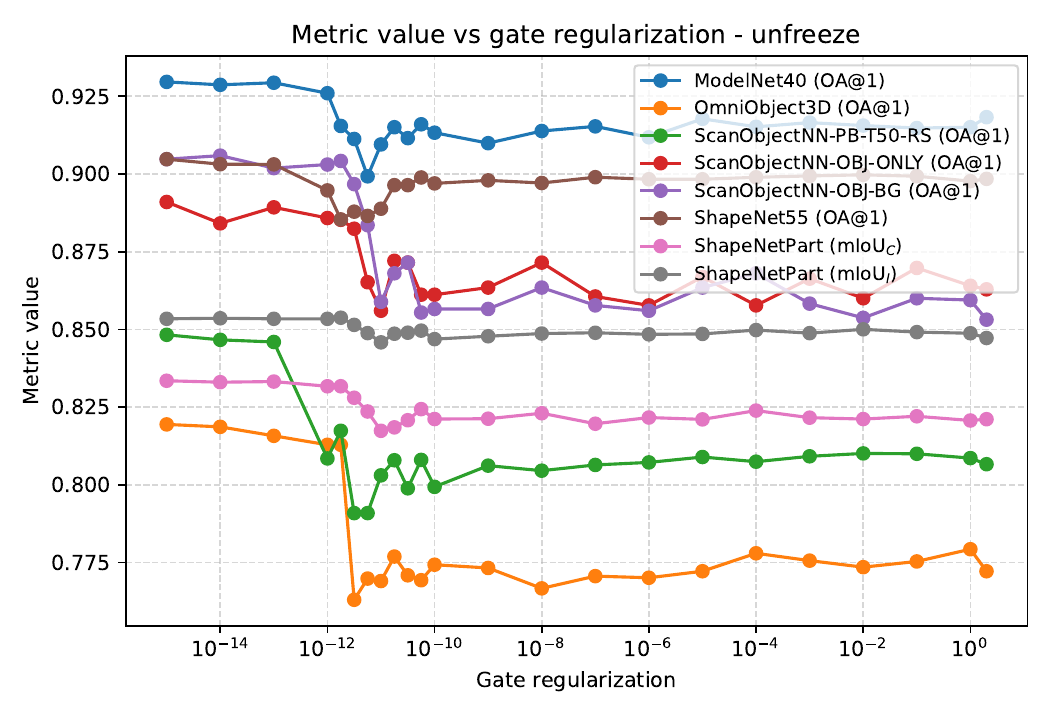}
        \caption{Unfrozen encoder during distillation}
    \end{subfigure}
    \caption{\textbf{Fine-tuning results vs $\lambda_{\text{gate}}$.} No weight decay used on the gate module during distillation are shown. All metrics have been computed on the validation set for each dataset.}
    \label{fig:v3:finetuning-metric-vs-gate-reg-frozen-unfreeze}
\end{figure}

\Cref{tab:v3-correlation-gate-reg-metrics} shows the Spearman correlation values for each dataset between final metrics and gate regularization as a single global value. We observe an inverse correlation between performances and gate regularization. The more we regularize for compressing tokens, the less downstream performance is obtained.
\input{tables/correlation-v3}

\Cref{fig:v3:distillation-loss-token-compression-rate-vs-gate-reg} presents the content distillation loss of distilled models for each gate regularization value with specification when we freeze/unfreeze the model during distillation and when we add weight decay on gate. For $\lambda_{\text{gate}}\!\in\![10^{-12}, 10^{-10}]$, we observe an increase in the token compression ratio, leading to 100\% of the tokens selected by the gate being merged. We also observe an increase in the distillation loss. By combining these two observations, we can state that the distillation loss has a correlation with the token compression ratio. Additionally, with the help of \cref{tab:v3-correlation-gate-reg-metrics}, by transitivity, the distillation loss is also correlated with downstream task performances.
\begin{figure}[ht]
    \centering
    \begin{subfigure}{0.45\textwidth}
        \centering
        \includegraphics[width=\textwidth]{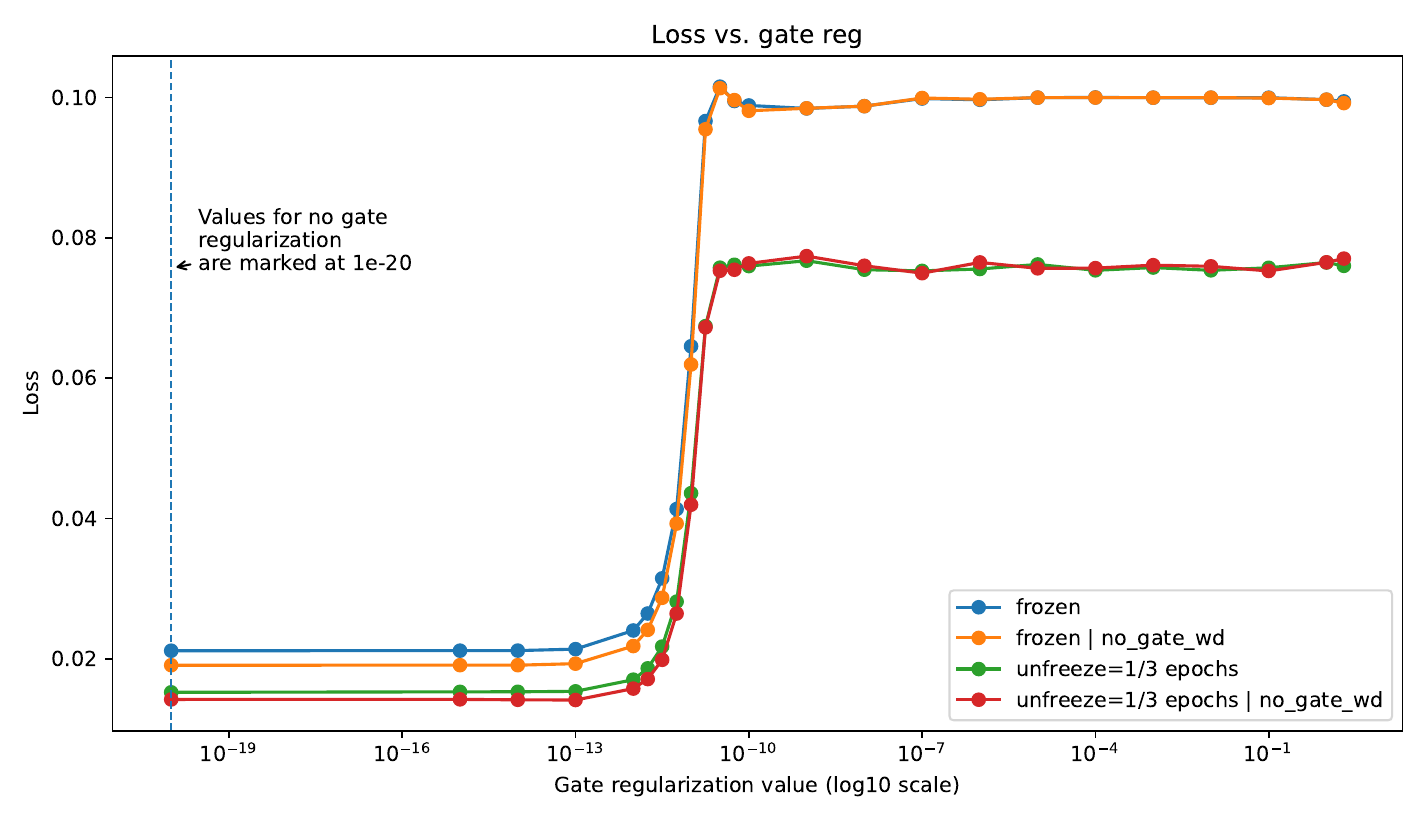}
        \caption{Distillation losses.}
    \end{subfigure}
    \hfill
    \begin{subfigure}{0.45\textwidth}
        \centering
        \includegraphics[width=\textwidth]{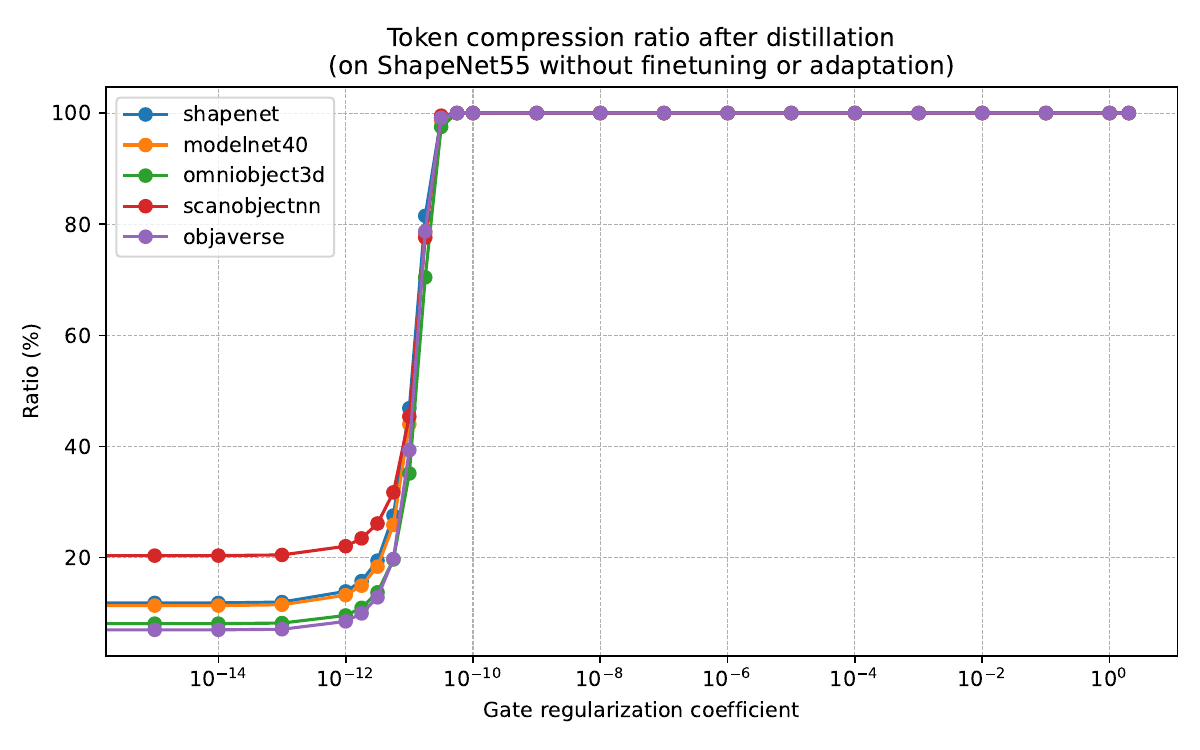}
        \caption{Token compression ratio (frozen backbone).}
    \end{subfigure}
    \caption{\textbf{Effect of gate regularization $\lambda_{\text{gate}}$ on distillation stability.} Distillation loss and token compression ratio curves for frozen and unfrozen backbones across a wide range of $\lambda_{\text{gate}}$. This figure shows the relationship between the distillation loss, token compression rate and gate regularization. (Left) Several distillation configurations are compared, frozen/unfrozen, add/remove weight decay on the gate module. We also provide token fusion ratios for the validation set on Objaverse \cite{deitke2023objaverse} for additional results on an unannotated dataset with more diverse objects. (Right) For $\lambda_{\text{gate}}\!\in\![10^{-12},10^{-10}]$, we observe the token compression ratio is increasing to obtain 100\% of tokens selected by the gate for being merged.}
    \label{fig:v3:distillation-loss-token-compression-rate-vs-gate-reg}
\end{figure}
}

\paragraph{Budget-Aware Inference with Selective Compression.}{
We budgeted the number of tokens that can be processed by the core encoder by, (i) fixing a maximum number of tokens that the encoder can take, (ii) changing the fusion ratio. These two techniques are used at inference-time and using the output probabilities $\pi_i$ of the gate. \Cref{fig:v3:finetuning-metric-vs-token-budget} illustrates the first technique. By limiting the number of tokens, we see different behavior. For the ShapeNet55 dataset, it is at its maximum when all tokens are fused. In contrast, for all others the maximum are when we don't fuse between 50/62.5 to 75/87.5\% of the input tokens (for 64/128 input tokens). \Cref{fig:v3:finetuning-metric-vs-ratio} shows the results for the second technique at inference-time. We observe that the best ratio value is for the value used for distilling and fine-tuning the model. We observe a large downfall for ShapeNet55, larger than for any other dataset. We think that it comes from the fact that it was used during pre-training, distillation and fine-tuning and by fixing a $r$, it does not succeed in inferring the inputs we give. As a hypothesis, we believe that during distillation and fine-tuning, in order to make the ratio more easily changeable at inference time, choosing a random fusion ratio could help models generalize to any value.
\begin{figure}[ht]
    \centering
    \begin{subfigure}{0.45\textwidth}
        \centering
        \includegraphics[width=\linewidth]{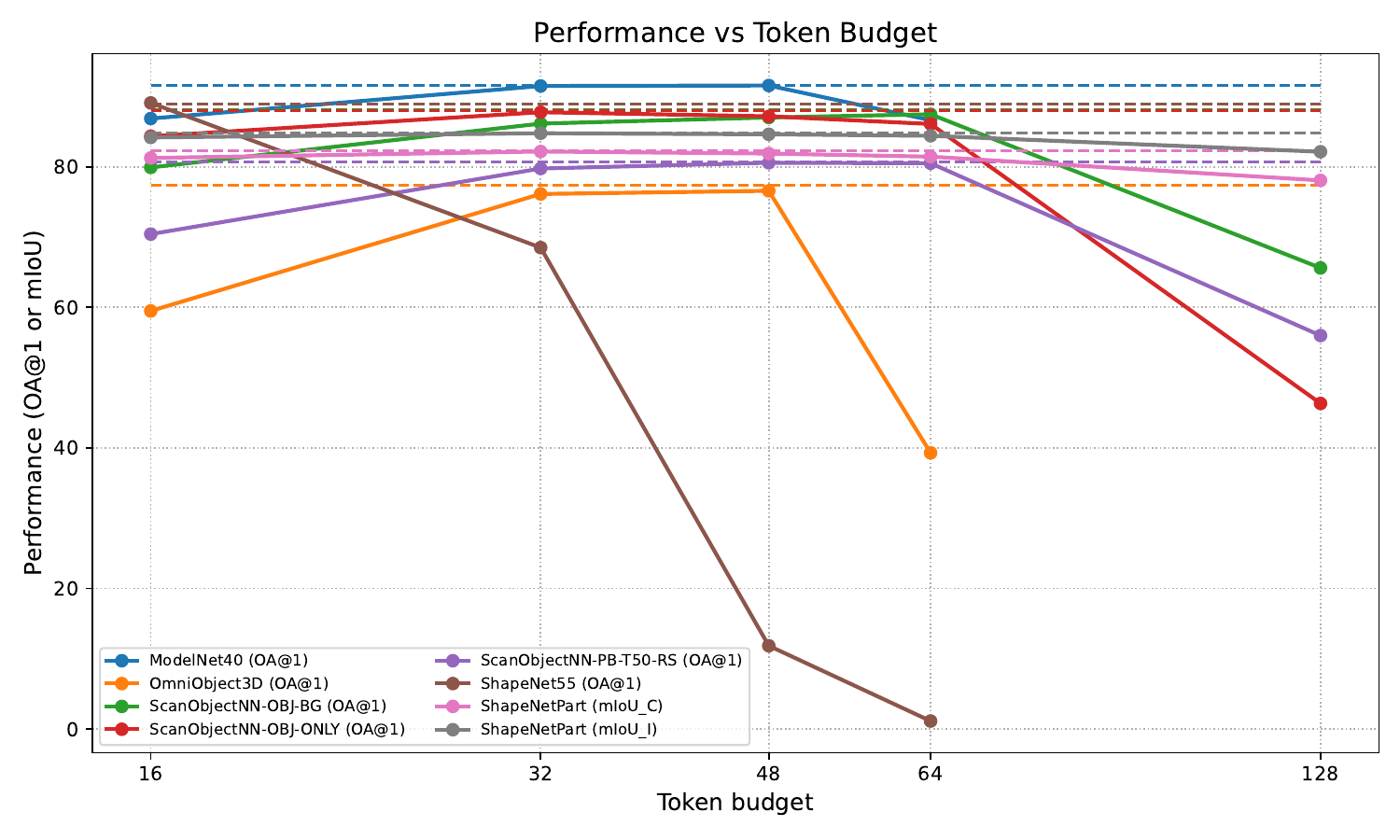}
        \caption{Inference results vs token budget for \paper-Gate.}
        \label{fig:v3:finetuning-metric-vs-token-budget}
    \end{subfigure}
    \hfill
    \begin{subfigure}{0.45\textwidth}
        \centering
        \includegraphics[width=\linewidth]{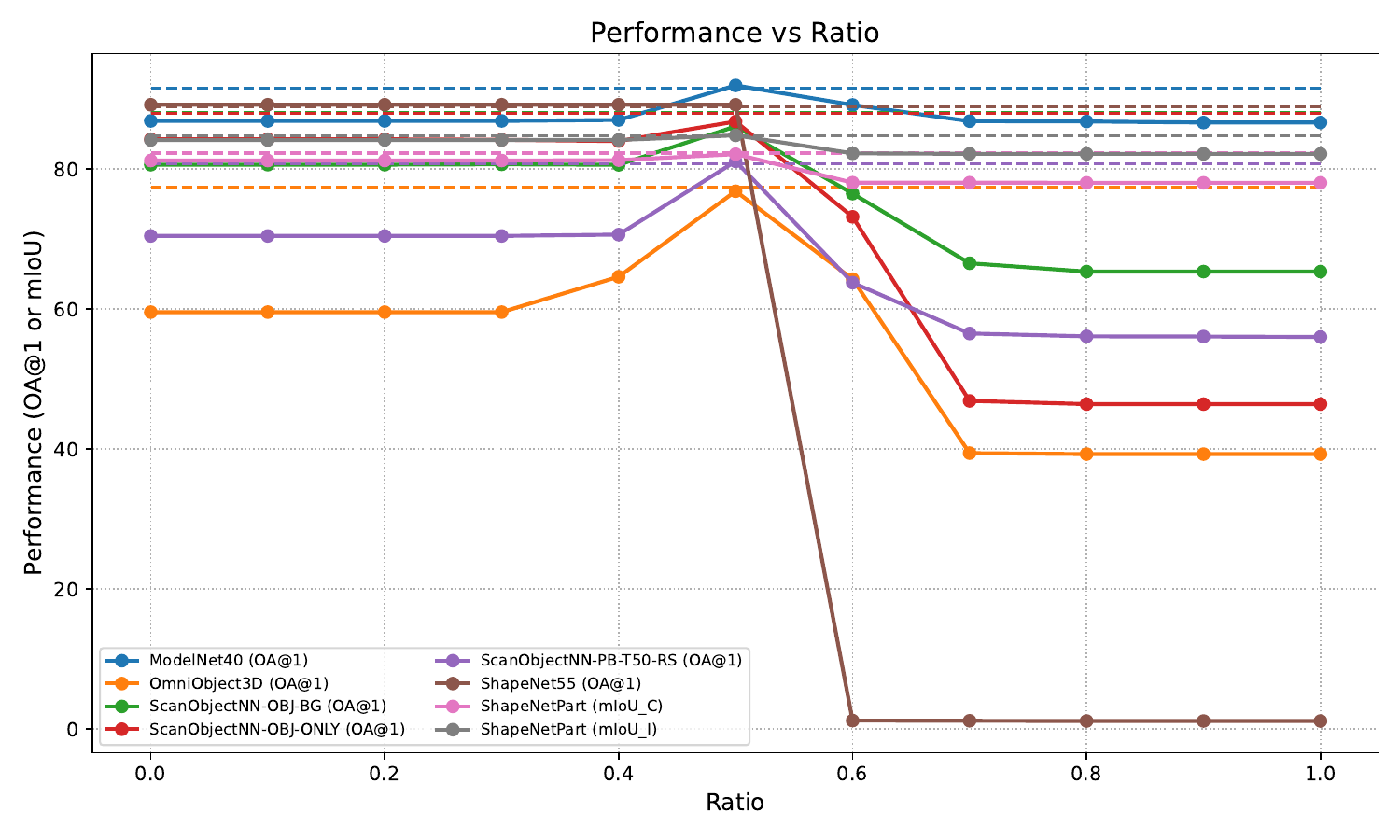}
        \caption{Inference results vs fusion ratio for \paper-Gate.}
        \label{fig:v3:finetuning-metric-vs-ratio}
    \end{subfigure}
    \caption{\textbf{Inference results for \paper-Gate.} We use the best backbone on each dataset with $\lambda_{\text{gate}}=10^{-11}$ and without weight decay. $r$ represents the fusion ratio threshold. All metrics have been computed on the validation set for each dataset.}
    \label{fig:v3:finetuning-metric-vs-token-budget-ratio}
\end{figure}
\Cref{tab:inference-v3-token-budget-ratio} contains all plotted values.
\input{tables/inference-v3-budgeted-tokens-and-ratio}
}

\paragraph{Empirical computational cost.}{
We provide GPU and CPU costs in \cref{tab:empirical-computational-cost-complete}. The used GPU is a NVIDIA RTX~A3000 6GB Laptop GPU and the CPU is Intel Core i9-11950H @ 2.60GHz. Reported values take into account tokenization and encoding steps. 
\input{tables/computational-cost-empiric}
}

\subsection{Visualizations}
\begin{figure*}[htbp]
  \centering
  \includegraphics[width=\linewidth]{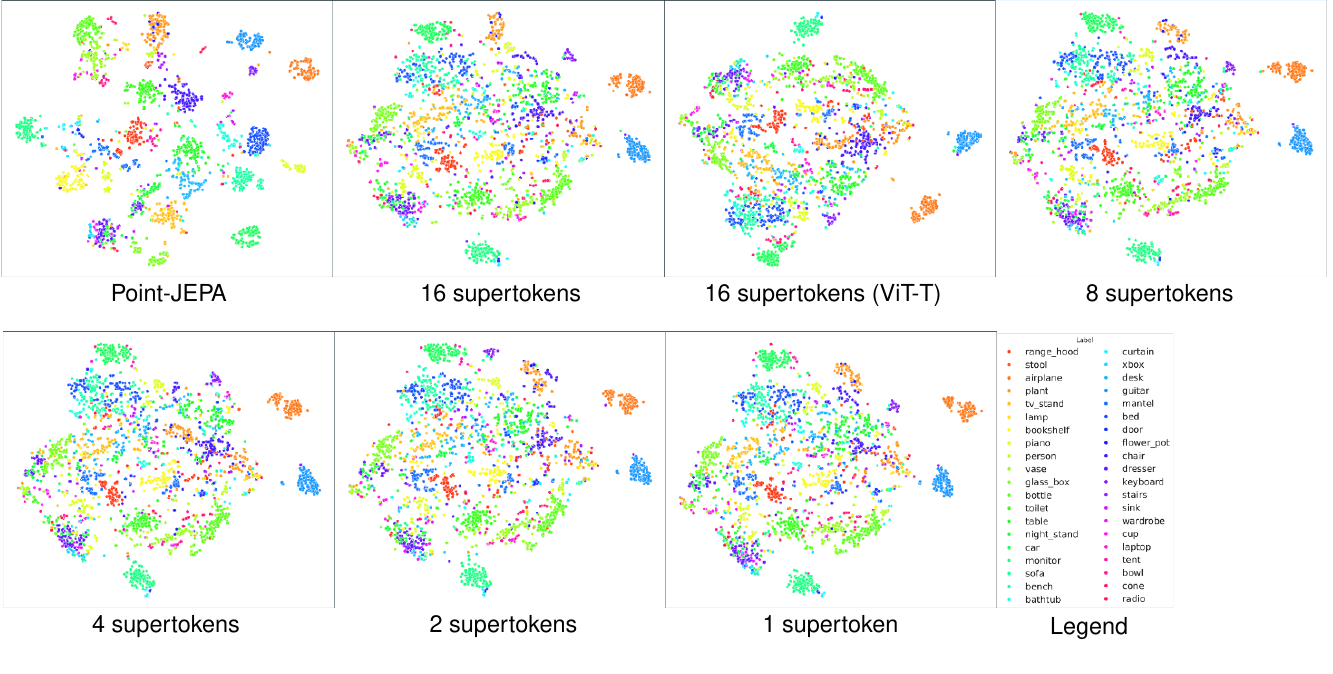}
  \caption{\textbf{Embedding visualization for \paper.} t-SNE \cite{JMLR:v9:vandermaaten08a} projections of the output representation of frozen backbone at distillation ending on ModelNet40 \cite{wu20153d} validation set. Point-JEPA outputs are from encoder and CAU for others. If not specified, the Transformer architecture used for student during distillation is ViT-S. We recommend zooming in to see details.}
  \label{fig:appendix:embeddings-supertokens}
\end{figure*}

\paragraph{Embedding visualization.}
In \cref{fig:appendix:embeddings-supertokens}, we observe that at the encoder output clear clusters appear, in opposition to outputs of the backbone with supertokens where the clusters are more mixed, which explains the degradation in fine-tuning results. This is due to our chosen architecture where in CAU, to recover tokens we upsample from encoder outputs (processed supertokens) and CAM from DSO and we add the output of the tokenizer to add some original information as done in 3DLST \cite{lu20243d}. Nonetheless, it proves that our method succeed to extract basis vectors of the representation space. Between embeddings of different compression levels, we can't say that one is better than the other. With ViT-T, we see more close clusters and the separation is less obvious compared to its left neighbor. We think that comes from both compressing the input and the model embedding size. The capacity of encoding data is less important but helps to speed up inferences.

\begin{figure}
    \centering
    \includegraphics[width=0.7\linewidth]{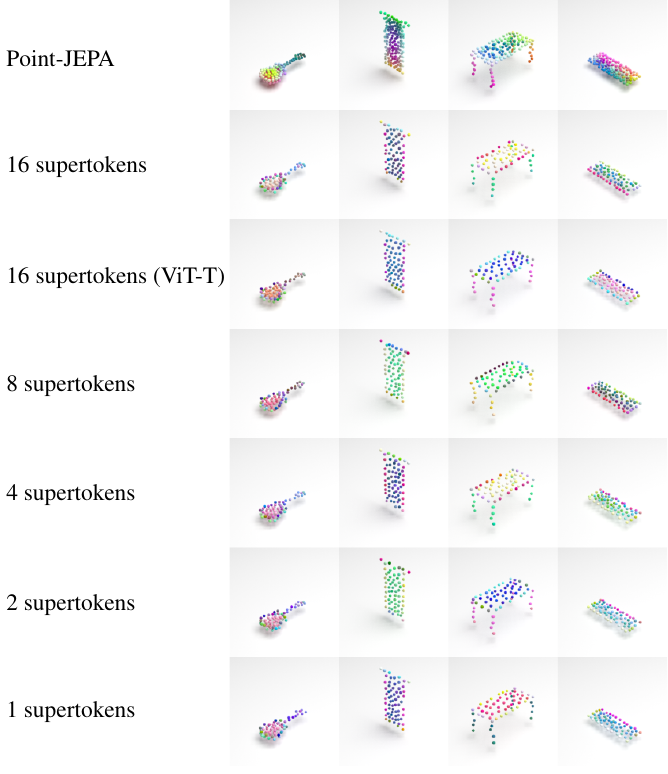}
    \caption{\textbf{PCA embedding visualization for \paper which changes feature quality by focusing on object edges for few examples.} PCA projections of the output representation of frozen backbone at distillation ending on ModelNet40 \cite{wu20153d} validation set in RGB space. If not specified, the Transformer architecture used for student during distillation is ViT-S. We recommend zooming in to see details. The associated labels for each object are ordered as follow: guitar, curtain, table and keyboard.}
    \label{fig:appendices:pca-vis-edges}
\end{figure}

\begin{figure*}
    \centering
    \includegraphics[width=\linewidth]{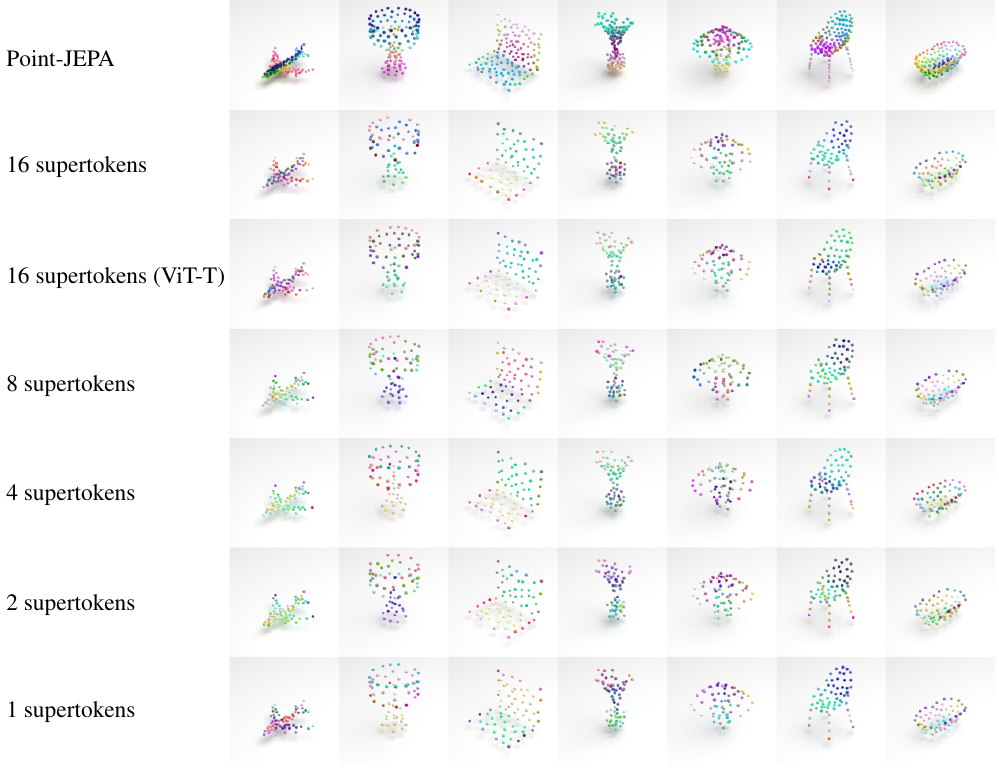}
    \caption{\textbf{PCA embedding visualization for \paper for keeping feature quality on some selected examples.} PCA projections of the output representation of frozen backbone at distillation ending on ModelNet40 \cite{wu20153d} validation set in RGB space. If not specified, the Transformer architecture used for student during distillation is ViT-S. We recommend zooming in to see details. The associated labels for each object are ordered as follow: airplane, lamp, laptop, plant, cone, chair and bathtub.}
    \label{fig:appendices:pca-vis-good}
\end{figure*}

\begin{figure*}
    \centering
    \includegraphics[width=\linewidth]{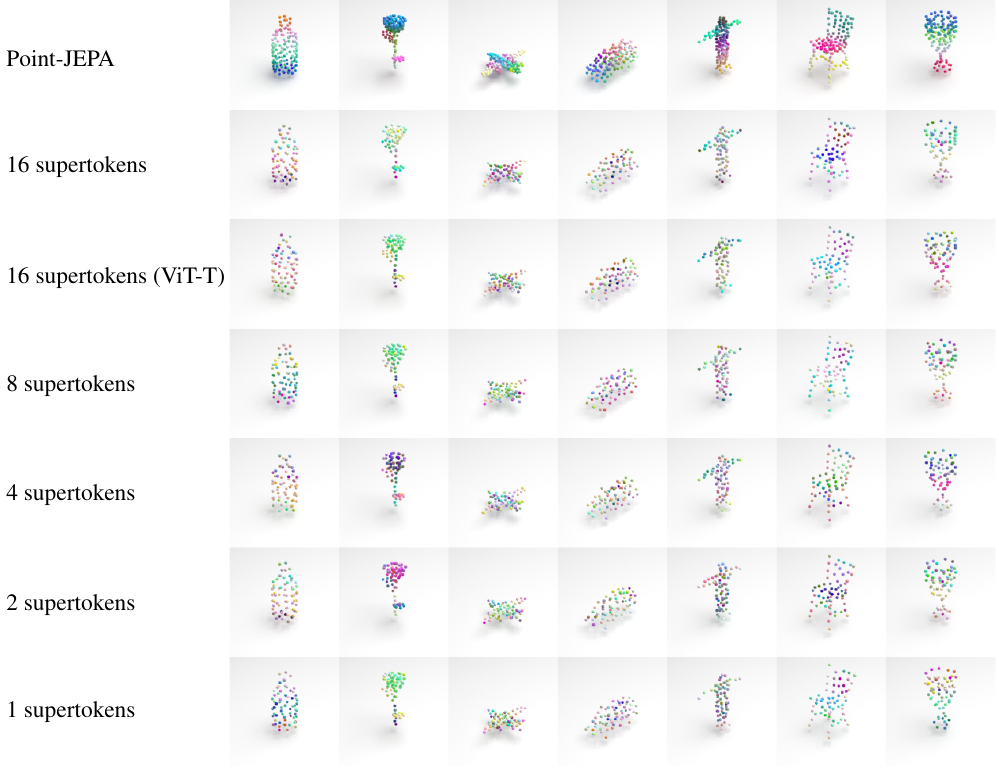}
    \caption{\textbf{PCA embedding visualization for \paper for lower feature quality examples.} PCA projections of the output representation of frozen backbone at distillation ending on ModelNet40 \cite{wu20153d} validation set in RGB space. If not specified, the Transformer architecture used for student during distillation is ViT-S. We recommend zooming in to see details. The associated labels for each object are ordered as follow: bottle, plant, airplane, car, person, chair and cup.}
    \label{fig:appendices:pca-vis-bad}
\end{figure*}

\paragraph{PCA embedding visualization.}{
In \cref{fig:appendices:pca-vis-edges}, we observe that the supertokens are more inclined to analyze edges rather than semantics by analyzing PCA embeddings per object. We think that the DSO/CAU blocks try to focus the encoder more on discriminative parts for simplifying analysis. This is why for complex objects, the distilled encoder succeeds in classifying them. Additionally, with \cref{fig:appendices:pca-vis-good}, we also share some examples where the semantics seem equal or even better. We also see that the edges are more present compared to the baseline. In \cref{fig:appendices:pca-vis-bad}, some semantics are degraded by being less precise like flower, plane or glass, or completely lost with the bottle or car when the number of supertokens is too low. This demonstrates the need to use enough supertokens not only to improve performances, but for keeping semantic information in the latent space.
}

\begin{figure*}[ht!]
\centering
\resizebox{\linewidth}{!}{
\begin{tabular}{l|ccc}
\toprule
Model & Example 1 & Example 2 & Example 3 \\
\midrule
16 supertokens & \includegraphics[width=.3\linewidth]{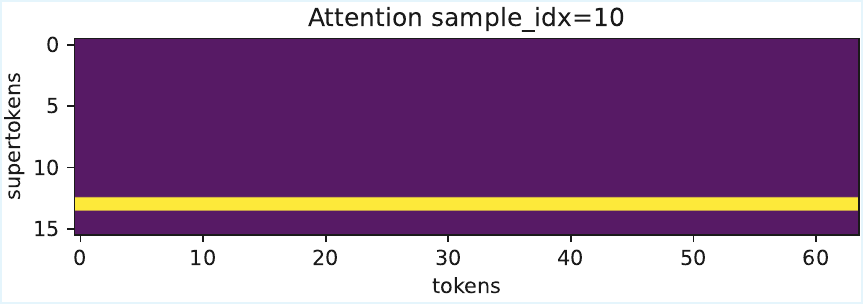} & \includegraphics[width=.3\linewidth]{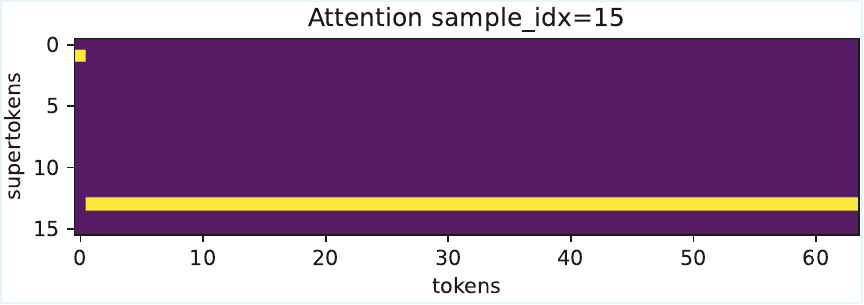} & \includegraphics[width=.3\linewidth]{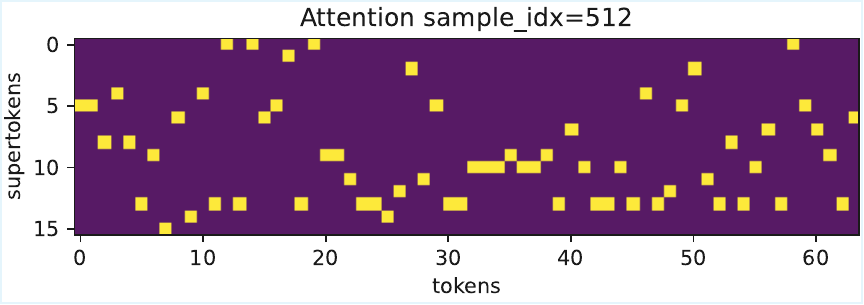} \\
16 supertokens (ViT-T) & \includegraphics[width=.3\linewidth]{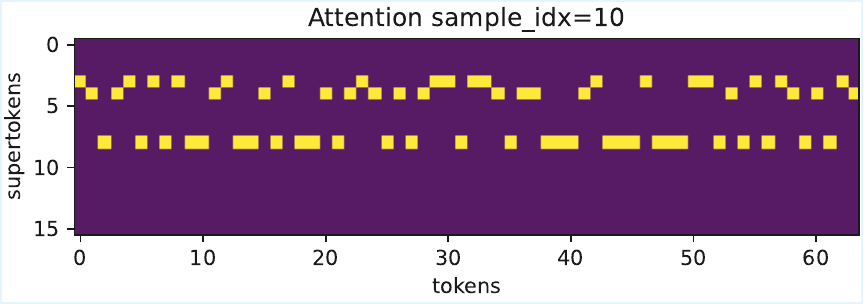} & \includegraphics[width=.3\linewidth]{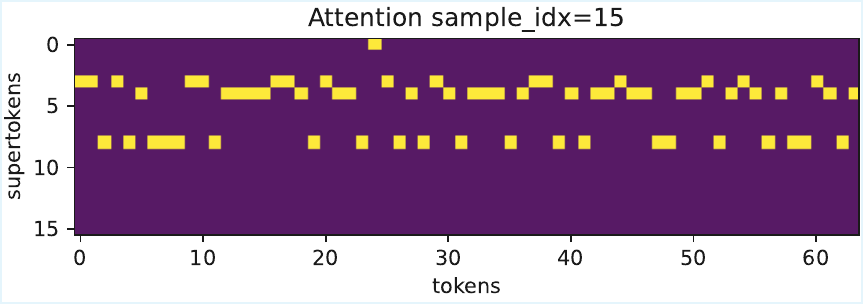} & \includegraphics[width=.3\linewidth]{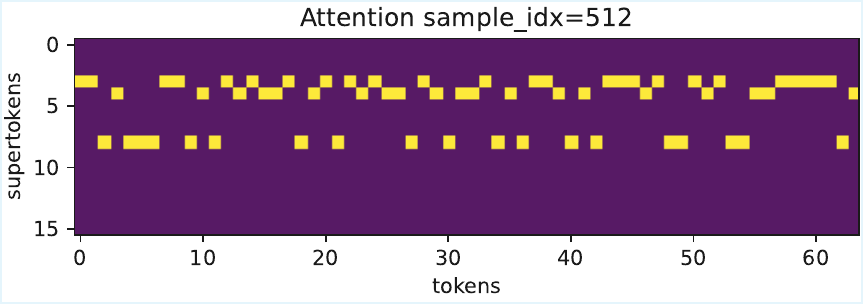} \\
8 supertokens & \includegraphics[width=.3\linewidth]{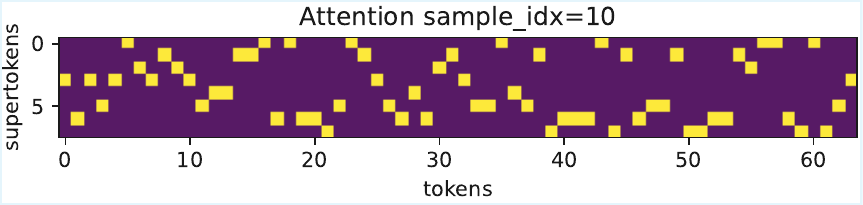} & \includegraphics[width=.3\linewidth]{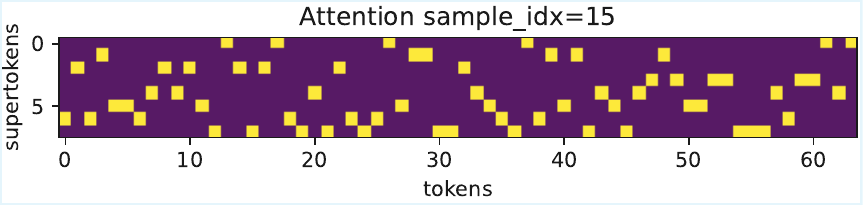} & \includegraphics[width=.3\linewidth]{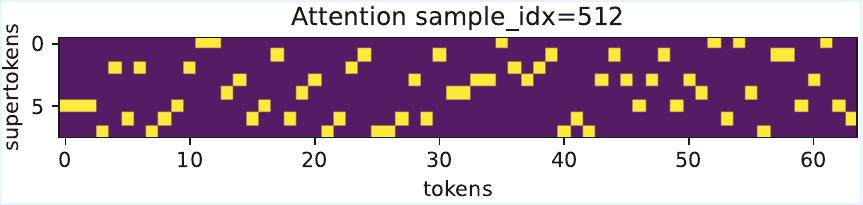} \\
4 supertokens & \includegraphics[width=.3\linewidth]{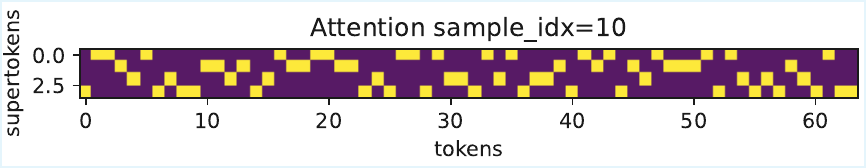} & \includegraphics[width=.3\linewidth]{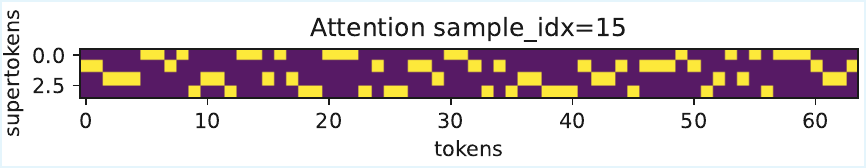} & \includegraphics[width=.3\linewidth]{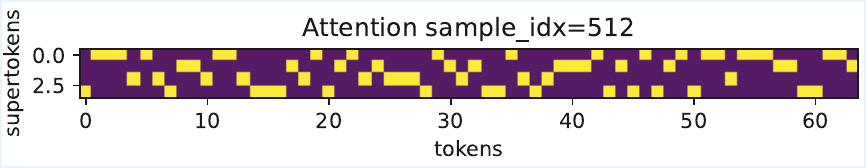} \\
2 supertokens & \includegraphics[width=.3\linewidth]{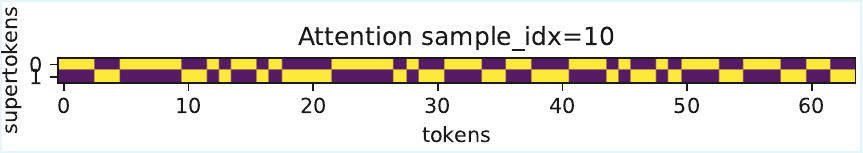} & \includegraphics[width=.3\linewidth]{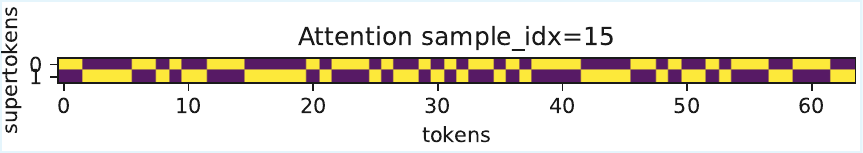} & \includegraphics[width=.3\linewidth]{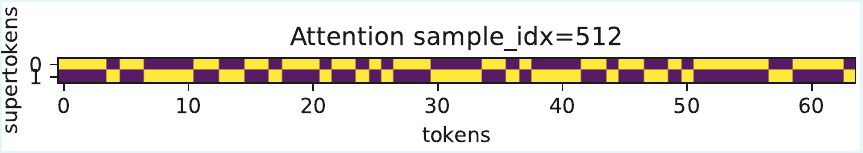} \\
1 supertoken & \includegraphics[width=.3\linewidth]{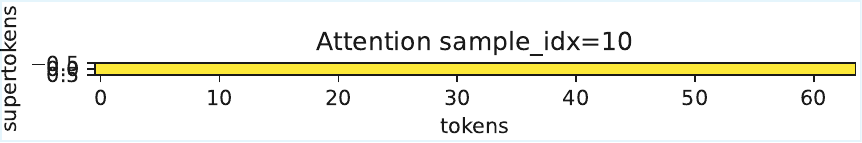} & \includegraphics[width=.3\linewidth]{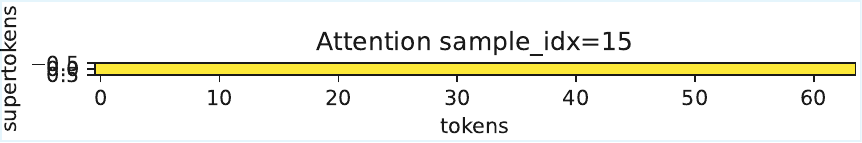} & \includegraphics[width=.3\linewidth]{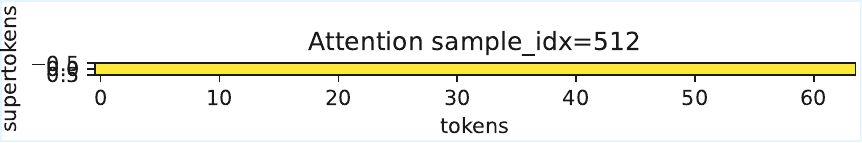} \\
\end{tabular}
}
\caption{\textbf{Cross-Attention Map (CAM) comparison.} Unless specified, ViT-S backbone is used for student branch during distillation. The first example represents a person, second a bottle and third a toilet object. The examples come from validation set of ModelNet40 \cite{wu20153d}. The used backbones come from the end of distillation stage.}
\label{fig:appendix:cams}
\end{figure*}

\paragraph{CAM visualization.}{
\Cref{fig:appendix:cams} shows the chosen supertokens for three random examples. Firstly, we observe that the number of chosen supertokens is not constant and it depends on the considered example. Some need more supertokens for representing the object and some only one. Also, we remark that the fewer supertokens we defined at the beginning, the more supertokens are used. For the first column, we observe that for 16 supertokens, one is needed but for 8, we use all. We think that when we let the model with larger number of supertokens, they are more specialized than smaller, which is expected behavior. Finally, reducing the embedding size 384 of ViT-S to 192 of ViT-T degrades the quality of supertokens. This result makes the link with the observation on the global embedding in \cref{fig:appendix:embeddings-supertokens}.
}

\paragraph{Supertokens visualization.}{
\Cref{fig:appendix:supertokens-vis:spatial} shows selected tokens on few objects. In this figure, we observe that different supertokens are used for each category of objects, even though some are shared between several classes. When divergent forms appear within the same category, other supertokens tend to be used. This is consistent with the fact that merging is performed after tokenization, without access to the context provided by neighboring tokens, which limits the consideration of more global semantic information. Thus, merging relies mainly on low-level cues—such as local patterns or geometric structures (e.g., the orientation or curvature of a shape). This explains why some supertokens are specialized for very local features, while others, which are more general, appear in several classes. In other words, the distribution of supertokens reflects both the internal geometric diversity within categories and the degree of invariance captured by the merging process.

\begin{figure*}[ht!]
\centering
\resizebox{\linewidth}{!}{
\begin{tabular}{cccccc}
\includegraphics[width=.15\linewidth]{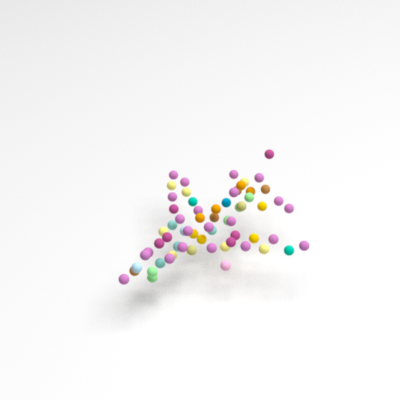} & \includegraphics[width=.15\linewidth]{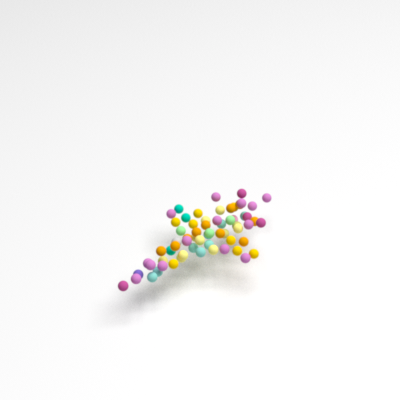} & \includegraphics[width=.15\linewidth]{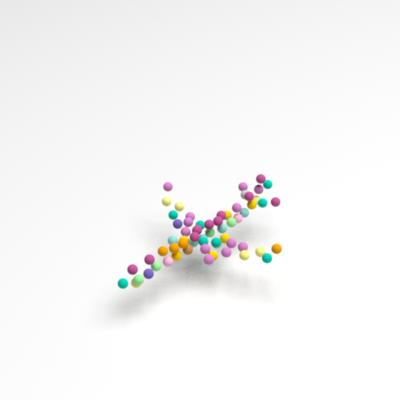} & \includegraphics[width=.15\linewidth]{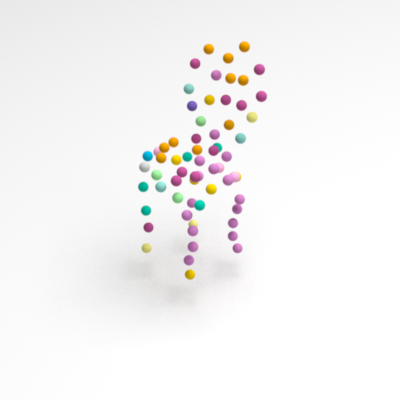} & \includegraphics[width=.15\linewidth]{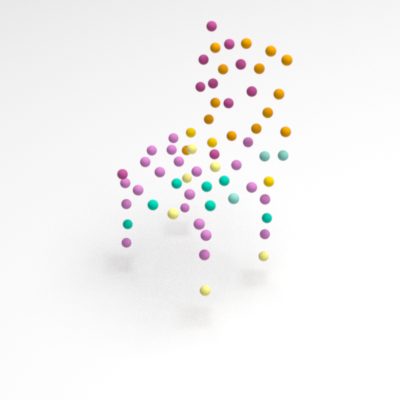} & \includegraphics[width=.15\linewidth]{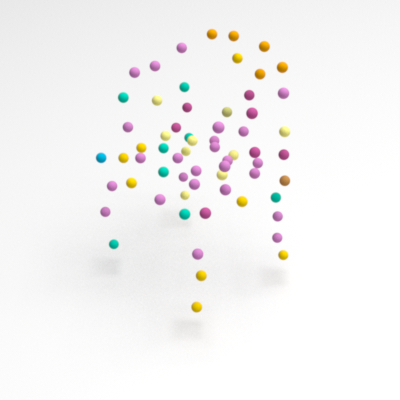} \\
\includegraphics[width=.15\linewidth]{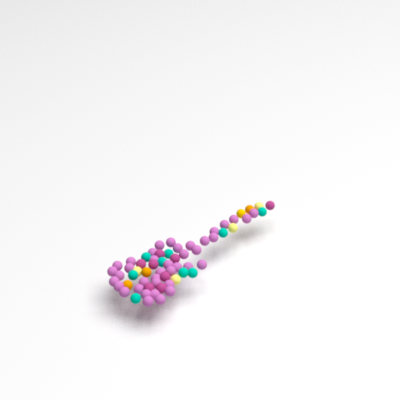} & \includegraphics[width=.15\linewidth]{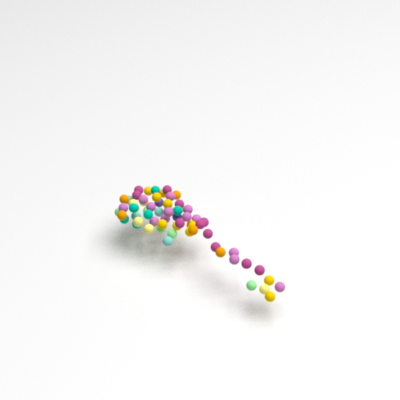} & \includegraphics[width=.15\linewidth]{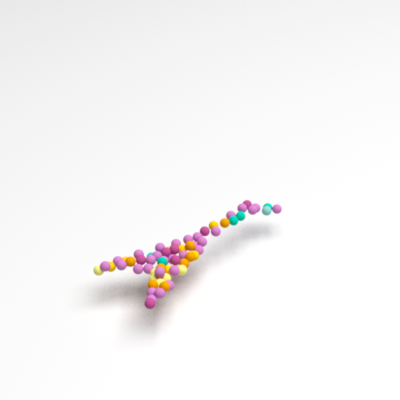} & \includegraphics[width=.15\linewidth]{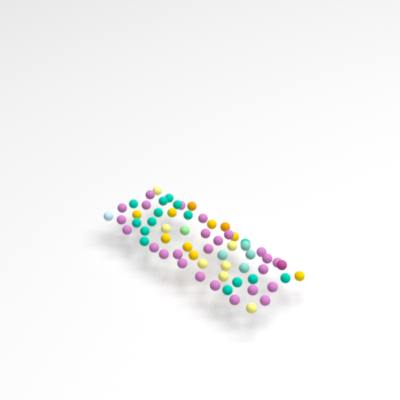} & \includegraphics[width=.15\linewidth]{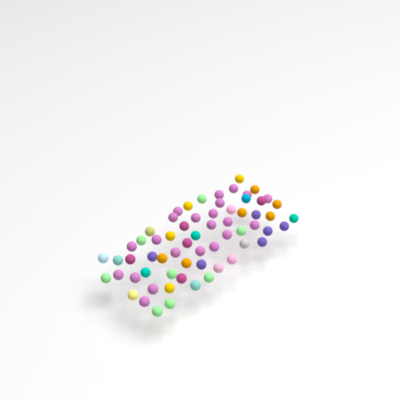} & \includegraphics[width=.15\linewidth]{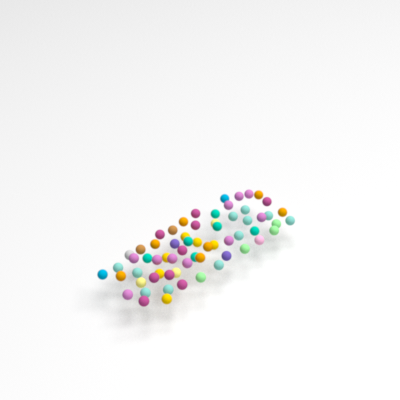} \\
\includegraphics[width=.15\linewidth]{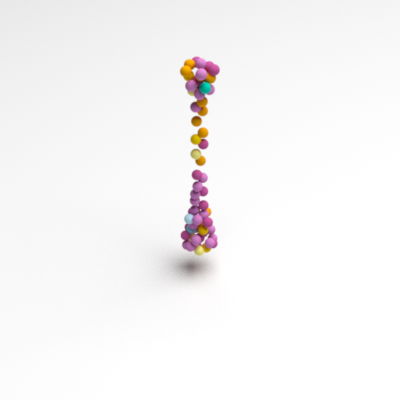} & \includegraphics[width=.15\linewidth]{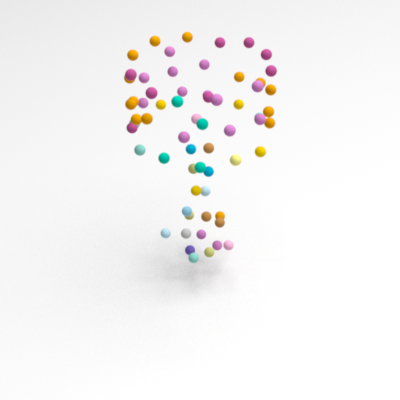} & \includegraphics[width=.15\linewidth]{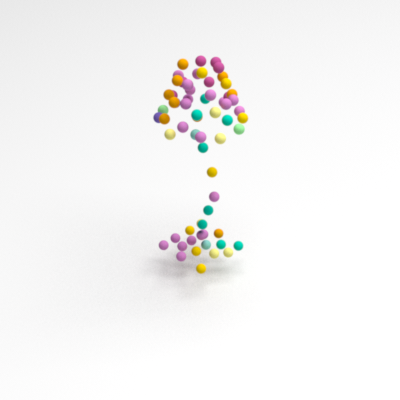} & \includegraphics[width=.15\linewidth]{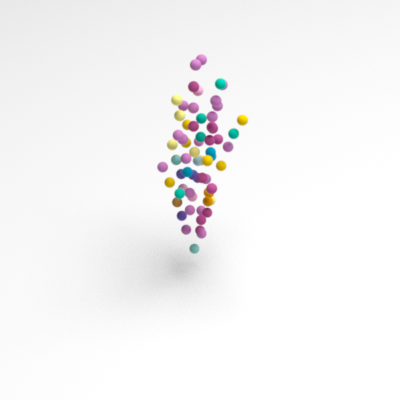} & \includegraphics[width=.15\linewidth]{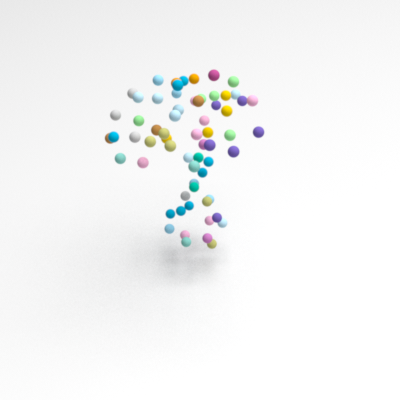} &
\includegraphics[width=.15\linewidth]{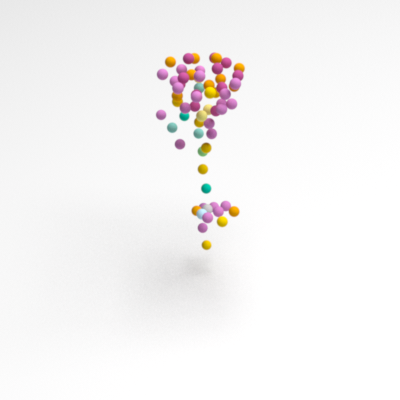} \\ \includegraphics[width=.15\linewidth]{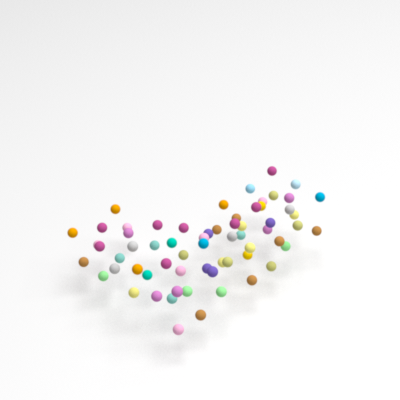} & \includegraphics[width=.15\linewidth]{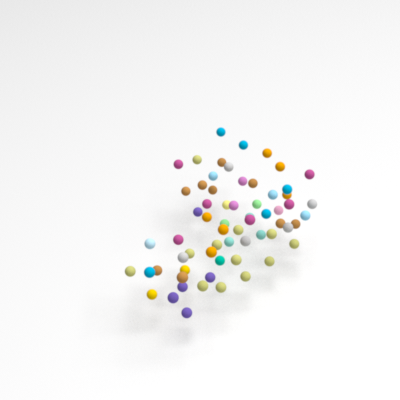} & \includegraphics[width=.15\linewidth]{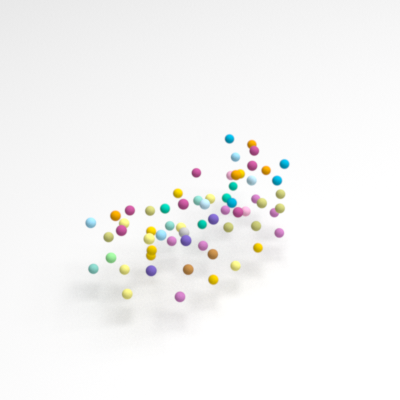} & \includegraphics[width=.15\linewidth]{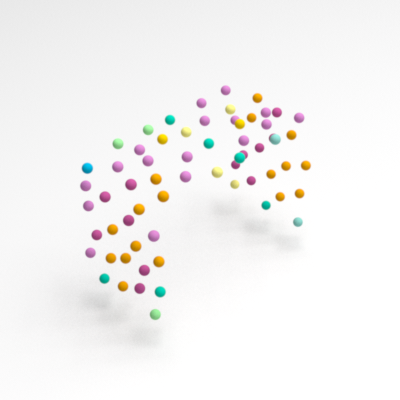} & \includegraphics[width=.15\linewidth]{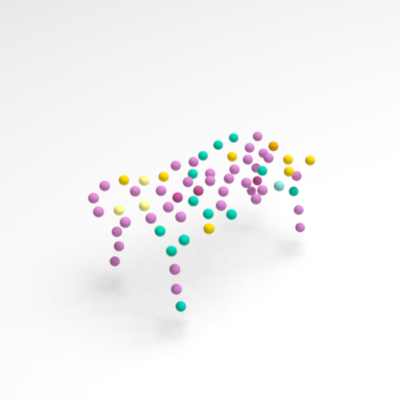} & \includegraphics[width=.15\linewidth]{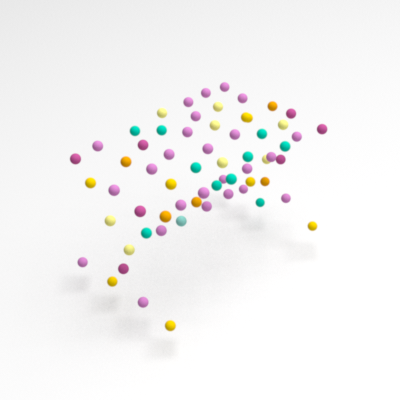} 
\end{tabular}
}
\caption{\textbf{Qualitative supertokens selection view in 3D space.} We select examples from following categories: airplane, chair, guitar, keybord, lamp, plant, sofa and, table. Colors are unique and are associated to supertokens throughout all selected examples. The examples come from validation set of ModelNet40 \cite{wu20153d}. Classifier is the 16 supertokens best checkpoint.}
\label{fig:appendix:supertokens-vis:spatial}
\end{figure*}
\Cref{fig:appendix:supertokens-vis:semantic-before-projections,fig:appendix:supertokens-vis:semantic-after-projections} respectively displays the embedding space of tokenized inputs and the supertokens set before and after being projected by $\mathbf{W_Q}$ and $\mathbf{W_K}$. In \cref{fig:appendix:supertokens-vis:semantic-before-projections}, we observe that the supertokens all appear in the same location in the token and supertoken space before projection. This result is expected, as the similarity used to construct the CAM is based on a linear projection applied identically to tokens and supertokens via the two weight matrices. \Cref{fig:appendix:supertokens-vis:semantic-after-projections} is therefore more informative: it represents the space in which the fusion is actually performed. In this space, supertokens are generally close to the tokens assigned to them. However, some are not. We believe that this discrepancy stems from the effects of t-SNE projection: the dimensions relevant for distinguishing certain tokens may not be correctly reproduced in 2D, which can mask the actual relationships that exist in high-dimensional space.
\begin{figure*}[ht!]
\centering
\resizebox{\linewidth}{!}{
\begin{tabular}{cc}
\includegraphics[width=.4\linewidth]{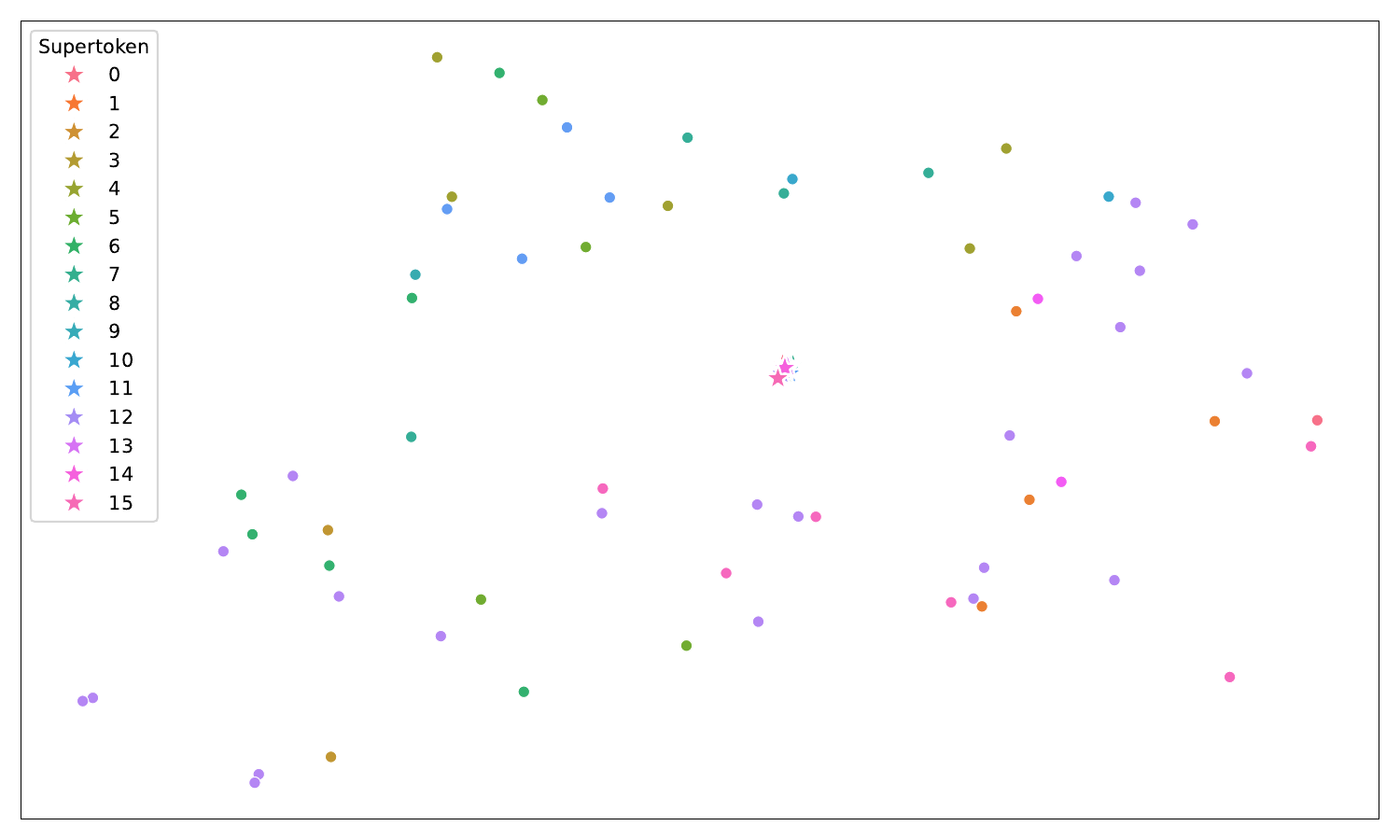} & \includegraphics[width=.4\linewidth]{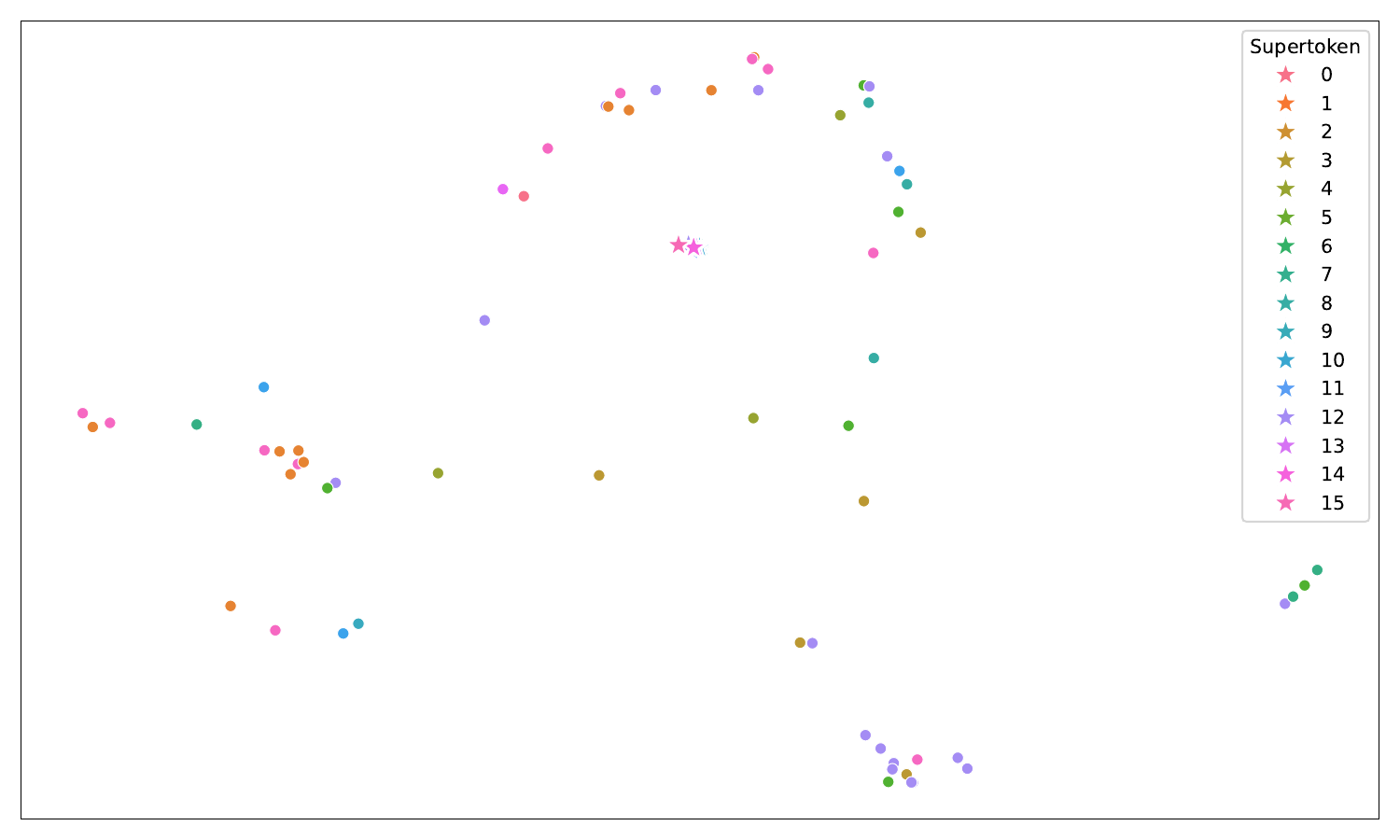} \\
\includegraphics[width=.4\linewidth]{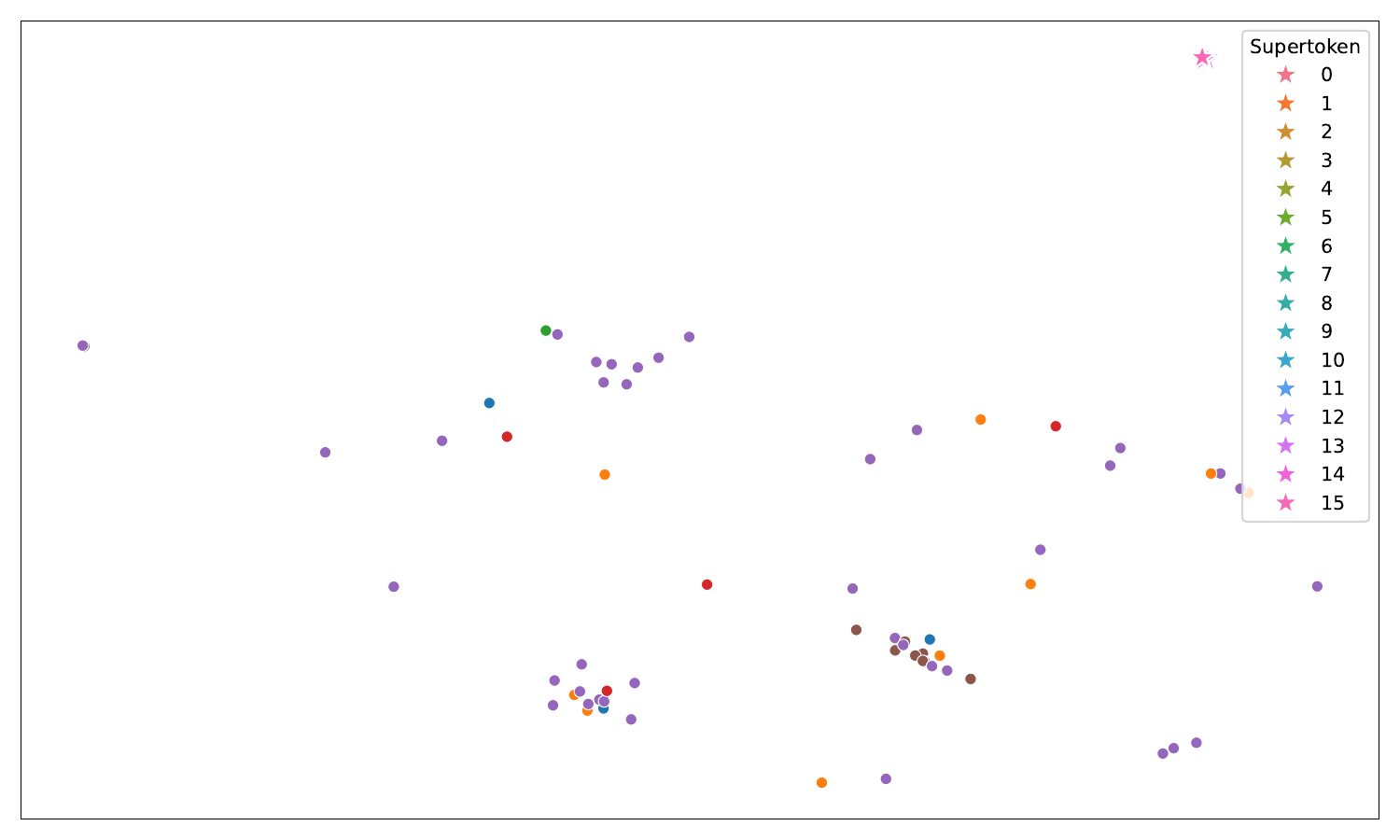} & \includegraphics[width=.4\linewidth]{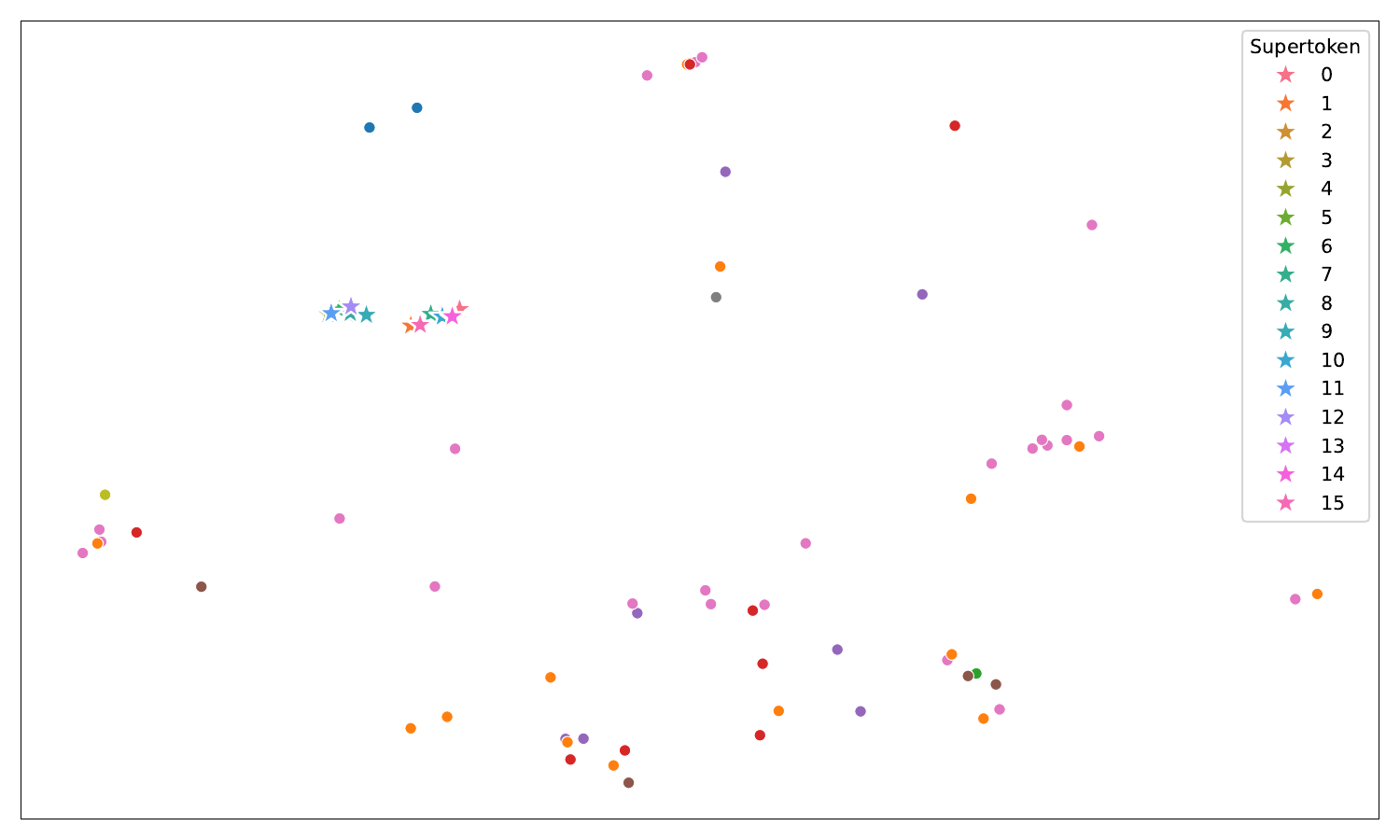} \\
\includegraphics[width=.4\linewidth]{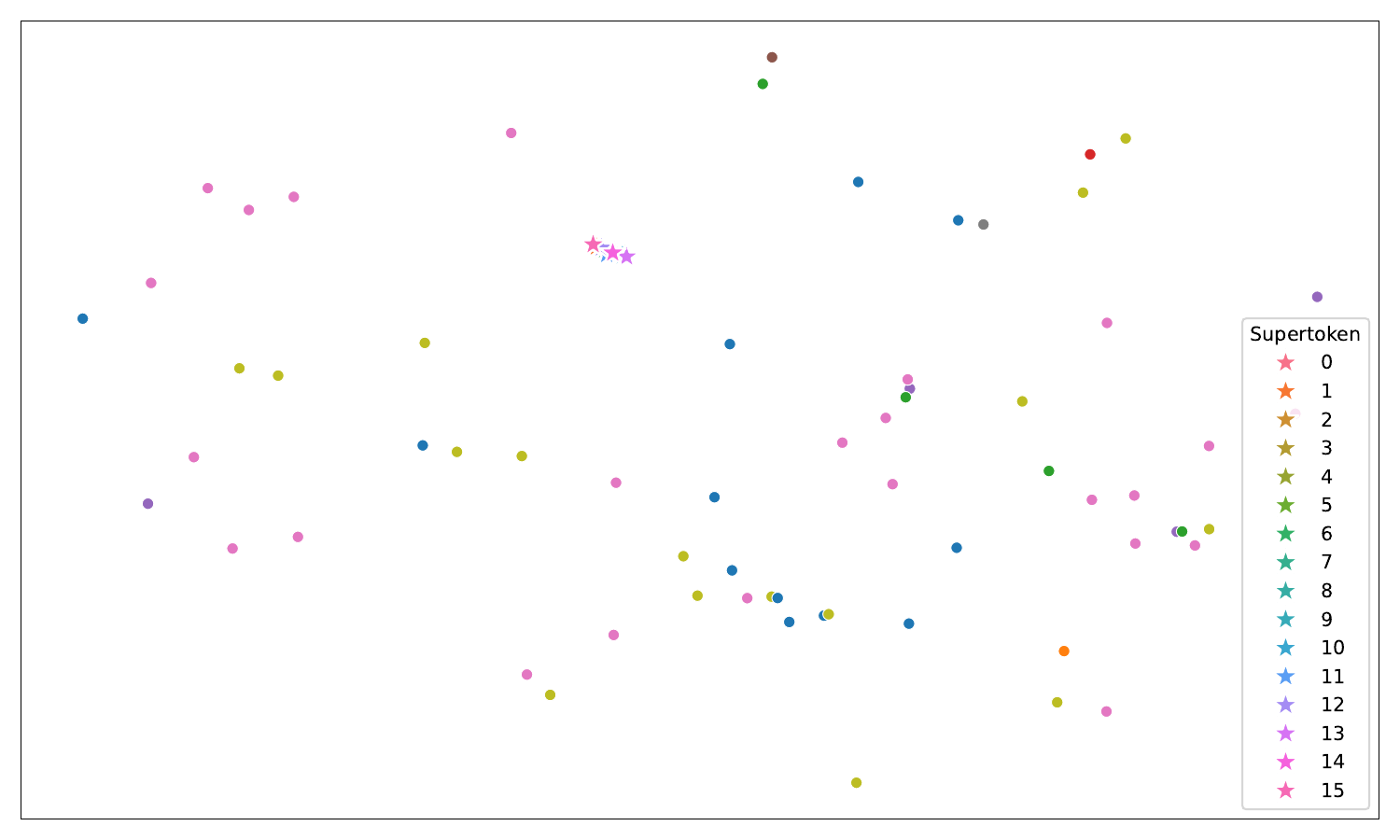} & \includegraphics[width=.4\linewidth]{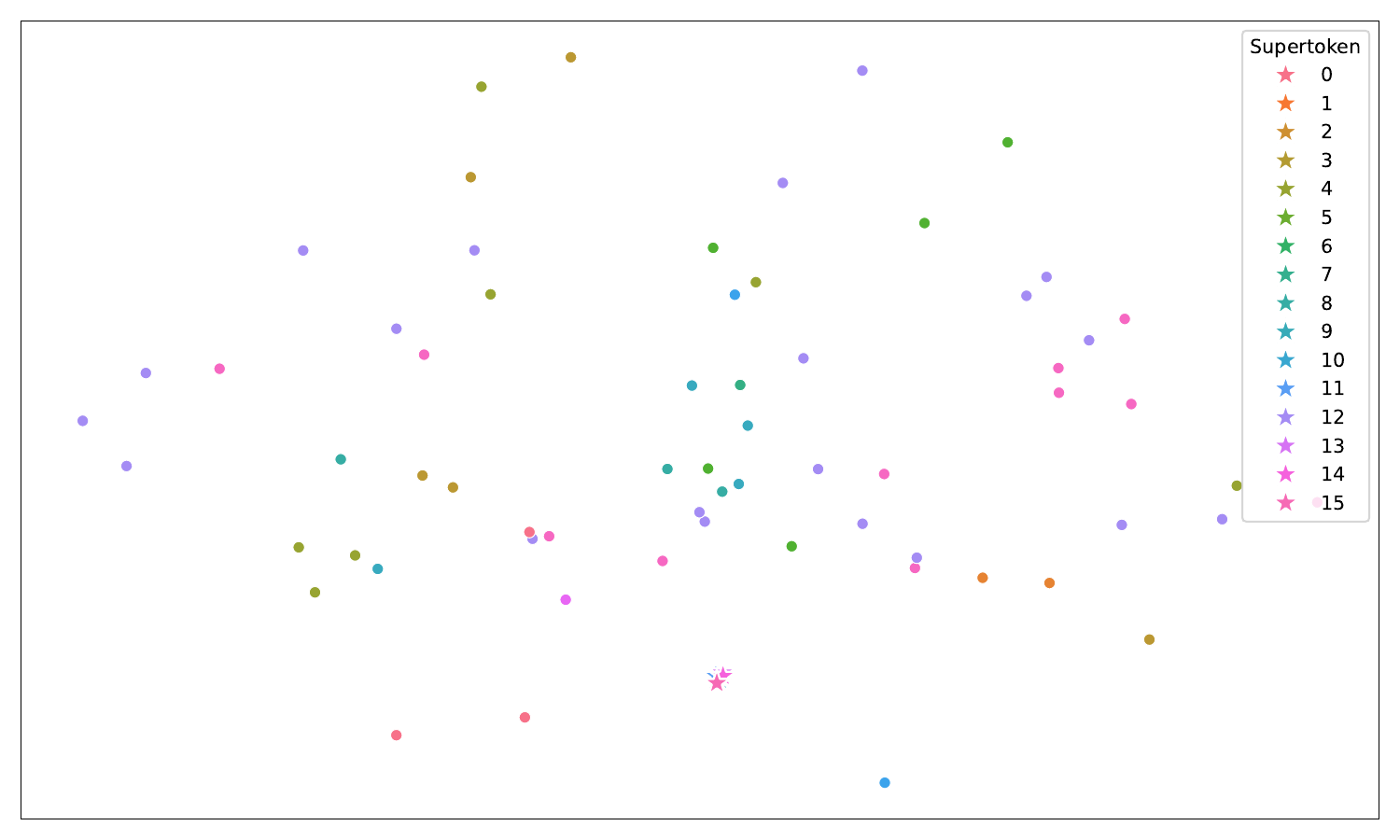} \\
\includegraphics[width=.4\linewidth]{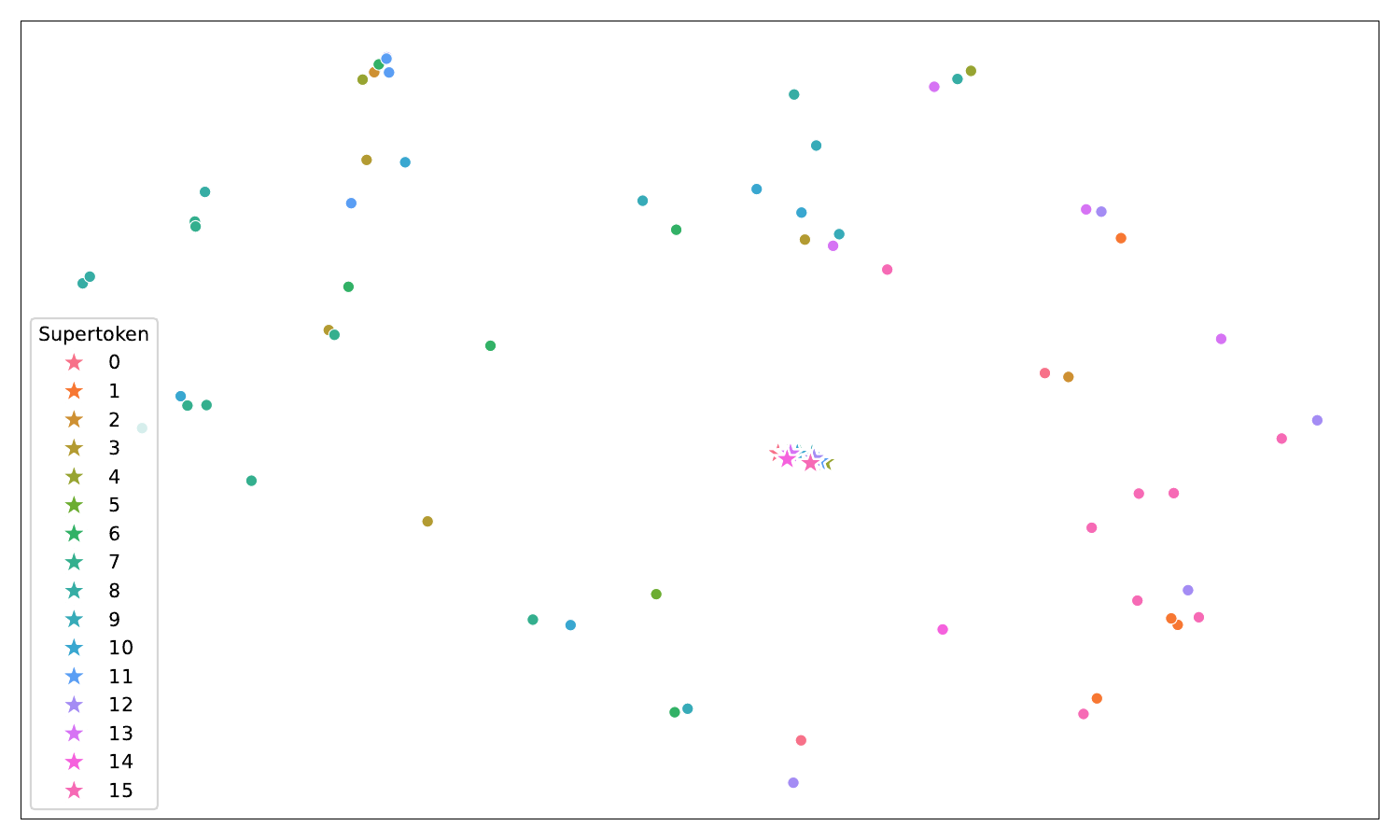} & \includegraphics[width=.4\linewidth]{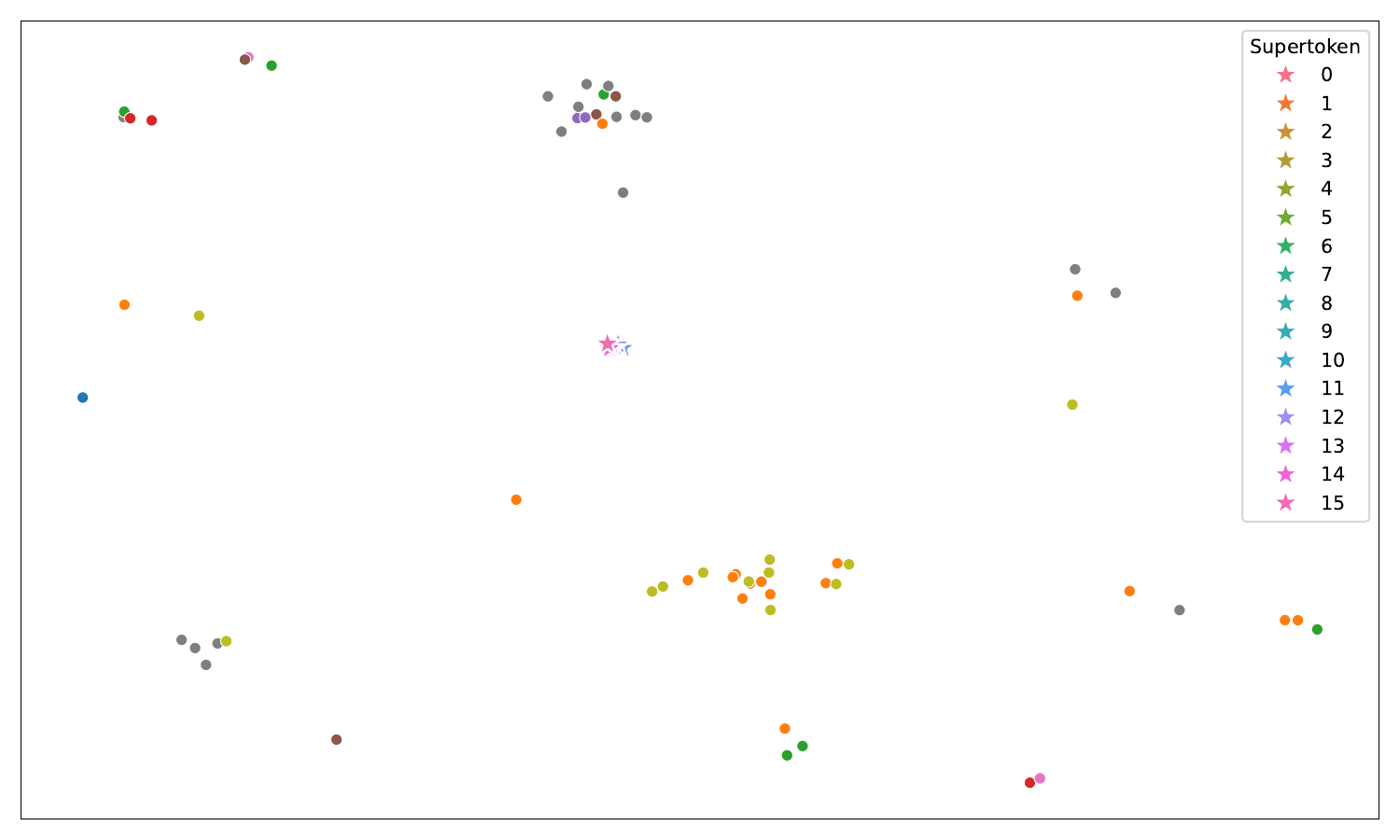}
\end{tabular}
}
\caption{\textbf{Qualitative supertokens selection view in token-supertoken space before projecting.} We select examples from following categories: airplane, chair, guitar, keybord, lamp, plant, sofa and, table. Colors are unique and are associated to supertokens throughout all selected examples. The examples come from validation set of ModelNet40 \cite{wu20153d} and are the same as \cref{fig:appendix:supertokens-vis:spatial}. Classifier is the 16 supertokens best checkpoint.}
\label{fig:appendix:supertokens-vis:semantic-before-projections}
\end{figure*}
\begin{figure*}[ht!]
\centering
\resizebox{\linewidth}{!}{
\begin{tabular}{cc}
\includegraphics[width=.4\linewidth]{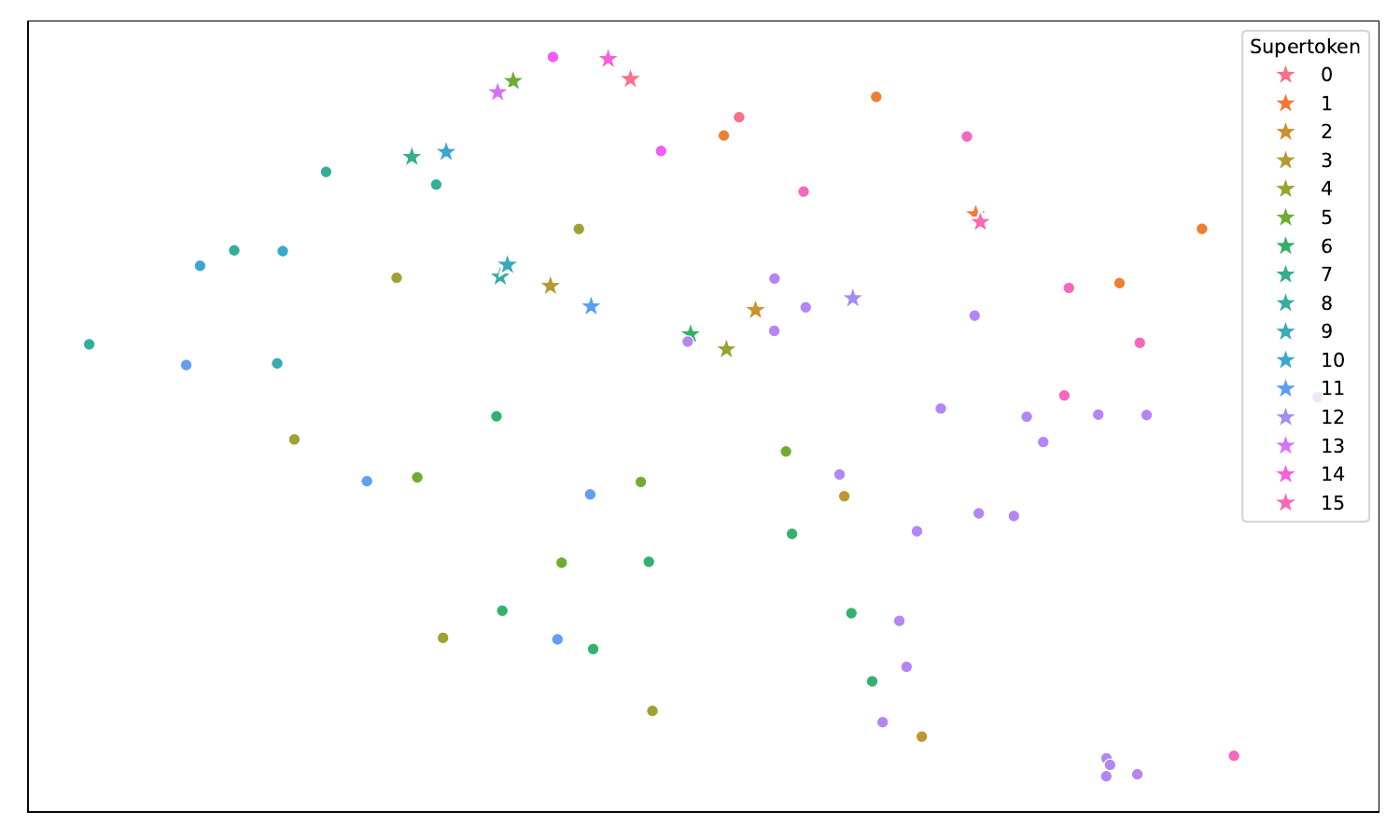} & \includegraphics[width=.4\linewidth]{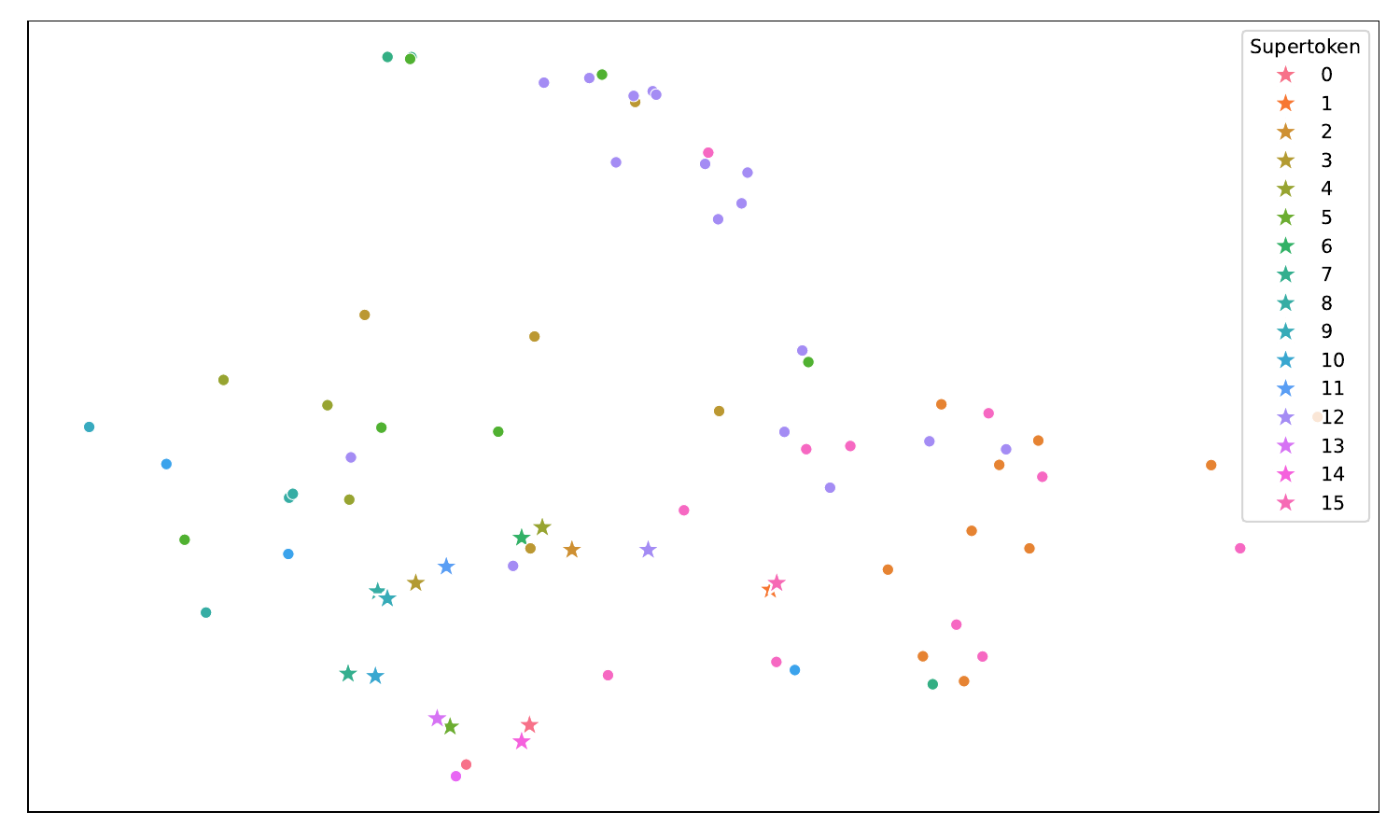} \\
\includegraphics[width=.4\linewidth]{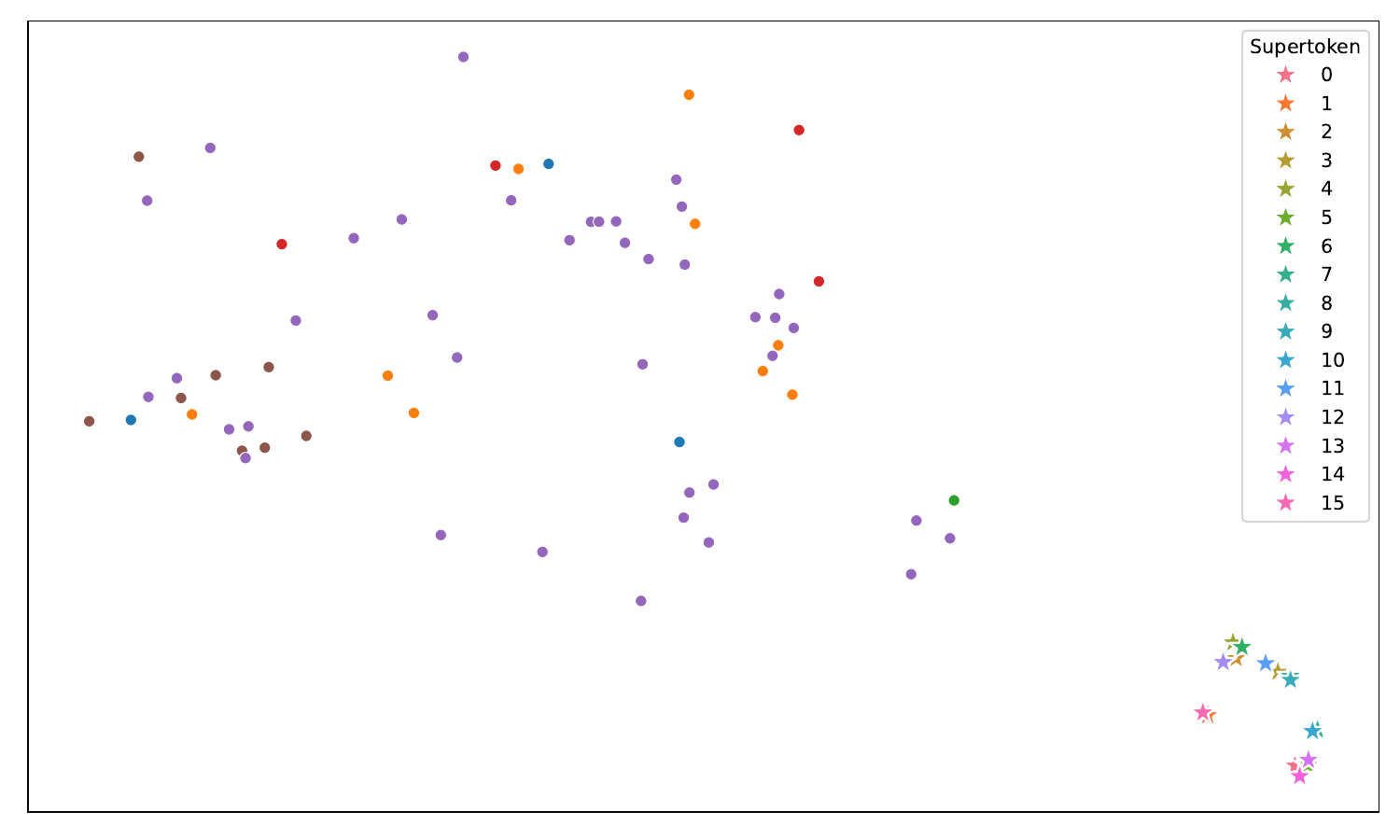} & \includegraphics[width=.4\linewidth]{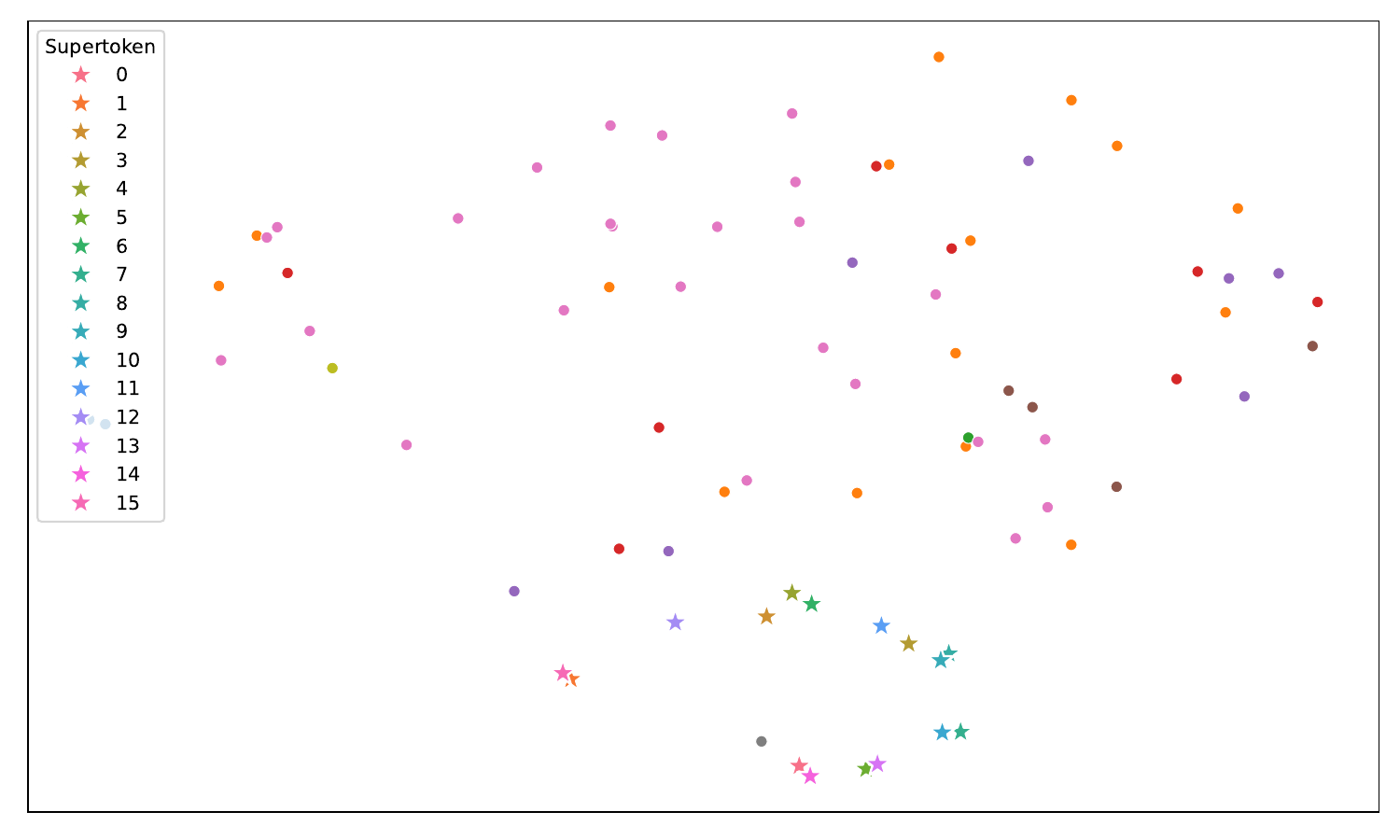} \\
\includegraphics[width=.4\linewidth]{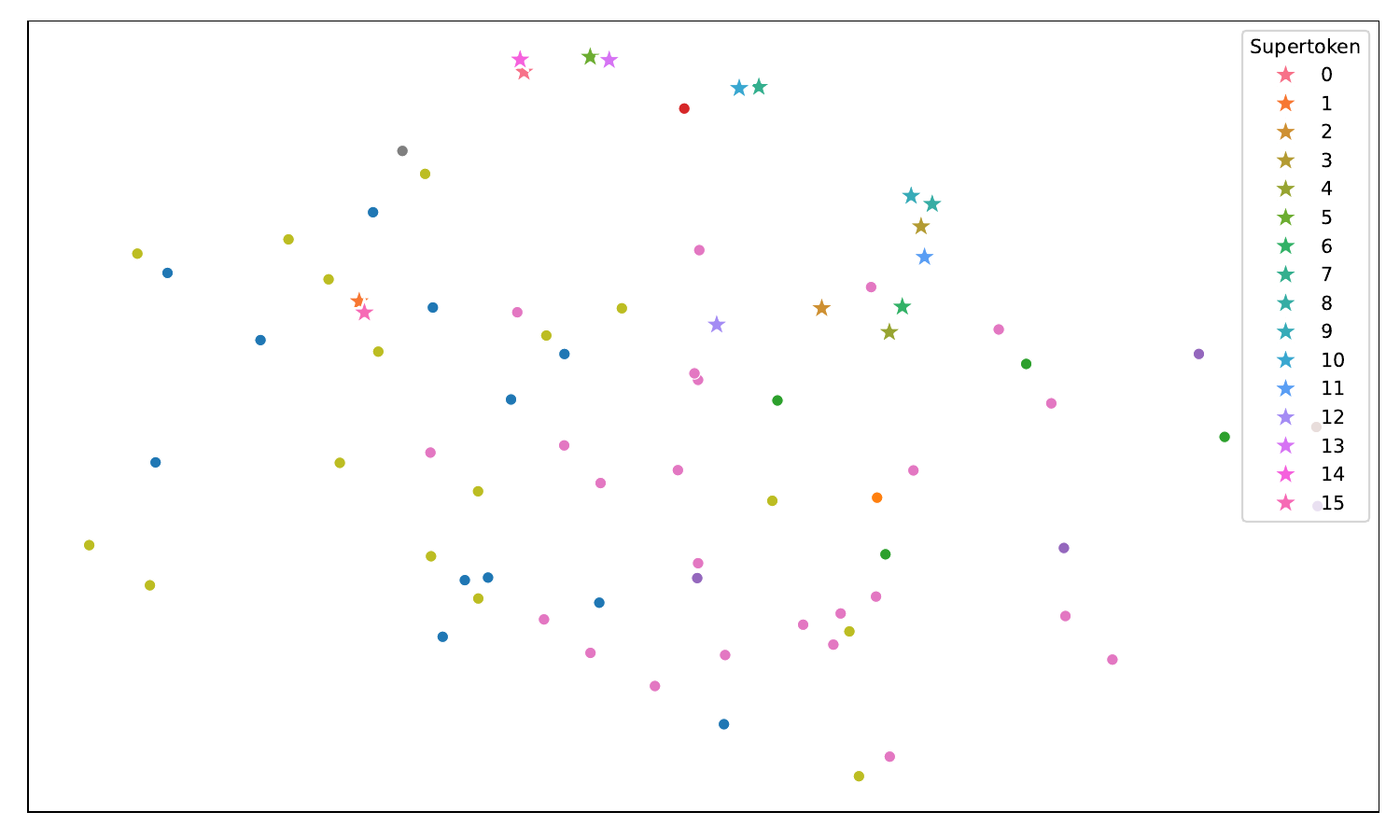} & \includegraphics[width=.4\linewidth]{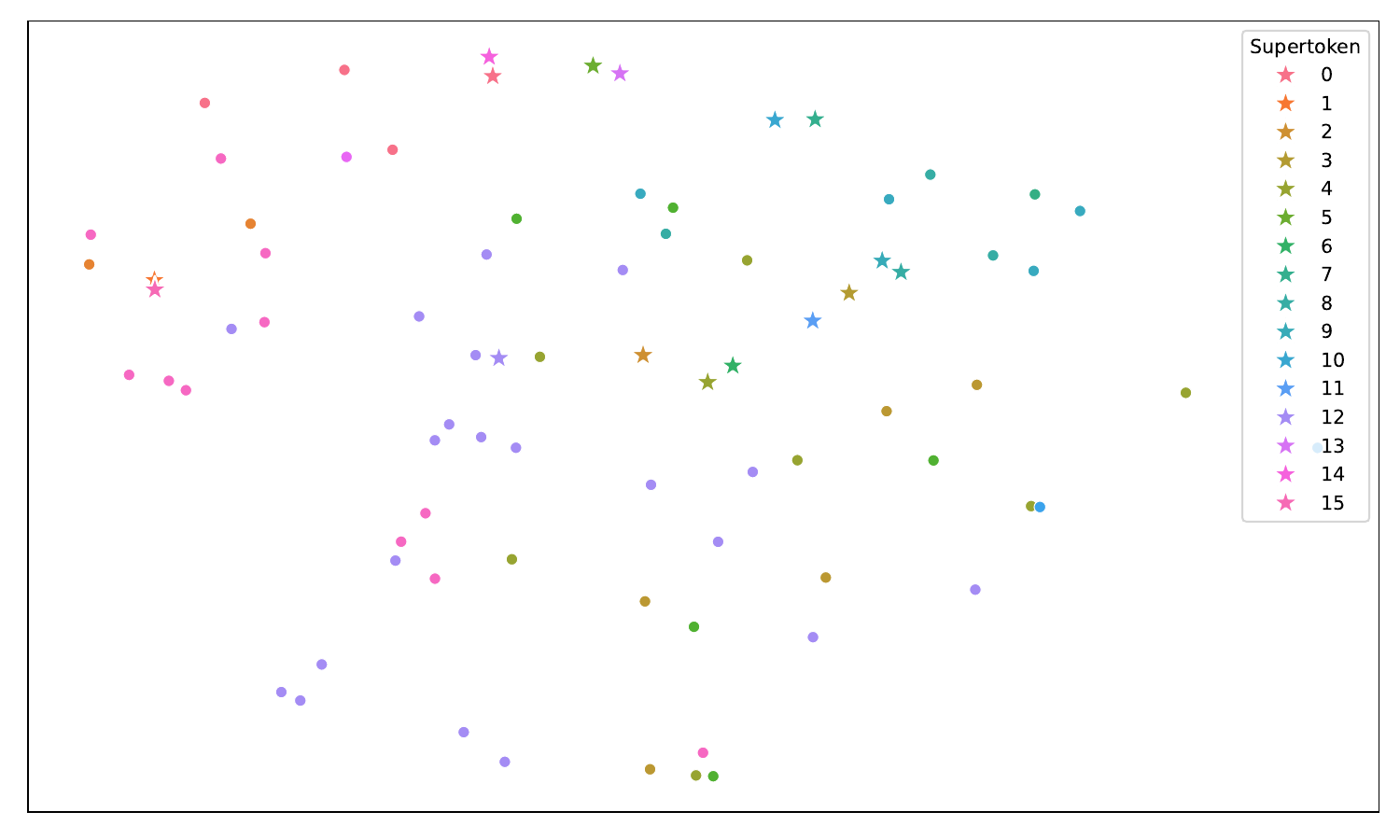} \\
\includegraphics[width=.4\linewidth]{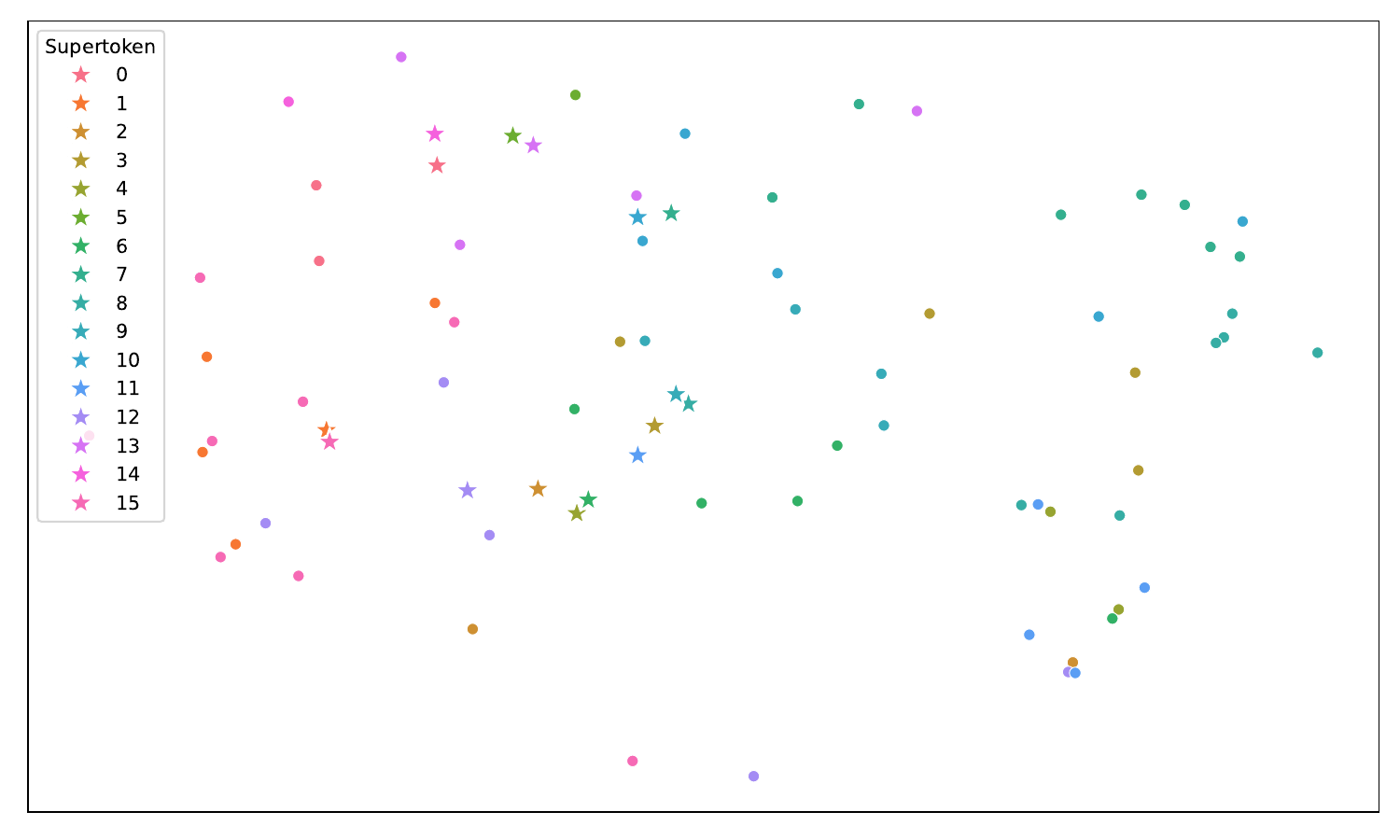} & \includegraphics[width=.4\linewidth]{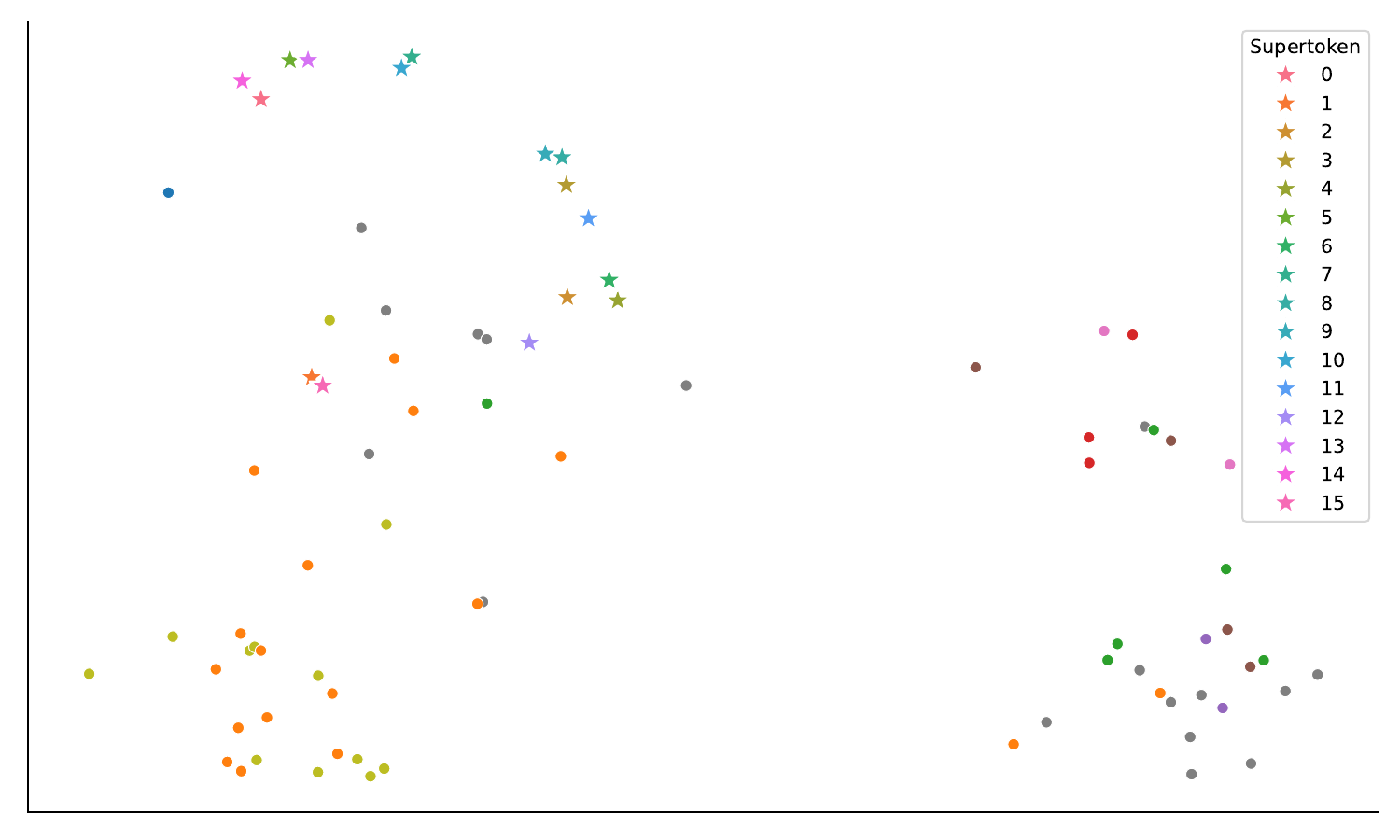}
\end{tabular}
}
\caption{\textbf{Qualitative supertokens selection view in token-supertoken space after projecting.} We select examples from following categories: airplane, chair, guitar, keybord, lamp, plant, sofa and, table. Colors are unique and are associated to supertokens throughout all selected examples. The examples come from validation set of ModelNet40 \cite{wu20153d} and are the same as \cref{fig:appendix:supertokens-vis:spatial}. Classifier is the 16 supertokens best checkpoint.}
\label{fig:appendix:supertokens-vis:semantic-after-projections}
\end{figure*}
}

\section{Hyper-parameters}
\label{sec:hp}
This section highlights the used hyper-parameters for distillation and fine-tunings. \Cref{tab:appendix:distillation-hp} shows used values for important hyper-parameters during distillation. For fine-tunings, we use traditional hyper-parameters and for those specific for \paper and \paper-Gate, we vary the value of the epoch when we unfreezing the encoder for classification (0, 100 and 150) and part segmentation (0 and 300) tasks and we fine-tuned only the distilled checkpoints which no weight decay is added for gate module. For OmniObject3D, we use ModelNet40 configuration.

\begin{table*}[t!]
    \centering
    \begin{tabular}{lp{0.0001cm}cc}
         \toprule
         Hyper-parameter && \papershort & \papershort-Gate \\
         \midrule
         \textit{data} \\
         datasets && ShapeNet55 \cite{chang2015shapenet} & ShapeNet55 \cite{chang2015shapenet} \\
         \# points && 1024 & 1024 \\
         \midrule
         \textit{augmentation} \\
         scale && (0.8, 1.2) & (0.8, 1.2) \\
         anisotropic scaling && enabled & enabled \\
         flip && y axis & y axis \\
         rotate Y axis && $[-\pi,\pi]$ & $[-\pi,\pi]$ \\
         \midrule
         \textit{optimization} \\
         batch size && 512 & 512 \\
         max epochs && 150 & 150 \\
         optimizer && AdamW \cite{DBLP:conf/iclr/LoshchilovH19} & AdamW \cite{DBLP:conf/iclr/LoshchilovH19} \\
         weight decay && $5\times 10^{-2}$ & $5\times 10^{-2}$ \\
         weight decay on gate && - & \{false, true\} \\
         momentum && $0.9$ & $0.9$ \\
         lr && $10^{-3}$ & $10^{-3}$ \\
         start lr && $10^{-6}$ & $10^{-6}$ \\
         final lr && $10^{-6}$ & $10^{-6}$ \\
         lr scheduler && Cosine annealing & Cosine annealing \\
         warmup type && linear & linear \\
         warmup epochs && 15 & 15 \\
         encoder unfreeze && \{50, max epochs\} & \{50, max epochs\} \\
         \midrule
         \textit{architecture} \\
         teacher encoder && ViT-S & ViT-S \\
         student encoder && ViT-T/S & ViT-S \\
         \# groups/tokens && 64 & 64 \\
         group size && 32 & 32 \\
         max \# supertokens && $\{2^i\}_{i=1}^4$ & $16$\\
         use supertoken $\mathbf{P}$ && false & false \\
         bias on DSO qkv && false & false \\
         $\arg\max$ in DSO && true & true \\
         $\lambda_{\text{gate}}$ values && - & $\{0\} \cup \{\,10^i\}_{i=-15}^{1} \cup \{\,10^i \mid i \in A\},$ \\
         &&& $A = \{-10.25,-10.5,-10.75,-11.25,-11.5,-11.75\}$\\
         \midrule
         \textit{hardware} \\
         dtype && float32 & float32 \\
         accelerator && 1 A100 80G & 1 A100 80G \\
         \bottomrule
    \end{tabular}
    \caption{\textbf{Distillation hyper-parameters for \paper and \paper-Gate}.}
    \label{tab:appendix:distillation-hp}
\end{table*}

\end{document}

%% file: tables/finetuning-v1-complete.tex
{\scriptsize\tabcolsep=2pt
\begin{longtable}{l|cc|cc|cccccccc}
\toprule
& Frozen & FT Unfreeze & Distillation & Avg & MN40 & OO3D & \multicolumn{3}{c}{SONN-} & SN55 & \multicolumn{2}{c}{SNP} \\
\# supertokens & student & epoch & loss &&&&\texttt{OBJ-BG}&\texttt{OBJ-ONLY}&\texttt{PB-T50-RS}&& $\text{mIoU}_C$ & $\text{mIoU}_I$ \\
\endfirsthead

\multicolumn{13}{c}%
{{\bfseries \tablename\ \thetable{} -- continued from previous page}} \\
\endhead

\hline \multicolumn{13}{c}{{Continued on next page}} \\ \hline
\endfoot
\endlastfoot

\midrule
baseline ($c=64$) & - & 100 & - & - & 93.02 \tiny{$\pm$ 0.14} & 82.25 \tiny{$\pm$ 0.37} & 91.84 \tiny{$\pm$ 0.47} & 88.38 \tiny{$\pm$ 0.41} & 86.05 \tiny{$\pm$ 0.42} & 90.54 \tiny{$\pm$ 0.14} & - & - \\
baseline ($c=64$) & - & 0   & - & - & - & - & - & - & - & - & 83.91 \tiny{$\pm$ 0.25} & 85.73 \tiny{$\pm$ 0.15} \\
baseline ($c=64$) & - & - & - & 87.72 & 93.02 \tiny{$\pm$ 0.14} & 82.25 \tiny{$\pm$ 0.37} & 91.84 \tiny{$\pm$ 0.47} & 88.38 \tiny{$\pm$ 0.41} & 86.05 \tiny{$\pm$ 0.42} & 90.54 \tiny{$\pm$ 0.14} & 83.91 \tiny{$\pm$ 0.25} & 85.73 \tiny{$\pm$ 0.15}\\

\midrule
\multicolumn{13}{c}{\textit{Random sampling - inference from baseline}} \\
$c=16$ & - & - & - & 75.20 & 90.57 \tiny{$\pm$ 0.41} & 65.19 \tiny{$\pm$ 0.81} & 78.95 \tiny{$\pm$ 1.19} & 72.70 \tiny{$\pm$ 1.14} & 69.74 \tiny{$\pm$ 0.91} & 87.80 \tiny{$\pm$ 0.22} & 66.96 \tiny{$\pm$ 0.71} & 69.71 \tiny{$\pm$ 0.16} \\
$c=8$ & - & - & - & 63.45 & 83.72 \tiny{$\pm$ 0.60} & 41.20 \tiny{$\pm$ 1.35} & 67.40 \tiny{$\pm$ 1.60} & 62.67 \tiny{$\pm$ 1.47} & 55.78 \tiny{$\pm$ 0.83} & 80.72 \tiny{$\pm$ 0.36} & 56.47 \tiny{$\pm$ 0.62} & 59.60 \tiny{$\pm$ 0.19} \\
$c=4$ & - & - & - & 47.46 & 65.45 \tiny{$\pm$ 0.80} & 19.74 \tiny{$\pm$ 1.08} & 50.79 \tiny{$\pm$ 0.69} & 45.30 \tiny{$\pm$ 1.41} & 39.39 \tiny{$\pm$ 0.72} & 62.51 \tiny{$\pm$ 0.38} & 46.35 \tiny{$\pm$ 0.46} & 50.11 \tiny{$\pm$ 0.23} \\
$c=2$ & - & - & - & 32.10 & 39.32 \tiny{$\pm$ 0.55} & 08.03 \tiny{$\pm$ 0.52} & 35.51 \tiny{$\pm$ 1.52} & 32.19 \tiny{$\pm$ 1.78} & 27.07 \tiny{$\pm$ 0.62} & 36.00 \tiny{$\pm$ 0.29} & 37.11 \tiny{$\pm$ 0.27} & 41.56 \tiny{$\pm$ 0.10} \\
$c=1$ & - & - & - & 21.46 & 18.33 \tiny{$\pm$ 0.66} & 03.82 \tiny{$\pm$ 0.32} & 23.22 \tiny{$\pm$ 1.55} & 21.31 \tiny{$\pm$ 1.35} & 19.90 \tiny{$\pm$ 0.68} & 15.54 \tiny{$\pm$ 0.23} & 33.70 \tiny{$\pm$ 0.30} & 35.89 \tiny{$\pm$ 0.16} \\

\midrule
\multicolumn{13}{c}{\textit{Updating group size to reduce number of generated tokens - inference from baseline}} \\
$c=16$ & - & - & - & 41.66 & 63.45 \tiny{$\pm$ 0.47} & 13.29 \tiny{$\pm$ 0.42} & 11.76 \tiny{$\pm$ 0.16} & 22.24 \tiny{$\pm$ 0.29} & 12.74 \tiny{$\pm$ 0.17} & 59.99 \tiny{$\pm$ 0.19} & 72.98 \tiny{$\pm$ 0.21} & 76.84 \tiny{$\pm$ 0.08} \\
$c=8$ & - & - & - & 26.36 & 20.57 \tiny{$\pm$ 0.19} & 03.62 \tiny{$\pm$ 0.19} & 09.29 \tiny{$\pm$ 0.00} & 17.33 \tiny{$\pm$ 0.36} & 09.37 \tiny{$\pm$ 0.00} & 23.12 \tiny{$\pm$ 0.18} & 61.36 \tiny{$\pm$ 0.35} & 66.24 \tiny{$\pm$ 0.13} \\
$c=4$ & - & - & - & 20.23 & 10.32 \tiny{$\pm$ 0.25} & 00.66 \tiny{$\pm$ 0.14} & 09.29 \tiny{$\pm$ 0.00} & 15.44 \tiny{$\pm$ 0.34} & 09.37 \tiny{$\pm$ 0.00} & 12.65 \tiny{$\pm$ 0.21} & 50.31 \tiny{$\pm$ 0.36} & 53.80 \tiny{$\pm$ 0.13} \\
$c=2$ & - & - & - & 15.12 & 04.79 \tiny{$\pm$ 0.15} & 00.28 \tiny{$\pm$ 0.10} & 09.29 \tiny{$\pm$ 0.00} & 12.67 \tiny{$\pm$ 0.36} & 09.37 \tiny{$\pm$ 0.00} & 05.53 \tiny{$\pm$ 0.11} & 37.08 \tiny{$\pm$ 0.09} & 41.95 \tiny{$\pm$ 0.08} \\
$c=1$ & - & - & - & 15.78 & 03.88 \tiny{$\pm$ 0.19} & 00.47 \tiny{$\pm$ 0.13} & 09.24 \tiny{$\pm$ 0.63} & 15.85 \tiny{$\pm$ 0.82} & 09.37 \tiny{$\pm$ 0.00} & 04.25 \tiny{$\pm$ 0.09} & 40.30 \tiny{$\pm$ 0.44} & 42.90 \tiny{$\pm$ 0.34} \\

\midrule
\multicolumn{13}{c}{\textit{ToMe \cite{DBLP:conf/iclr/BolyaFDZFH23} - inference from baseline (no additional training needed)}} \\
$s=16$ & - & - & - & 73.80 & 89.10 \tiny{$\pm$ 0.23} & 50.51 \tiny{$\pm$ 0.39} & 82.19 \tiny{$\pm$ 0.56} & 84.22 \tiny{$\pm$ 0.62} & 73.87 \tiny{$\pm$ 0.50} & 84.89 \tiny{$\pm$ 0.24} & 63.09 \tiny{$\pm$ 0.40} & 62.51 \tiny{$\pm$ 0.08} \\
$s=8$ & - & - & - & 66.79 & 84.02 \tiny{$\pm$ 0.38} & 33.54 \tiny{$\pm$ 0.78} & 76.87 \tiny{$\pm$ 1.30} & 79.88 \tiny{$\pm$ 0.96} & 67.32 \tiny{$\pm$ 0.62} & 79.68 \tiny{$\pm$ 0.26} & 56.81 \tiny{$\pm$ 0.43} & 56.23 \tiny{$\pm$ 0.14} \\
$s=4$ & - & - & - & 59.88 & 74.92 \tiny{$\pm$ 0.65} & 20.91 \tiny{$\pm$ 0.59} & 71.57 \tiny{$\pm$ 1.05} & 74.29 \tiny{$\pm$ 0.54} & 61.52 \tiny{$\pm$ 0.49} & 72.05 \tiny{$\pm$ 0.30} & 51.59 \tiny{$\pm$ 0.52} & 52.19 \tiny{$\pm$ 0.12} \\
$s=2$ & - & - & - & 53.19 & 62.76 \tiny{$\pm$ 0.44} & 11.21 \tiny{$\pm$ 0.36} & 67.76 \tiny{$\pm$ 1.03} & 68.49 \tiny{$\pm$ 0.84} & 56.30 \tiny{$\pm$ 0.62} & 61.66 \tiny{$\pm$ 0.30} & 47.47 \tiny{$\pm$ 0.45} & 49.88 \tiny{$\pm$ 0.12} \\
$s=1$ & - & - & - & 46.15 & 48.35 \tiny{$\pm$ 0.31} & 07.20 \tiny{$\pm$ 0.20} & 61.20 \tiny{$\pm$ 0.89} & 62.87 \tiny{$\pm$ 0.95} & 51.69 \tiny{$\pm$ 0.46} & 42.15 \tiny{$\pm$ 0.18} & 45.36 \tiny{$\pm$ 0.19} & 50.37 \tiny{$\pm$ 0.09} \\

\midrule
\multicolumn{13}{c}{\textit{ToMe \cite{DBLP:conf/iclr/BolyaFDZFH23} - Trained using \textbf{FMD (Ours)} + finetuning}} \\
$s=16$ & \xmark & - & 0.0658 & 86.52 & 93.07 & 80.95 & 90.02 & 87.44 & 83.66 & 89.84 & 82.41 & 84.76 \\
$s=1$  & \xmark & - & 0.1509 & 85.87 & 92.14 & 80.08 & 89.85 & 86.40 & 83.00 & 89.84 & 81.14 & 84.50 \\
\midrule
\multicolumn{13}{c}{\textit{PiToMe \cite{DBLP:conf/nips/TranNNNLXSZNN24} - Trained using \textbf{FMD (Ours)} + finetuning}} \\
$s=16$ & \xmark & - & 0.0643 & 86.50 & 92.83 & 81.97 & 89.33 & 87.95 & 83.66 & 89.88 & 81.77 & 84.65 \\
$s=1$  & \xmark & - & 0.1508 & 86.14 & 92.55 & 80.95 & 90.19 & 86.23 & 82.89 & 89.76 & 81.84 & 84.74 \\
\midrule
\multicolumn{13}{c}{\textit{PatchMerger \cite{DBLP:journals/corr/abs-2202-12015} - Trained using \textbf{FMD (Ours)} + finetuning}} \\
$s=16$ & \xmark & - & 0.0326 & 87.59 & 92.83 & 82.76 & 91.39 & 88.99 & 85.53 & 90.11 & 83.59 & 85.55 \\
$s=1$  & \xmark & - & 0.2811 & 87.14 & 92.18 & 82.13 & 90.71 & 88.12 & 85.05 & 89.71 & 83.64 & 85.54 \\

\midrule
\multicolumn{13}{c}{\textit{\textbf{\paper (Ours)}}} \\
$s=16$ & \cmark & 300 & 0.1006 & - & - & - & - & - & - & - & 81.55 & 84.59 \\
$s=16$ & \cmark & 150 & 0.1006 & - & 91.09 & 76.61 & 84.68 & 85.54 & 80.81 & 89.68 & - & - \\
$s=16$ & \cmark & 100 & 0.1006 & - & 91.73 & 77.48 & 85.20 & 84.17 & 80.29 & 89.67 & - & - \\
$s=16$ & \cmark &   0 & 0.1006 & - & 91.41 & 77.80 & 84.34 & 85.20 & 78.70 & 89.32 & 82.21 & 85.05 \\
$s=16$ & \cmark & - & 0.1006 & 84.33 & 91.41 \tiny{$\pm$ 0.32} & 77.30 \tiny{$\pm$ 0.61} & 84.74 \tiny{$\pm$ 0.43} & 84.97 \tiny{$\pm$ 0.72} & 79.93 \tiny{$\pm$ 1.10} & 89.56 \tiny{$\pm$ 0.21} & 81.88 \tiny{$\pm$ 0.47} & 84.82 \tiny{$\pm$ 0.32} \\
$s=16$ & \xmark & 300 & 0.0736 & - & - & - & - & - & - & - & 81.90 & 84.87 \\
$s=16$ & \xmark & 150 & 0.0736 & - & 91.69 & 77.72 & 86.06 & 86.75 & 80.43 & 89.88 & - & - \\
$s=16$ & \xmark & 100 & 0.0736 & - & 92.02 & 78.03 & 86.75 & 85.89 & 81.23 & 89.95 & - & - \\
$s=16$ & \xmark &   0 & 0.0736 & - & 91.53 & 77.64 & 85.89 & 86.23 & 79.42 & 89.78 & 81.85 & 84.77 \\
$s=16$ & \xmark & - & 0.0736 & 84.87 & 91.75 \tiny{$\pm$ 0.25} & 77.80 \tiny{$\pm$ 0.21} & 86.23 \tiny{$\pm$ 0.46} & 86.29 \tiny{$\pm$ 0.43} & 80.36 \tiny{$\pm$ 0.90} & 89.87 \tiny{$\pm$ 0.09} & 81.87 \tiny{$\pm$ 0.04} & 84.82 \tiny{$\pm$ 0.07} \\
$s=16$ (ViT-T) & \xmark & 300 & 0.1040 & - & - & - & - & - & - & - & 81.85 & 84.49 \\
$s=16$ (ViT-T) & \xmark & 150 & 0.1040 & - & 91.69 & 75.75 & 84.34 & 83.99 & 79.49 & 89.04 & - & - \\
$s=16$ (ViT-T) & \xmark & 100 & 0.1040 & - & 91.77 & 76.22 & 83.30 & 85.89 & 79.46 & 88.56 & - & - \\
$s=16$ (ViT-T) & \xmark &   0 & 0.1040 & - & 91.00 & 75.98 & 82.62 & 84.51 & 79.18 & 88.41 & 81.51 & 84.63 \\
$s=16$ (ViT-T) & \xmark & - & 0.1040 & 83.75 & 91.49 \tiny{$\pm$ 0.42} & 75.98 \tiny{$\pm$ 0.24} & 83.42 \tiny{$\pm$ 0.87} & 84.80 \tiny{$\pm$ 0.98} & 79.38 \tiny{$\pm$ 0.17} & 88.67 \tiny{$\pm$ 0.33} & 81.68 \tiny{$\pm$ 0.24} & 84.56 \tiny{$\pm$ 0.10} \\
$s=8$ & \cmark & 300 & 0.0993 & - & - & - & - & - & - & - & 82.08 & 84.53 \\
$s=8$ & \cmark & 150 & 0.0993 & - & 91.69 & 76.69 & 85.37 & 84.51 & 81.05 & 89.48 & - & - \\
$s=8$ & \cmark & 100 & 0.0993 & - & 91.82 & 76.69 & 86.23 & 84.68 & 80.64 & 89.56 & - & - \\
$s=8$ & \cmark &   0 & 0.0993 & - & 91.61 & 76.06 & 85.20 & 86.23 & 79.15 & 89.23 & 81.60 & 84.60 \\
$s=8$ & \cmark & - & 0.0993 & 84.38 & 91.71 \tiny{$\pm$ 0.10} & 76.48 \tiny{$\pm$ 0.36} & 85.60 \tiny{$\pm$ 0.55} & 85.14 \tiny{$\pm$ 0.95} & 80.28 \tiny{$\pm$ 1.00} & 89.42 \tiny{$\pm$ 0.17} & 81.84 \tiny{$\pm$ 0.34} & 84.57 \tiny{$\pm$ 0.05} \\
$s=8$ & \xmark & 300 & 0.0749 & - & - & - & - & - & - & - & 82.09 & 84.86 \\
$s=8$ & \xmark & 150 & 0.0749 & - & 91.49 & 78.11 & 86.23 & 86.57 & 80.46 & 89.67 & - & - \\
$s=8$ & \xmark & 100 & 0.0749 & - & 91.86 & 77.17 & 85.89 & 84.85 & 81.19 & 89.81 & - & - \\
$s=8$ & \xmark &   0 & 0.0749 & - & 91.53 & 76.69 & 86.57 & 86.75 & 79.74 & 89.34 & 81.81 & 84.84 \\
$s=8$ & \xmark & - & 0.0749 & 84.76 & 91.63 \tiny{$\pm$ 0.20} & 77.32 \tiny{$\pm$ 0.72} & 86.23 \tiny{$\pm$ 0.34} & 86.06 \tiny{$\pm$ 1.05} & 80.46 \tiny{$\pm$ 0.73} & 89.61 \tiny{$\pm$ 0.24} & 81.95 \tiny{$\pm$ 0.19} & 84.85 \tiny{$\pm$ 0.01} \\
$s=4$ & \cmark & 300 & 0.0986 & - & - & - & - & - & - & - & 81.97 & 84.72 \\
$s=4$ & \cmark & 150 & 0.0986 & - & 91.09 & 77.24 & 84.51 & 84.85 & 79.53 & 89.41 & - & - \\
$s=4$ & \cmark & 100 & 0.0986 & - & 91.33 & 76.93 & 85.03 & 86.92 & 79.77 & 89.43 & - & - \\
$s=4$ & \cmark &   0 & 0.0986 & - & 91.33 & 76.69 & 83.82 & 86.06 & 79.53 & 88.95 & 81.71 & 84.61 \\
$s=4$ & \cmark & - & 0.0986 & 84.25 & 91.25 \tiny{$\pm$ 0.14} & 76.96 \tiny{$\pm$ 0.28} & 84.45 \tiny{$\pm$ 0.60} & 85.94 \tiny{$\pm$ 1.04} & 79.61 \tiny{$\pm$ 0.14} & 89.26 \tiny{$\pm$ 0.27} & 81.84 \tiny{$\pm$ 0.18} & 84.67 \tiny{$\pm$ 0.08} \\
$s=4$ & \xmark & 300 & 0.0788 & - & - & - & - & - & - & - & 81.57 & 84.76 \\
$s=4$ & \xmark & 150 & 0.0788 & - & 91.45 & 76.14 & 86.23 & 85.89 & 79.67 & 89.81 & - & - \\
$s=4$ & \xmark & 100 & 0.0788 & - & 91.41 & 77.87 & 87.09 & 86.40 & 80.95 & 89.74 & - & - \\
$s=4$ & \xmark &   0 & 0.0788 & - & 91.57 & 76.06 & 84.68 & 85.37 & 81.47 & 89.68 & 81.99 & 84.86 \\
$s=4$ & \xmark & - & 0.0788 & 84.64 & 91.48 \tiny{$\pm$ 0.08} & 76.69 \tiny{$\pm$ 1.02} & 86.00 \tiny{$\pm$ 1.22} & 85.89 \tiny{$\pm$ 0.52} & 80.70 \tiny{$\pm$ 0.93} & 89.74 \tiny{$\pm$ 0.06} & 81.78 \tiny{$\pm$ 0.30} & 84.81 \tiny{$\pm$ 0.07} \\
$s=2$ & \cmark & 300 & 0.0979 & - & - & - & - & - & - & - & 81.58 & 84.47 \\
$s=2$ & \cmark & 150 & 0.0979 & - & 92.06 & 77.09 & 83.82 & 83.30 & 79.84 & 89.26 & - & - \\
$s=2$ & \cmark & 100 & 0.0979 & - & 91.00 & 76.77 & 84.68 & 85.37 & 80.12 & 89.50 & - & - \\
$s=2$ & \cmark &   0 & 0.0979 & - & 91.49 & 77.40 & 85.03 & 84.85 & 79.98 & 89.27 & 82.10 & 84.63 \\
$s=2$ & \cmark & - & 0.0979 & 84.17 & 91.52 \tiny{$\pm$ 0.53} & 77.09 \tiny{$\pm$ 0.32} & 84.51 \tiny{$\pm$ 0.62} & 84.51 \tiny{$\pm$ 1.08} & 79.98 \tiny{$\pm$ 0.14} & 89.34 \tiny{$\pm$ 0.14} & 81.84 \tiny{$\pm$ 0.36} & 84.55 \tiny{$\pm$ 0.11} \\
$s=2$ & \xmark & 300 & 0.0795 & - & - & - & - & - & - & - & 82.13 & 84.50 \\
$s=2$ & \xmark & 150 & 0.0795 & - & 91.65 & 78.66 & 85.37 & 86.40 & 80.57 & 89.68 & - & - \\
$s=2$ & \xmark & 100 & 0.0795 & - & 92.02 & 78.74 & 85.71 & 86.40 & 81.16 & 89.67 & - & - \\
$s=2$ & \xmark &   0 & 0.0795 & - & 91.61 & 77.56 & 86.40 & 85.54 & 80.60 & 89.62 & 81.84 & 84.44 \\
$s=2$ & \xmark & - & 0.0795 & 84.86 & 91.76 \tiny{$\pm$ 0.22} & 78.32 \tiny{$\pm$ 0.66} & 85.83 \tiny{$\pm$ 0.53} & 86.12 \tiny{$\pm$ 0.50} & 80.78 \tiny{$\pm$ 0.33} & 89.66 \tiny{$\pm$ 0.04} & 81.98 \tiny{$\pm$ 0.21} & 84.47 \tiny{$\pm$ 0.04} \\
$s=1$ & \cmark & 300 & 0.0967 & - & - & - & - & - & - & - & 81.51 & 84.61 \\
$s=1$ & \cmark & 150 & 0.0967 & - & 91.94 & 76.93 & 84.68 & 85.37 & 79.77 & 89.14 & - & - \\
$s=1$ & \cmark & 100 & 0.0967 & - & 91.49 & 76.69 & 85.37 & 84.17 & 79.53 & 89.37 & - & - \\
$s=1$ & \cmark &   0 & 0.0967 & - & 91.37 & 77.95 & 84.34 & 85.54 & 79.11 & 89.35 & 81.40 & 84.47 \\
$s=1$ & \cmark & - & 0.0967 & 84.17 & 91.60 \tiny{$\pm$ 0.30} & 77.19 \tiny{$\pm$ 0.67} & 84.80 \tiny{$\pm$ 0.53} & 85.03 \tiny{$\pm$ 0.75} & 79.47 \tiny{$\pm$ 0.33} & 89.29 \tiny{$\pm$ 0.13} & 81.46 \tiny{$\pm$ 0.08} & 84.54 \tiny{$\pm$ 0.10} \\
$s=1$ & \xmark & 300 & 0.0792 & - & - & - & - & - & - & - & 81.66 & 84.56 \\
$s=1$ & \xmark & 150 & 0.0792 & - & 91.86 & 77.64 & 86.06 & 85.20 & 81.09 & 89.67 & - & - \\
$s=1$ & \xmark & 100 & 0.0792 & - & 91.57 & 77.72 & 83.65 & 87.09 & 80.40 & 89.69 & - & - \\
$s=1$ & \xmark &   0 & 0.0792 & - & 91.73 & 77.80 & 83.99 & 85.89 & 80.46 & 89.58 & 81.86 & 84.43 \\
$s=1$ & \xmark & - & 0.0792 & 84.58 & 91.72 \tiny{$\pm$ 0.14} & 77.72 \tiny{$\pm$ 0.08} & 84.57 \tiny{$\pm$ 1.30} & 86.06 \tiny{$\pm$ 0.96} & 80.65 \tiny{$\pm$ 0.38} & 89.65 \tiny{$\pm$ 0.06} & 81.76 \tiny{$\pm$ 0.14} & 84.50 \tiny{$\pm$ 0.09} \\
\bottomrule

\caption{\textbf{Detailed finetuning results of \paper.} All presented results are finetunings from a frozen or unfrozen backbone during distillation (no distillation stage for the baseline). For the classification task on ModelNet40 (MN40), OmniObject3D (OO3D), the three splits of ScanObjectNN (SONN-\{\texttt{PB-T50-RS}, \texttt{OBJ-BG}, \texttt{OBJ-ONLY}\}) and ShapeNet55 (SN55), the used metric is the top-1 accuracy on the validation set. For part segmentation with ShapeNetPart (SNP), we show for both category and instance mIoUs. Mean $\pm$ Std Dev over 10 runs for the baseline. "FT Unfreeze Epoch" denotes for the epoch when we completely unfreeze the ViT backbone during the finetuning stage ('-' in this column means aggregation over the results above). In columns with results, '-' stands for no running was performed. ToMe \cite{DBLP:conf/iclr/BolyaFDZFH23} and PiToMe \cite{DBLP:conf/nips/TranNNNLXSZNN24} halve the number of tokens at each layer up to the specified number of tokens (supertokens, to simplify the table columns). For distilled ToMe, PiToMe and PatchMerger \cite{DBLP:journals/corr/abs-2202-12015}, the distilled encoder is unfrozen at 100 epochs for classification datasets and at 0 epochs for part segmentation.}
\label{tab:finetunings-v1-detailed}
\end{longtable}
}

%% file: tables/finetuning-v3-complete.tex
{\scriptsize\tabcolsep=2pt
\begin{longtable}{l|cc|cc|cccccccc}
\toprule
& Frozen & FT Unfreeze & Distillation & Avg & MN40 & OO3D & \multicolumn{3}{c}{SONN-} & SN55 & \multicolumn{2}{c}{SNP} \\
$\lambda_{\text{gate}}$ & student & epoch & loss &&&&\texttt{OBJ-BG}&\texttt{OBJ-ONLY}&\texttt{PB-T50-RS}&& $\text{mIoU}_C$ & $\text{mIoU}_I$ \\
\endfirsthead

\multicolumn{13}{c}%
{{\bfseries \tablename\ \thetable{} -- continued from previous page}} \\
\endhead

\hline \multicolumn{13}{c}{{Continued on next page}} \\ \hline
\endfoot
\endlastfoot

\midrule
baseline & - & - & - & 87.72 & 93.02 \tiny{$\pm$ 0.14} & 82.25 \tiny{$\pm$ 0.37} & 91.84 \tiny{$\pm$ 0.47} & 88.38 \tiny{$\pm$ 0.41} & 86.05 \tiny{$\pm$ 0.42} & 90.54 \tiny{$\pm$ 0.14} & 83.91 \tiny{$\pm$ 0.25} & 85.73 \tiny{$\pm$ 0.15} \\
\midrule
\multicolumn{13}{c}{\textit{\textbf{\paper-Gate (Ours)}}} \\
$0$ & \cmark & - & 0.0191 & - & - & - & - & - & - & - & - & - \\
$0$ & \xmark & - & 0.0152 & - & - & - & - & - & - & - & - & - \\
$10^{-15}$ & \cmark & 300 & 0.0191 & - & - & - & - & - & - & - & 82.64 & 85.24 \\
$10^{-15}$ & \cmark & 150 & 0.0191 & - & 93.03 & 79.37 & 89.67 & 87.95 & 83.10 & 90.30 & - & - \\
$10^{-15}$ & \cmark & 100 & 0.0191 & - & 93.48 & 81.97 & 92.77 & 88.12 & 86.02 & 90.75 & - & - \\
$10^{-15}$ & \cmark & 0 & 0.0191 & - & 92.22 & 83.23 & 90.53 & 88.30 & 84.84 & 90.31 & 83.77 & 85.63 \\
$10^{-15}$ & \cmark & - & 0.0191 & 87.16 & 92.91 \tiny{$\pm$ 0.64} & 81.52 \tiny{$\pm$ 1.97} & 90.99 \tiny{$\pm$ 1.60} & 88.12 \tiny{$\pm$ 0.17} & 84.65 \tiny{$\pm$ 1.47} & 90.45 \tiny{$\pm$ 0.26} & 83.21 \tiny{$\pm$ 0.79} & 85.44 \tiny{$\pm$ 0.27} \\
$10^{-15}$ & \xmark & 300 & 0.0142 & - & - & - & - & - & - & - & 82.76 & 85.06 \\
$10^{-15}$ & \xmark & 150 & 0.0142 & - & 93.56 & 80.24 & 88.30 & 87.95 & 82.86 & 90.50 & - & - \\
$10^{-15}$ & \xmark & 100 & 0.0142 & - & 93.19 & 83.23 & 92.25 & 89.67 & 86.12 & 90.55 & - & - \\
$10^{-15}$ & \xmark & 0 & 0.0142 & - & 92.14 & 82.36 & 90.88 & 89.67 & 85.50 & 90.38 & 83.93 & 85.62 \\
$10^{-15}$ & \xmark & - & 0.0142 & 87.31 & 92.96 \tiny{$\pm$ 0.74} & 81.94 \tiny{$\pm$ 1.54} & 90.48 \tiny{$\pm$ 2.01} & 89.10 \tiny{$\pm$ 0.99} & 84.83 \tiny{$\pm$ 1.73} & 90.48 \tiny{$\pm$ 0.09} & 83.35 \tiny{$\pm$ 0.83} & 85.34 \tiny{$\pm$ 0.39} \\
$10^{-14}$ & \cmark & 300 & 0.0191 & - & - & - & - & - & - & - & 82.78 & 85.11 \\
$10^{-14}$ & \cmark & 150 & 0.0191 & - & 93.31 & 79.61 & 90.71 & 88.47 & 83.07 & 90.36 & - & - \\
$10^{-14}$ & \cmark & 100 & 0.0191 & - & 92.87 & 81.73 & 91.22 & 87.78 & 86.36 & 90.58 & - & - \\
$10^{-14}$ & \cmark & 0 & 0.0191 & - & 92.59 & 82.28 & 90.36 & 87.95 & 85.60 & 90.24 & 83.66 & 85.42 \\
$10^{-14}$ & \cmark & - & 0.0191 & 87.11 & 92.92 \tiny{$\pm$ 0.37} & 81.21 \tiny{$\pm$ 1.41} & 90.76 \tiny{$\pm$ 0.43} & 88.07 \tiny{$\pm$ 0.36} & 85.01 \tiny{$\pm$ 1.73} & 90.39 \tiny{$\pm$ 0.17} & 83.22 \tiny{$\pm$ 0.62} & 85.26 \tiny{$\pm$ 0.21} \\
$10^{-14}$ & \xmark & 300 & 0.0141 & - & - & - & - & - & - & - & 82.55 & 85.12 \\
$10^{-14}$ & \xmark & 150 & 0.0141 & - & 93.15 & 80.31 & 90.02 & 88.47 & 83.69 & 90.22 & - & - \\
$10^{-14}$ & \xmark & 100 & 0.0141 & - & 93.15 & 82.60 & 90.71 & 89.16 & 85.91 & 90.52 & - & - \\
$10^{-14}$ & \xmark & 0 & 0.0141 & - & 92.30 & 82.68 & 91.05 & 87.61 & 84.39 & 90.21 & 84.06 & 85.60 \\
$10^{-14}$ & \xmark & - & 0.0141 & 87.17 & 92.87 \tiny{$\pm$ 0.49} & 81.86 \tiny{$\pm$ 1.34} & 90.59 \tiny{$\pm$ 0.53} & 88.41 \tiny{$\pm$ 0.78} & 84.66 \tiny{$\pm$ 1.14} & 90.32 \tiny{$\pm$ 0.18} & 83.30 \tiny{$\pm$ 1.07} & 85.36 \tiny{$\pm$ 0.34} \\
$10^{-13}$ & \cmark & 300 & 0.0193 & - & - & - & - & - & - & - & 82.68 & 85.11 \\
$10^{-13}$ & \cmark & 150 & 0.0193 & - & 92.95 & 80.08 & 89.67 & 87.09 & 83.21 & 90.43 & - & - \\
$10^{-13}$ & \cmark & 100 & 0.0193 & - & 92.74 & 82.36 & 91.22 & 87.78 & 86.09 & 90.53 & - & - \\
$10^{-13}$ & \cmark & 0 & 0.0193 & - & 92.018 & 81.65 & 91.05 & 87.09 & 85.36 & 90.08 & 83.88 & 85.50 \\
$10^{-13}$ & \cmark & - & 0.0193 & 86.96 & 92.57 \tiny{$\pm$ 0.49} & 81.36 \tiny{$\pm$ 1.17} & 90.65 \tiny{$\pm$ 0.85} & 87.32 \tiny{$\pm$ 0.40} & 84.88 \tiny{$\pm$ 1.50} & 90.35 \tiny{$\pm$ 0.23} & 83.28 \tiny{$\pm$ 0.85} & 85.30 \tiny{$\pm$ 0.28} \\
$10^{-13}$ & \xmark & 300 & 0.0141 & - & - & - & - & - & - & - & 82.85 & 85.07 \\
$10^{-13}$ & \xmark & 150 & 0.0141 & - & 93.23 & 79.84 & 90.19 & 89.16 & 83.03 & 90.15 & - & - \\
$10^{-13}$ & \xmark & 100 & 0.0141 & - & 93.07 & 82.68 & 90.02 & 88.12 & 85.95 & 90.47 & - & - \\
$10^{-13}$ & \xmark & 0 & 0.0141 & - & 92.50 & 82.20 & 90.36 & 89.50 & 84.80 & 90.30 & 83.79 & 85.61 \\
$10^{-13}$ & \xmark & - & 0.0141 & 87.15 & 92.94 \tiny{$\pm$ 0.38} & 81.57 \tiny{$\pm$ 1.52} & 90.19 \tiny{$\pm$ 0.17} & 88.93 \tiny{$\pm$ 0.72} & 84.59 \tiny{$\pm$ 1.47} & 90.31 \tiny{$\pm$ 0.16} & 83.32 \tiny{$\pm$ 0.66} & 85.34 \tiny{$\pm$ 0.39} \\
$10^{-12}$ & \cmark & 300 & 0.0218 & - & - & - & - & - & - & - & 82.76 & 85.26 \\
$10^{-12}$ & \cmark & 150 & 0.0218 & - & 93.19 & 79.61 & 89.33 & 87.95 & 82.17 & 89.71 & - & - \\
$10^{-12}$ & \cmark & 100 & 0.0218 & - & 93.15 & 81.89 & 91.05 & 88.47 & 85.91 & 89.78 & - & - \\
$10^{-12}$ & \cmark & 0 & 0.0218 & - & 91.69 & 81.89 & 89.85 & 88.64 & 74.98 & 89.34 & 83.97 & 85.76 \\
$10^{-12}$ & \cmark & - & 0.0218 & 86.47 & 92.68 \tiny{$\pm$ 0.85} & 81.13 \tiny{$\pm$ 1.32} & 90.07 \tiny{$\pm$ 0.88} & 88.35 \tiny{$\pm$ 0.36} & 81.02 \tiny{$\pm$ 5.55} & 89.61 \tiny{$\pm$ 0.24} & 83.37 \tiny{$\pm$ 0.86} & 85.51 \tiny{$\pm$ 0.35} \\
$10^{-12}$ & \xmark & 300 & 0.0157 & - & - & - & - & - & - & - & 82.64 & 84.99 \\
$10^{-12}$ & \xmark & 150 & 0.0157 & - & 92.99 & 80.00 & 89.16 & 88.47 & 83.14 & 88.60 & - & - \\
$10^{-12}$ & \xmark & 100 & 0.0157 & - & 93.15 & 82.36 & 91.74 & 88.81 & 84.84 & 90.07 & - & - \\
$10^{-12}$ & \xmark & 0 & 0.0157 & - & 91.65 & 81.50 & 90.02 & 88.47 & 74.57 & 89.74 & 83.70 & 85.69 \\
$10^{-12}$ & \xmark & - & 0.0157 & 86.45 & 92.60 \tiny{$\pm$ 0.82} & 81.29 \tiny{$\pm$ 1.20} & 90.30 \tiny{$\pm$ 1.31} & 88.58 \tiny{$\pm$ 0.20} & 80.85 \tiny{$\pm$ 5.50} & 89.47 \tiny{$\pm$ 0.77} & 83.17 \tiny{$\pm$ 0.75} & 85.34 \tiny{$\pm$ 0.49} \\
$10^{-11.75}$ & \cmark & 300 & 0.024 & - & - & - & - & - & - & - & 82.69 & 85.23 \\
$10^{-11.75}$ & \cmark & 150 & 0.024 & - & 92.95 & 78.90 & 88.64 & 87.78 & 82.41 & 87.16 & - & - \\
$10^{-11.75}$ & \cmark & 100 & 0.024 & - & 92.46 & 81.02 & 92.25 & 87.09 & 85.25 & 87.96 & - & - \\
$10^{-11.75}$ & \cmark & 0 & 0.024 & - & 89.06 & 82.05 & 90.02 & 88.64 & 74.36 & 87.84 & 83.82 & 85.58 \\
$10^{-11.75}$ & \cmark & - & 0.024 & 85.91 & 91.49 \tiny{$\pm$ 2.12} & 80.66 \tiny{$\pm$ 1.61} & 90.30 \tiny{$\pm$ 1.82} & 87.84 \tiny{$\pm$ 0.78} & 80.67 \tiny{$\pm$ 5.65} & 87.66 \tiny{$\pm$ 0.43} & 83.25 \tiny{$\pm$ 0.80} & 85.40 \tiny{$\pm$ 0.25} \\
$10^{-11.75}$ & \xmark & 300 & 0.0171 & - & - & - & - & - & - & - & 82.39 & 84.97 \\
$10^{-11.75}$ & \xmark & 150 & 0.0171 & - & 92.71 & 79.92 & 89.85 & 88.81 & 82.03 & 88.45 & - & - \\
$10^{-11.75}$ & \xmark & 100 & 0.0171 & - & 92.99 & 82.68 & 91.22 & 88.12 & 85.53 & 88.25 & - & - \\
$10^{-11.75}$ & \xmark & 0 & 0.0171 & - & 88.94 & 81.26 & 90.19 & 88.64 & 77.65 & 88.91 & 83.95 & 85.78 \\
$10^{-11.75}$ & \xmark & - & 0.0171 & 86.33 & 91.55 \tiny{$\pm$ 2.26} & 81.29 \tiny{$\pm$ 1.38} & 90.42 \tiny{$\pm$ 0.72} & 88.53 \tiny{$\pm$ 0.36} & 81.74 \tiny{$\pm$ 3.95} & 88.54 \tiny{$\pm$ 0.34} & 83.17 \tiny{$\pm$ 1.11} & 85.37 \tiny{$\pm$ 0.57} \\
$10^{-11.5}$ & \cmark & 300 & 0.029 & - & - & - & - & - & - & - & 82.08 & 84.43 \\
$10^{-11.5}$ & \cmark & 150 & 0.029 & - & 90.76 & 79.29 & 88.30 & 86.06 & 82.30 & 88.48 & - & - \\
$10^{-11.5}$ & \cmark & 100 & 0.029 & - & 92.71 & 79.92 & 90.53 & 87.78 & 84.49 & 87.99 & - & - \\
$10^{-11.5}$ & \cmark & 0 & 0.029 & - & 90.40 & 77.87 & 89.50 & 87.61 & 79.98 & 88.13 & 83.80 & 85.66 \\
$10^{-11.50}$ & \cmark & - & 0.029 & 85.67 & 91.29 \tiny{$\pm$ 1.24} & 79.03 \tiny{$\pm$ 1.05} & 89.44 \tiny{$\pm$ 1.12} & 87.15 \tiny{$\pm$ 0.95} & 82.26 \tiny{$\pm$ 2.26} & 88.20 \tiny{$\pm$ 0.25} & 82.94 \tiny{$\pm$ 1.22} & 85.05 \tiny{$\pm$ 0.87} \\
$10^{-11.5}$ & \xmark & 300 & 0.0198 & - & - & - & - & - & - & - & 82.00 & 84.66 \\
$10^{-11.5}$ & \xmark & 150 & 0.0198 & - & 91.69 & 75.98 & 89.67 & 88.64 & 80.22 & 88.71 & - & - \\
$10^{-11.5}$ & \xmark & 100 & 0.0198 & - & 92.02 & 81.50 & 91.91 & 89.33 & 78.14 & 88.46 & - & - \\
$10^{-11.5}$ & \xmark & 0 & 0.0198 & - & 89.67 & 71.42 & 87.44 & 86.75 & 78.90 & 89.20 & 83.60 & 85.63 \\
$10^{-11.50}$ & \xmark & - & 0.0198 & 85.15 & 91.13 \tiny{$\pm$ 1.27} & 76.30 \tiny{$\pm$ 5.05} & 89.67 \tiny{$\pm$ 2.24} & 88.24 \tiny{$\pm$ 1.34} & 79.09 \tiny{$\pm$ 1.05} & 88.79 \tiny{$\pm$ 0.38} & 82.80 \tiny{$\pm$ 1.13} & 85.15 \tiny{$\pm$ 0.68} \\
$10^{-11.25}$ & \cmark & 300 & 0.039 & - & - & - & - & - & - & - & 82.02 & 84.75 \\
$10^{-11.25}$ & \cmark & 150 & 0.039 & - & 89.34 & 78.11 & 89.16 & 87.44 & 81.40 & 88.36 & - & - \\
$10^{-11.25}$ & \cmark & 100 & 0.039 & - & 88.29 & 81.50 & 88.47 & 87.78 & 79.91 & 88.75 & - & - \\
$10^{-11.25}$ & \cmark & 0 & 0.039 & - & 90.60 & 74.02 & 86.23 & 87.44 & 79.91 & 88.41 & 82.29 & 85.00 \\
$10^{-11.25}$ & \cmark & - & 0.039 & 84.84 & 89.41 \tiny{$\pm$ 1.16} & 77.87 \tiny{$\pm$ 3.75} & 87.95 \tiny{$\pm$ 1.53} & 87.55 \tiny{$\pm$ 0.20} & 80.41 \tiny{$\pm$ 0.86} & 88.51 \tiny{$\pm$ 0.21} & 82.16 \tiny{$\pm$ 0.19} & 84.87 \tiny{$\pm$ 0.18} \\
$10^{-11.25}$ & \xmark & 300 & 0.0264 & - & - & - & - & - & - & - & 81.87 & 84.65 \\
$10^{-11.25}$ & \xmark & 150 & 0.0264 & - & 89.63 & 78.11 & 89.85 & 88.12 & 78.24 & 88.55 & - & - \\
$10^{-11.25}$ & \xmark & 100 & 0.0264 & - & 89.51 & 78.35 & 90.71 & 86.06 & 78.38 & 88.68 & - & - \\
$10^{-11.25}$ & \xmark & 0 & 0.0264 & - & 90.64 & 74.49 & 84.51 & 85.37 & 80.64 & 88.72 & 82.85 & 85.11 \\
$10^{-11.25}$ & \xmark & - & 0.0264 & 84.59 & 89.92 \tiny{$\pm$ 0.62} & 76.98 \tiny{$\pm$ 2.16} & 88.35 \tiny{$\pm$ 3.36} & 86.52 \tiny{$\pm$ 1.43} & 79.09 \tiny{$\pm$ 1.34} & 88.65 \tiny{$\pm$ 0.09} & 82.36 \tiny{$\pm$ 0.69} & 84.88 \tiny{$\pm$ 0.33} \\
$10^{-11}$ & \cmark & 300 & 0.0619 & - & - & - & - & - & - & - & 81.99 & 84.70 \\
$10^{-11}$ & \cmark & 150 & 0.0619 & - & 91.57 & 77.40 & 87.95 & 86.92 & 77.72 & 88.61 & - & - \\
$10^{-11}$ & \cmark & 100 & 0.0619 & - & 90.11 & 73.46 & 88.12 & 87.95 & 76.86 & 88.27 & - & - \\
$10^{-11}$ & \cmark & 0 & 0.0619 & - & 91.29 & 76.77 & 84.85 & 87.95 & 79.18 & 88.90 & 82.28 & 84.79 \\
$10^{-11}$ & \cmark & - & 0.0619 & 84.36 & 90.99 \tiny{$\pm$ 0.77} & 75.88 \tiny{$\pm$ 2.11} & 86.98 \tiny{$\pm$ 1.84} & 87.61 \tiny{$\pm$ 0.60} & 77.92 \tiny{$\pm$ 1.17} & 88.59 \tiny{$\pm$ 0.31} & 82.14 \tiny{$\pm$ 0.21} & 84.75 \tiny{$\pm$ 0.06} \\
$10^{-11}$ & \xmark & 300 & 0.0419 & - & - & - & - & - & - & - & 81.63 & 84.47 \\
$10^{-11}$ & \xmark & 150 & 0.0419 & - & 90.60 & 76.77 & 87.26 & 83.65 & 80.74 & 88.84 & - & - \\
$10^{-11}$ & \xmark & 100 & 0.0419 & - & 90.76 & 77.40 & 83.99 & 86.57 & 80.01 & 88.91 & - & - \\
$10^{-11}$ & \xmark & 0 & 0.0419 & - & 91.49 & 76.54 & 86.40 & 86.57 & 80.15 & 88.90 & 81.84 & 84.70 \\
$10^{-11}$ & \xmark & - & 0.0419 & 84.36 & 90.95 \tiny{$\pm$ 0.47} & 76.90 \tiny{$\pm$ 0.45} & 85.89 \tiny{$\pm$ 1.70} & 85.60 \tiny{$\pm$ 1.69} & 80.30 \tiny{$\pm$ 0.39} & 88.88 \tiny{$\pm$ 0.04} & 81.74 \tiny{$\pm$ 0.15} & 84.58 \tiny{$\pm$ 0.16} \\
$10^{-10.75}$ & \cmark & 300 & 0.0955 & - & - & - & - & - & - & - & 81.65 & 84.53 \\
$10^{-10.75}$ & \cmark & 150 & 0.0955 & - & 91.00 & 75.04 & 85.54 & 86.23 & 78.31 & 89.02 & - & - \\
$10^{-10.75}$ & \cmark & 100 & 0.0955 & - & 91.13 & 78.82 & 88.64 & 86.92 & 77.41 & 89.09 & - & - \\
$10^{-10.75}$ & \cmark & 0 & 0.0955 & - & 91.65 & 77.80 & 85.89 & 87.61 & 80.01 & 89.03 & 81.67 & 84.89 \\
$10^{-10.75}$ & \cmark & - & 0.0955 & 84.51 & 91.26 \tiny{$\pm$ 0.34} & 77.22 \tiny{$\pm$ 1.95} & 86.69 \tiny{$\pm$ 1.70} & 86.92 \tiny{$\pm$ 0.69} & 78.58 \tiny{$\pm$ 1.32} & 89.04 \tiny{$\pm$ 0.04} & 81.66 \tiny{$\pm$ 0.02} & 84.71 \tiny{$\pm$ 0.25} \\
$10^{-10.75}$ & \xmark & 300 & 0.0672 & - & - & - & - & - & - & - & 81.73 & 84.77 \\
$10^{-10.75}$ & \xmark & 150 & 0.0672 & - & 91.17 & 76.69 & 85.71 & 86.57 & 81.02 & 89.68 & - & - \\
$10^{-10.75}$ & \xmark & 100 & 0.0672 & - & 92.06 & 78.74 & 86.75 & 88.12 & 81.26 & 90.17 & - & - \\
$10^{-10.75}$ & \xmark & 0 & 0.0672 & - & 91.29 & 77.64 & 87.95 & 86.92 & 80.08 & 89.07 & 81.97 & 84.95 \\
$10^{-10.75}$ & \xmark & - & 0.0672 & 85.04 & 91.50 \tiny{$\pm$ 0.48} & 77.69 \tiny{$\pm$ 1.02} & 86.80 \tiny{$\pm$ 1.12} & 87.21 \tiny{$\pm$ 0.81} & 80.79 \tiny{$\pm$ 0.62} & 89.64 \tiny{$\pm$ 0.55} & 81.85 \tiny{$\pm$ 0.17} & 84.86 \tiny{$\pm$ 0.13} \\
$10^{-10.5}$ & \cmark & 300 & 0.1013 & - & - & - & - & - & - & - & 81.73 & 84.52 \\
$10^{-10.5}$ & \cmark & 150 & 0.1013 & - & 91.57 & 77.09 & 84.51 & 84.34 & 79.01 & 88.95 & - & - \\
$10^{-10.5}$ & \cmark & 100 & 0.1013 & - & 91.17 & 78.11 & 84.68 & 85.37 & 78.73 & 89.35 & - & - \\
$10^{-10.5}$ & \cmark & 0 & 0.1013 & - & 90.92 & 77.80 & 85.54 & 84.51 & 70.37 & 89.26 & 82.13 & 84.87 \\
$10^{-10.50}$ & \cmark & - & 0.1013 & 83.80 & 91.22 \tiny{$\pm$ 0.33} & 77.66 \tiny{$\pm$ 0.52} & 84.91 \tiny{$\pm$ 0.55} & 84.74 \tiny{$\pm$ 0.55} & 76.04 \tiny{$\pm$ 4.91} & 89.19 \tiny{$\pm$ 0.21} & 81.93 \tiny{$\pm$ 0.29} & 84.70 \tiny{$\pm$ 0.25} \\
$10^{-10.5}$ & \xmark & 300 & 0.0753 & - & - & - & - & - & - & - & 82.27 & 85.00 \\
$10^{-10.5}$ & \xmark & 150 & 0.0753 & - & 91.21 & 76.61 & 87.95 & 86.75 & 80.33 & 89.88 & - & - \\
$10^{-10.5}$ & \xmark & 100 & 0.0753 & - & 91.53 & 77.56 & 87.09 & 87.61 & 80.88 & 89.87 & - & - \\
$10^{-10.5}$ & \xmark & 0 & 0.0753 & - & 90.72 & 77.09 & 86.40 & 87.09 & 78.45 & 89.16 & 81.88 & 84.79 \\
$10^{-10.50}$ & \xmark & - & 0.0753 & 84.88 & 91.15 \tiny{$\pm$ 0.41} & 77.09 \tiny{$\pm$ 0.47} & 87.15 \tiny{$\pm$ 0.78} & 87.15 \tiny{$\pm$ 0.43} & 79.89 \tiny{$\pm$ 1.27} & 89.64 \tiny{$\pm$ 0.41} & 82.08 \tiny{$\pm$ 0.27} & 84.89 \tiny{$\pm$ 0.15} \\
$10^{-10.25}$ & \cmark & 300 & 0.0996 & - & - & - & - & - & - & - & 81.80 & 84.53 \\
$10^{-10.25}$ & \cmark & 150 & 0.0996 & - & 91.00 & 76.30 & 83.48 & 84.68 & 79.35 & 89.31 & - & - \\
$10^{-10.25}$ & \cmark & 100 & 0.0996 & - & 90.96 & 77.80 & 84.51 & 84.85 & 79.15 & 89.29 & - & - \\
$10^{-10.25}$ & \cmark & 0 & 0.0996 & - & 91.13 & 77.40 & 83.82 & 84.51 & 79.04 & 89.08 & 82.36 & 84.85 \\
$10^{-10.25}$ & \cmark & - & 0.0996 & 84.00 & 91.03 \tiny{$\pm$ 0.08} & 77.17 \tiny{$\pm$ 0.78} & 83.94 \tiny{$\pm$ 0.53} & 84.68 \tiny{$\pm$ 0.17} & 79.18 \tiny{$\pm$ 0.16} & 89.22 \tiny{$\pm$ 0.13} & 82.08 \tiny{$\pm$ 0.39} & 84.69 \tiny{$\pm$ 0.23} \\
$10^{-10.25}$ & \xmark & 300 & 0.0754 & - & - & - & - & - & - & - & 82.37 & 85.01 \\
$10^{-10.25}$ & \xmark & 150 & 0.0754 & - & 91.90 & 76.93 & 85.37 & 86.75 & 80.57 & 89.89 & - & - \\
$10^{-10.25}$ & \xmark & 100 & 0.0754 & - & 91.53 & 77.40 & 85.20 & 85.54 & 80.88 & 89.98 & - & - \\
$10^{-10.25}$ & \xmark & 0 & 0.0754 & - & 91.37 & 76.46 & 86.06 & 86.06 & 80.95 & 89.79 & 82.49 & 84.91 \\
$10^{-10.25}$ & \xmark & - & 0.0754 & 84.78 & 91.60 \tiny{$\pm$ 0.27} & 76.93 \tiny{$\pm$ 0.47} & 85.54 \tiny{$\pm$ 0.46} & 86.12 \tiny{$\pm$ 0.60} & 80.80 \tiny{$\pm$ 0.20} & 89.89 \tiny{$\pm$ 0.10} & 82.43 \tiny{$\pm$ 0.08} & 84.96 \tiny{$\pm$ 0.08} \\
$10^{-10}$ & \cmark & 300 & 0.0981 & - & - & - & - & - & - & - & 82.04 & 84.68 \\
$10^{-10}$ & \cmark & 150 & 0.0981 & - & 91.41 & 76.69 & 83.65 & 85.37 & 79.49 & 89.41 & - & - \\
$10^{-10}$ & \cmark & 100 & 0.0981 & - & 91.33 & 76.46 & 83.99 & 84.85 & 79.25 & 89.58 & - & - \\
$10^{-10}$ & \cmark & 0 & 0.0981 & - & 90.56 & 75.75 & 83.48 & 84.85 & 78.73 & 88.97 & 82.22 & 84.67 \\
$10^{-10}$ & \cmark & - & 0.0981 & 83.93 & 91.10 \tiny{$\pm$ 0.47} & 76.30 \tiny{$\pm$ 0.49} & 83.71 \tiny{$\pm$ 0.26} & 85.03 \tiny{$\pm$ 0.30} & 79.16 \tiny{$\pm$ 0.39} & 89.32 \tiny{$\pm$ 0.31} & 82.13 \tiny{$\pm$ 0.13} & 84.67 \tiny{$\pm$ 0.01} \\
$10^{-10}$ & \xmark & 300 & 0.0763 & - & - & - & - & - & - & - & 82.05 & 84.70 \\
$10^{-10}$ & \xmark & 150 & 0.0763 & - & 91.41 & 77.64 & 86.75 & 85.03 & 80.43 & 89.68 & - & - \\
$10^{-10}$ & \xmark & 100 & 0.0763 & - & 91.13 & 77.87 & 85.20 & 86.75 & 80.71 & 89.95 & - & - \\
$10^{-10}$ & \xmark & 0 & 0.0763 & - & 91.45 & 76.77 & 85.03 & 86.57 & 78.66 & 89.46 & 82.18 & 84.67 \\
$10^{-10}$ & \xmark & - & 0.0763 & 84.62 & 91.33 \tiny{$\pm$ 0.18} & 77.43 \tiny{$\pm$ 0.58} & 85.66 \tiny{$\pm$ 0.95} & 86.12 \tiny{$\pm$ 0.95} & 79.93 \tiny{$\pm$ 1.11} & 89.70 \tiny{$\pm$ 0.25} & 82.12 \tiny{$\pm$ 0.09} & 84.69 \tiny{$\pm$ 0.02} \\
$10^{-9}$ & \cmark & 300 & 0.0984 & - & - & - & - & - & - & - & 82.20 & 84.56 \\
$10^{-9}$ & \cmark & 150 & 0.0984 & - & 91.21 & 77.09 & 85.54 & 85.54 & 79.98 & 89.26 & - & - \\
$10^{-9}$ & \cmark & 100 & 0.0984 & - & 91.17 & 77.32 & 85.37 & 84.85 & 78.80 & 89.43 & - & - \\
$10^{-9}$ & \cmark & 0 & 0.0984 & - & 91.29 & 77.32 & 85.54 & 84.17 & 79.49 & 89.00 & 82.02 & 84.71 \\
$10^{-9}$ & \cmark & - & 0.0984 & 84.27 & 91.22 \tiny{$\pm$ 0.06} & 77.24 \tiny{$\pm$ 0.14} & 85.48 \tiny{$\pm$ 0.10} & 84.85 \tiny{$\pm$ 0.69} & 79.42 \tiny{$\pm$ 0.59} & 89.23 \tiny{$\pm$ 0.22} & 82.11 \tiny{$\pm$ 0.12} & 84.64 \tiny{$\pm$ 0.10} \\
$10^{-9}$ & \xmark & 300 & 0.0774 & - & - & - & - & - & - & - & 82.23 & 84.75 \\
$10^{-9}$ & \xmark & 150 & 0.0774 & - & 91.29 & 76.85 & 85.20 & 86.57 & 80.81 & 89.96 & - & - \\
$10^{-9}$ & \xmark & 100 & 0.0774 & - & 91.09 & 77.32 & 86.06 & 86.06 & 80.53 & 89.79 & - & - \\
$10^{-9}$ & \xmark & 0 & 0.0774 & - & 90.60 & 77.80 & 85.71 & 86.40 & 80.50 & 89.64 & 82.02 & 84.82 \\
$10^{-9}$ & \xmark & - & 0.0774 & 84.71 & 90.99 \tiny{$\pm$ 0.35} & 77.32 \tiny{$\pm$ 0.47} & 85.66 \tiny{$\pm$ 0.43} & 86.35 \tiny{$\pm$ 0.26} & 80.62 \tiny{$\pm$ 0.17} & 89.80 \tiny{$\pm$ 0.16} & 82.13 \tiny{$\pm$ 0.15} & 84.78 \tiny{$\pm$ 0.05} \\
$10^{-8}$ & \cmark & 300 & 0.0987 & - & - & - & - & - & - & - & 81.73 & 84.62 \\
$10^{-8}$ & \cmark & 150 & 0.0987 & - & 91.41 & 78.43 & 84.34 & 85.37 & 79.94 & 89.53 & - & - \\
$10^{-8}$ & \cmark & 100 & 0.0987 & - & 91.13 & 77.01 & 83.48 & 85.54 & 80.19 & 89.65 & - & - \\
$10^{-8}$ & \cmark & 0 & 0.0987 & - & 91.00 & 77.64 & 83.48 & 85.37 & 79.04 & 89.26 & 81.71 & 84.81 \\
$10^{-8}$ & \cmark & - & 0.0987 & 84.21 & 91.18 \tiny{$\pm$ 0.21} & 77.69 \tiny{$\pm$ 0.71} & 83.76 \tiny{$\pm$ 0.50} & 85.43 \tiny{$\pm$ 0.10} & 79.72 \tiny{$\pm$ 0.60} & 89.48 \tiny{$\pm$ 0.20} & 81.72 \tiny{$\pm$ 0.01} & 84.72 \tiny{$\pm$ 0.13} \\
$10^{-8}$ & \xmark & 300 & 0.0760 & - & - & - & - & - & - & - & 82.35 & 84.79 \\
$10^{-8}$ & \xmark & 150 & 0.0760 & - & 91.09 & 76.22 & 86.75 & 85.37 & 80.46 & 89.72 & - & - \\
$10^{-8}$ & \xmark & 100 & 0.0760 & - & 91.61 & 77.32 & 86.06 & 86.92 & 80.29 & 89.83 & - & - \\
$10^{-8}$ & \xmark & 0 & 0.0760 & - & 91.45 & 76.46 & 86.23 & 89.16 & 80.60 & 89.58 & 82.26 & 84.94 \\
$10^{-8}$ & \xmark & - & 0.0760 & 84.86 & 91.38 \tiny{$\pm$ 0.27} & 76.67 \tiny{$\pm$ 0.58} & 86.35 \tiny{$\pm$ 0.36} & 87.15 \tiny{$\pm$ 1.90} & 80.45 \tiny{$\pm$ 0.16} & 89.71 \tiny{$\pm$ 0.12} & 82.30 \tiny{$\pm$ 0.06} & 84.87 \tiny{$\pm$ 0.11} \\
$10^{-7}$ & \cmark & 300 & 0.0999 & - & - & - & - & - & - & - & 81.95 & 84.72 \\
$10^{-7}$ & \cmark & 150 & 0.0999 & - & 91.33 & 76.61 & 85.03 & 84.34 & 79.15 & 89.26 & - & - \\
$10^{-7}$ & \cmark & 100 & 0.0999 & - & 91.29 & 76.77 & 84.51 & 84.51 & 80.15 & 89.44 & - & - \\
$10^{-7}$ & \cmark & 0 & 0.0999 & - & 90.88 & 76.93 & 83.99 & 84.51 & 78.70 & 89.00 & 81.91 & 84.91 \\
$10^{-7}$ & \cmark & - & 0.0999 & 84.03 & 91.17 \tiny{$\pm$ 0.25} & 76.77 \tiny{$\pm$ 0.16} & 84.51 \tiny{$\pm$ 0.52} & 84.45 \tiny{$\pm$ 0.10} & 79.33 \tiny{$\pm$ 0.75} & 89.23 \tiny{$\pm$ 0.22} & 81.93 \tiny{$\pm$ 0.03} & 84.81 \tiny{$\pm$ 0.13} \\
$10^{-7}$ & \xmark & 300 & 0.0749 & - & - & - & - & - & - & - & 82.05 & 84.89 \\
$10^{-7}$ & \xmark & 150 & 0.0749 & - & 91.45 & 77.17 & 85.89 & 85.89 & 80.40 & 89.97 & - & - \\
$10^{-7}$ & \xmark & 100 & 0.0749 & - & 91.61 & 77.48 & 85.89 & 86.57 & 80.71 & 90.00 & - & - \\
$10^{-7}$ & \xmark & 0 & 0.0749 & - & 91.53 & 76.54 & 85.54 & 85.71 & 80.81 & 89.72 & 81.88 & 84.89 \\
$10^{-7}$ & \xmark & - & 0.0749 & 84.73 & 91.53 \tiny{$\pm$ 0.08} & 77.06 \tiny{$\pm$ 0.48} & 85.77 \tiny{$\pm$ 0.20} & 86.06 \tiny{$\pm$ 0.46} & 80.64 \tiny{$\pm$ 0.22} & 89.90 \tiny{$\pm$ 0.15} & 81.96 \tiny{$\pm$ 0.12} & 84.89 \tiny{$\pm$ 0.00} \\
$10^{-6}$ & \cmark & 300 & 0.0997 & - & - & - & - & - & - & - & 81.67 & 84.71 \\
$10^{-6}$ & \cmark & 150 & 0.0997 & - & 91.37 & 76.54 & 83.99 & 87.61 & 78.94 & 89.68 & - & - \\
$10^{-6}$ & \cmark & 100 & 0.0997 & - & 90.84 & 77.01 & 84.34 & 85.71 & 78.87 & 89.55 & - & - \\
$10^{-6}$ & \cmark & 0 & 0.0997 & - & 90.92 & 76.14 & 83.65 & 85.54 & 78.94 & 89.20 & 82.37 & 84.93 \\
$10^{-6}$ & \cmark & - & 0.0997 & 84.14 & 91.05 \tiny{$\pm$ 0.28} & 76.56 \tiny{$\pm$ 0.43} & 83.99 \tiny{$\pm$ 0.34} & 86.29 \tiny{$\pm$ 1.15} & 78.92 \tiny{$\pm$ 0.04} & 89.48 \tiny{$\pm$ 0.25} & 82.02 \tiny{$\pm$ 0.49} & 84.82 \tiny{$\pm$ 0.15} \\
$10^{-6}$ & \xmark & 300 & 0.0765 & - & - & - & - & - & - & - & 82.28 & 84.80 \\
$10^{-6}$ & \xmark & 150 & 0.0765 & - & 91.21 & 77.24 & 85.89 & 85.20 & 80.67 & 89.95 & - & - \\
$10^{-6}$ & \xmark & 100 & 0.0765 & - & 91.61 & 77.01 & 85.37 & 86.06 & 81.05 & 90.03 & - & - \\
$10^{-6}$ & \xmark & 0 & 0.0765 & - & 90.72 & 76.77 & 85.54 & 86.06 & 80.43 & 89.51 & 82.05 & 84.88 \\
$10^{-6}$ & \xmark & - & 0.0765 & 84.64 & 91.18 \tiny{$\pm$ 0.45} & 77.01 \tiny{$\pm$ 0.24} & 85.60 \tiny{$\pm$ 0.26} & 85.77 \tiny{$\pm$ 0.50} & 80.72 \tiny{$\pm$ 0.31} & 89.83 \tiny{$\pm$ 0.28} & 82.16 \tiny{$\pm$ 0.17} & 84.84 \tiny{$\pm$ 0.05} \\
$10^{-5}$ & \cmark & 300 & 0.1000 & - & - & - & - & - & - & - & 81.62 & 84.64 \\
$10^{-5}$ & \cmark & 150 & 0.1000 & - & 91.37 & 77.32 & 83.65 & 85.89 & 79.39 & 89.67 & - & - \\
$10^{-5}$ & \cmark & 100 & 0.1000 & - & 90.96 & 76.77 & 83.65 & 83.82 & 79.01 & 89.68 & - & - \\
$10^{-5}$ & \cmark & 0 & 0.1000 & - & 91.37 & 76.14 & 83.99 & 85.29 & 78.30 & 89.13 & 82.02 & 84.88 \\
$10^{-5}$ & \cmark & - & 0.1000 & 83.99 & 91.23 \tiny{$\pm$ 0.23} & 76.75 \tiny{$\pm$ 0.59} & 83.76 \tiny{$\pm$ 0.20} & 85.20 \tiny{$\pm$ 1.19} & 78.93 \tiny{$\pm$ 0.51} & 89.49 \tiny{$\pm$ 0.31} & 81.82 \tiny{$\pm$ 0.28} & 84.76 \tiny{$\pm$ 0.17} \\
$10^{-5}$ & \xmark & 300 & 0.0756 & - & - & - & - & - & - & - & 82.25 & 84.86 \\
$10^{-5}$ & \xmark & 150 & 0.0756 & - & 91.90 & 77.17 & 86.92 & 86.57 & 80.81 & 89.92 & - & - \\
$10^{-5}$ & \xmark & 100 & 0.0756 & - & 91.73 & 77.24 & 86.57 & 86.50 & 81.16 & 89.91 & - & - \\
$10^{-5}$ & \xmark & 0 & 0.0756 & - & 91.69 & 77.24 & 85.54 & 86.75 & 80.71 & 89.66 & 81.96 & 84.85 \\
$10^{-5}$ & \xmark & - & 0.0756 & 84.96 & 91.77 \tiny{$\pm$ 0.11} & 77.22 \tiny{$\pm$ 0.05} & 86.35 \tiny{$\pm$ 0.72} & 86.69 \tiny{$\pm$ 0.10} & 80.89 \tiny{$\pm$ 0.24} & 89.83 \tiny{$\pm$ 0.15} & 82.10 \tiny{$\pm$ 0.20} & 84.85 \tiny{$\pm$ 0.01} \\
$10^{-4}$ & \cmark & 300 & 0.1000 & - & - & - & - & - & - & - & 82.03 & 84.63 \\
$10^{-4}$ & \cmark & 150 & 0.1000 & - & 91.21 & 78.03 & 82.44 & 85.89 & 78.70 & 89.57 & - & - \\
$10^{-4}$ & \cmark & 100 & 0.1000 & - & 90.84 & 76.69 & 84.51 & 85.89 & 78.73 & 89.50 & - & - \\
$10^{-4}$ & \cmark & 0 & 0.1000 & - & 91.05 & 77.87 & 83.48 & 85.54 & 78.49 & 89.20 & 81.95 & 85.03 \\
$10^{-4}$ & \cmark & - & 0.1000 & 84.09 & 91.03 \tiny{$\pm$ 0.18} & 77.53 \tiny{$\pm$ 0.73} & 83.48 \tiny{$\pm$ 1.03} & 85.77 \tiny{$\pm$ 0.20} & 78.64 \tiny{$\pm$ 0.13} & 89.42 \tiny{$\pm$ 0.20} & 81.99 \tiny{$\pm$ 0.06} & 84.83 \tiny{$\pm$ 0.28} \\
$10^{-4}$ & \xmark & 300 & 0.0757 & - & - & - & - & - & - & - & 82.29 & 84.91 \\
$10^{-4}$ & \xmark & 150 & 0.0757 & - & 91.45 & 78.35 & 86.23 & 86.75 & 80.26 & 89.81 & - & - \\
$10^{-4}$ & \xmark & 100 & 0.0757 & - & 91.37 & 77.64 & 86.75 & 85.37 & 81.33 & 90.06 & - & - \\
$10^{-4}$ & \xmark & 0 & 0.0757 & - & 91.73 & 77.40 & 87.44 & 85.20 & 80.64 & 89.81 & 82.49 & 85.05 \\
$10^{-4}$ & \xmark & - & 0.0757 & 84.99 & 91.52 \tiny{$\pm$ 0.19} & 77.80 \tiny{$\pm$ 0.49} & 86.80 \tiny{$\pm$ 0.60} & 85.77 \tiny{$\pm$ 0.85} & 80.74 \tiny{$\pm$ 0.55} & 89.89 \tiny{$\pm$ 0.15} & 82.39 \tiny{$\pm$ 0.14} & 84.98 \tiny{$\pm$ 0.10} \\
$10^{-3}$ & \cmark & 300 & 0.1000 & - & - & - & - & - & - & - & 81.81 & 84.69 \\
$10^{-3}$ & \cmark & 150 & 0.1000 & - & 91.25 & 77.01 & 85.37 & 84.51 & 79.46 & 89.40 & - & - \\
$10^{-3}$ & \cmark & 100 & 0.1000 & - & 91.41 & 76.85 & 84.34 & 86.06 & 79.35 & 89.60 & - & - \\
$10^{-3}$ & \cmark & 0 & 0.1000 & - & 91.05 & 77.87 & 83.48 & 85.54 & 79.01 & 89.00 & 82.04 & 85.06 \\
$10^{-3}$ & \cmark & - & 0.1000 & 84.20 & 91.23 \tiny{$\pm$ 0.18} & 77.24 \tiny{$\pm$ 0.55} & 84.39 \tiny{$\pm$ 0.95} & 85.37 \tiny{$\pm$ 0.79} & 79.27 \tiny{$\pm$ 0.24} & 89.33 \tiny{$\pm$ 0.30} & 81.93 \tiny{$\pm$ 0.16} & 84.87 \tiny{$\pm$ 0.26} \\
$10^{-3}$ & \xmark & 300 & 0.0761 & - & - & - & - & - & - & - & 82.42 & 84.89 \\
$10^{-3}$ & \xmark & 150 & 0.0761 & - & 92.02 & 77.87 & 86.23 & 87.09 & 80.81 & 90.03 & - & - \\
$10^{-3}$ & \xmark & 100 & 0.0761 & - & 91.33 & 78.19 & 85.20 & 87.26 & 81.92 & 90.00 & - & - \\
$10^{-3}$ & \xmark & 0 & 0.0761 & - & 91.61 & 76.61 & 86.06 & 85.54 & 80.01 & 89.79 & 81.90 & 84.87 \\
$10^{-3}$ & \xmark & - & 0.0761 & 84.95 & 91.65 \tiny{$\pm$ 0.35} & 77.56 \tiny{$\pm$ 0.83} & 85.83 \tiny{$\pm$ 0.55} & 86.63 \tiny{$\pm$ 0.95} & 80.92 \tiny{$\pm$ 0.96} & 89.94 \tiny{$\pm$ 0.13} & 82.16 \tiny{$\pm$ 0.37} & 84.88 \tiny{$\pm$ 0.01} \\
$10^{-2}$ & \cmark & 300 & 0.1000 & - & - & - & - & - & - & - & 82.14 & 84.69 \\
$10^{-2}$ & \cmark & 150 & 0.1000 & - & 91.00 & 77.24 & 85.20 & 85.37 & 78.80 & 89.51 & - & - \\
$10^{-2}$ & \cmark & 100 & 0.1000 & - & 91.09 & 77.40 & 83.99 & 86.23 & 79.77 & 89.59 & - & - \\
$10^{-2}$ & \cmark & 0 & 0.1000 & - & 90.96 & 76.46 & 83.48 & 84.51 & 79.46 & 89.03 & 82.20 & 84.92 \\
$10^{-2}$ & \cmark & - & 0.1000 & 84.17 & 91.02 \tiny{$\pm$ 0.06} & 77.03 \tiny{$\pm$ 0.51} & 84.22 \tiny{$\pm$ 0.88} & 85.37 \tiny{$\pm$ 0.86} & 79.34 \tiny{$\pm$ 0.50} & 89.38 \tiny{$\pm$ 0.30} & 82.17 \tiny{$\pm$ 0.05} & 84.81 \tiny{$\pm$ 0.16} \\
$10^{-2}$ & \xmark & 300 & 0.0759 & - & - & - & - & - & - & - & 82.31 & 85.04 \\
$10^{-2}$ & \xmark & 150 & 0.0759 & - & 91.77 & 77.17 & 85.03 & 85.37 & 80.43 & 89.94 & - & - \\
$10^{-2}$ & \xmark & 100 & 0.0759 & - & 91.65 & 77.95 & 85.71 & 85.89 & 82.10 & 90.28 & - & - \\
$10^{-2}$ & \xmark & 0 & 0.0759 & - & 91.25 & 76.93 & 85.37 & 86.75 & 80.50 & 89.69 & 81.92 & 84.97 \\
$10^{-2}$ & \xmark & - & 0.0759 & 84.80 & 91.56 \tiny{$\pm$ 0.28} & 77.35 \tiny{$\pm$ 0.54} & 85.37 \tiny{$\pm$ 0.34} & 86.00 \tiny{$\pm$ 0.70} & 81.01 \tiny{$\pm$ 0.94} & 89.97 \tiny{$\pm$ 0.30} & 82.11 \tiny{$\pm$ 0.27} & 85.00 \tiny{$\pm$ 0.05} \\
$10^{-1}$ & \cmark & 300 & 0.1000 & - & - & - & - & - & - & - & 81.95 & 84.78 \\
$10^{-1}$ & \cmark & 150 & 0.1000 & - & 91.37 & 77.01 & 85.03 & 84.85 & 78.87 & 89.58 & - & - \\
$10^{-1}$ & \cmark & 100 & 0.1000 & - & 91.45 & 77.56 & 86.06 & 84.85 & 79.98 & 89.62 & - & - \\
$10^{-1}$ & \cmark & 0 & 0.1000 & - & 90.84 & 76.61 & 85.37 & 85.71 & 79.39 & 89.36 & 82.04 & 85.04 \\
$10^{-1}$ & \cmark & - & 0.1000 & 84.34 & 91.22 \tiny{$\pm$ 0.33} & 77.06 \tiny{$\pm$ 0.47} & 85.48 \tiny{$\pm$ 0.53} & 85.14 \tiny{$\pm$ 0.50} & 79.41 \tiny{$\pm$ 0.56} & 89.52 \tiny{$\pm$ 0.14} & 82.00 \tiny{$\pm$ 0.06} & 84.91 \tiny{$\pm$ 0.18} \\
$10^{-1}$ & \xmark & 300 & 0.0753 & - & - & - & - & - & - & - & 82.28 & 85.02 \\
$10^{-1}$ & \xmark & 150 & 0.0753 & - & 91.41 & 78.11 & 86.23 & 86.92 & 80.99 & 89.76 & - & - \\
$10^{-1}$ & \xmark & 100 & 0.0753 & - & 91.37 & 77.24 & 85.20 & 87.09 & 81.33 & 90.13 & - & - \\
$10^{-1}$ & \xmark & 0 & 0.0753 & - & 91.65 & 77.24 & 86.57 & 86.92 & 80.67 & 89.89 & 82.13 & 84.80 \\
$10^{-1}$ & \xmark & - & 0.0753 & 85.00 & 91.48 \tiny{$\pm$ 0.15} & 77.53 \tiny{$\pm$ 0.50} & 86.00 \tiny{$\pm$ 0.72} & 86.98 \tiny{$\pm$ 0.10} & 81.00 \tiny{$\pm$ 0.33} & 89.93 \tiny{$\pm$ 0.19} & 82.20 \tiny{$\pm$ 0.10} & 84.91 \tiny{$\pm$ 0.16} \\
$1$ & \cmark & 300 & 0.0997 & - & - & - & - & - & - & - & 81.79 & 84.76 \\
$1$ & \cmark & 150 & 0.0997 & - & 92.10 & 77.48 & 85.03 & 85.71 & 79.04 & 89.05 & - & - \\
$1$ & \cmark & 100 & 0.0997 & - & 91.17 & 77.09 & 83.48 & 84.68 & 79.98 & 89.67 & - & - \\
$1$ & \cmark & 0 & 0.0997 & - & 91.57 & 77.01 & 83.13 & 86.23 & 78.94 & 88.81 & 81.95 & 84.85 \\
$1$ & \cmark & - & 0.0997 & 84.17 & 91.61 \tiny{$\pm$ 0.47} & 77.19 \tiny{$\pm$ 0.25} & 83.88 \tiny{$\pm$ 1.01} & 85.54 \tiny{$\pm$ 0.79} & 79.32 \tiny{$\pm$ 0.57} & 89.17 \tiny{$\pm$ 0.44} & 81.87 \tiny{$\pm$ 0.11} & 84.80 \tiny{$\pm$ 0.06} \\
$1$ & \xmark & 300 & 0.0765 & - & - & - & - & - & - & - & 82.19 & 84.89 \\
$1$ & \xmark & 150 & 0.0765 & - & 91.65 & 78.11 & 86.57 & 86.57 & 80.33 & 90.00 & - & - \\
$1$ & \xmark & 100 & 0.0765 & - & 91.57 & 77.87 & 85.71 & 86.23 & 81.40 & 89.89 & - & - \\
$1$ & \xmark & 0 & 0.0765 & - & 91.29 & 77.80 & 85.54 & 86.40 & 80.85 & 89.42 & 81.95 & 84.86 \\
$1$ & \xmark & - & 0.0765 & 84.92 & 91.50 \tiny{$\pm$ 0.19} & 77.93 \tiny{$\pm$ 0.16} & 85.94 \tiny{$\pm$ 0.55} & 86.40 \tiny{$\pm$ 0.17} & 80.86 \tiny{$\pm$ 0.54} & 89.77 \tiny{$\pm$ 0.31} & 82.07 \tiny{$\pm$ 0.17} & 84.88 \tiny{$\pm$ 0.02} \\
$2$ & \cmark & 300 & 0.0992 & - & - & - & - & - & - & - & 82.07 & 84.78 \\
$2$ & \cmark & 150 & 0.0992 & - & 91.17 & 76.69 & 84.51 & 83.13 & 79.11 & 89.41 & - & - \\
$2$ & \cmark & 100 & 0.0992 & - & 90.96 & 77.09 & 83.13 & 84.51 & 79.46 & 89.51 & - & - \\
$2$ & \cmark & 0 & 0.0992 & - & 91.17 & 77.17 & 83.48 & 83.48 & 78.49 & 89.07 & 81.93 & 84.65 \\
$2$ & \cmark & - & 0.0992 & 83.82 & 91.10 \tiny{$\pm$ 0.12} & 76.98 \tiny{$\pm$ 0.25} & 83.71 \tiny{$\pm$ 0.72} & 83.71 \tiny{$\pm$ 0.72} & 79.02 \tiny{$\pm$ 0.49} & 89.33 \tiny{$\pm$ 0.23} & 82.00 \tiny{$\pm$ 0.11} & 84.71 \tiny{$\pm$ 0.09} \\
$2$ & \xmark & 300 & 0.0770 & - & - & - & - & - & - & - & 82.17 & 84.74 \\
$2$ & \xmark & 150 & 0.0770 & - & 92.22 & 77.01 & 84.51 & 86.57 & 80.19 & 89.82 & - & - \\
$2$ & \xmark & 100 & 0.0770 & - & 91.49 & 77.01 & 86.23 & 85.89 & 80.74 & 89.93 & - & - \\
$2$ & \xmark & 0 & 0.0770 & - & 91.77 & 77.64 & 85.20 & 86.40 & 81.05 & 89.78 & 82.05 & 84.70 \\
$2$ & \xmark & - & 0.0770 & 84.75 & 91.83 \tiny{$\pm$ 0.37} & 77.22 \tiny{$\pm$ 0.36} & 85.31 \tiny{$\pm$ 0.87} & 86.29 \tiny{$\pm$ 0.36} & 80.66 \tiny{$\pm$ 0.44} & 89.84 \tiny{$\pm$ 0.08} & 82.11 \tiny{$\pm$ 0.08} & 84.72 \tiny{$\pm$ 0.03} \\
\bottomrule

\caption{\textbf{Detailed finetuning results of \paper-Gate.} All presented results are finetunings from a frozen or unfrozen backbone during distillation using 16 supertokens. During distillation and finetunings, no weight decay is applied on the gate module. For the classification task on ModelNet40 (MN40), OmniObject3D (OO3D), the three splits of ScanObjectNN (SONN-\{\texttt{PB-T50-RS}, \texttt{OBJ-BG}, \texttt{OBJ-ONLY}\}) and ShapeNet55 (SN55), the used metric is the top-1 accuracy on the validation set. For part segmentation with ShapeNetPart (SNP), we show for both category and instance mIoUs. Mean $\pm$ Std Dev over 10 runs for the baseline, 3 for the classification task, and 2 for the part segmentation task.}
\label{tab:finetunings-v3-detailed}
\end{longtable}
}

%% file: tables/correlation-v3.tex
\begin{table*}[ht!]
\resizebox{\linewidth}{!}{
\begin{tabular}{l|cccccccc}
\toprule
Frozen student & MN40 & OO3D & SONN-\texttt{OBJ-BG} & SONN-\texttt{OBJ-ONLY} & SONN-\texttt{PB-T50-RS} & SN55 & \multicolumn{2}{c}{SNP} \\
 & OA@1 & OA@1 & OA@1 & OA@1 & OA@1 & OA@1 & $\text{mIoU}_C$ & $\text{mIoU}_I$ \\
\midrule
\cmark & -0.4542 & -0.6583 & -0.5198 & -0.8024 & -0.6953 & -0.0257 & -0.6630 & -0.4397 \\
\xmark & -0.1339 & -0.2451 & -0.0934 & -0.7556 & -0.6225 & 0.1948 & -0.5702 & -0.5099 \\
All & -0.1897 & -0.5278 & -0.3330 & -0.8103 & -0.7359 & 0.1472 & -0.5741 & -0.4091 \\
\bottomrule
\end{tabular}
}
\caption{\textbf{Correlation between downstream performances and gate regularization of \paper-Gate.} All presented results are finetunings from a frozen or unfrozen backbone during distillation using 16 supertokens. We compute the Spearman correlation on ModelNet40 (MN40), OmniObject3D (OO3D), the three splits of ScanObjectNN (SONN-\{\texttt{PB-T50-RS}, \texttt{OBJ-BG}, \texttt{OBJ-ONLY}\}), ShapeNet55 (SN55) and ShapeNetPart (SNP) between the mean of the considered metrics for each dataset on the validation set and $\lambda_{\text{gate}}$.}
\label{tab:v3-correlation-gate-reg-metrics}
\end{table*}

%% file: tables/inference-v3-budgeted-tokens-and-ratio.tex
\begin{table*}[ht!]
\resizebox{\linewidth}{!}{
\begin{tabular}{l|c|c|cccccccc}
\toprule
Token & \paper-Gate & Avg & MN40 & OO3D & SONN-\texttt{OBJ-BG} & SONN-\texttt{OBJ-ONLY} & SONN-\texttt{PB-T50-RS} & SN55 & \multicolumn{2}{c}{SNP} \\
budget & $r$ &&&&&&&& $\text{mIoU}_C$ & $\text{mIoU}_I$ \\
\midrule
\multicolumn{11}{c}{\textit{\textbf{\paper-Gate with $\lambda_{\text{gate}}=10^{-11}$ baseline (Ours)}}} \\
- & - & 84.36 & 90.99 \footnotesize{$\pm$ 0.77} & 75.88 \footnotesize{$\pm$ 2.11} & 86.98 \footnotesize{$\pm$ 1.84} & 87.61 \footnotesize{$\pm$ 0.60} & 77.92 \footnotesize{$\pm$ 1.17} & 88.59 \footnotesize{$\pm$ 0.31} & 82.14 \footnotesize{$\pm$ 0.21} & 84.75 \footnotesize{$\pm$ 0.06} \\

\midrule
\multicolumn{11}{c}{\textit{\textbf{\paper-Gate (Ours) - Inference token-budgeted}}} \\
16 & - & 79.44 & 86.86 \footnotesize{$\pm$ 0.22} & 59.49 \footnotesize{$\pm$ 0.59} & 79.91 \footnotesize{$\pm$ 0.59} & 84.34 \footnotesize{$\pm$ 0.82} & 70.42 \footnotesize{$\pm$ 0.43} & 89.09 \footnotesize{$\pm$ 0.12} & 81.25 \footnotesize{$\pm$ 0.13} & 84.17 \footnotesize{$\pm$ 0.06} \\
32 & - & 82.09 & 91.49 \footnotesize{$\pm$ 0.37} & 76.13 \footnotesize{$\pm$ 0.69} & 86.14 \footnotesize{$\pm$ 0.57} & 87.76 \footnotesize{$\pm$ 0.56} & 79.75 \footnotesize{$\pm$ 0.39} & 68.50 \footnotesize{$\pm$ 0.19} & 82.17 \footnotesize{$\pm$ 0.11} & 84.77 \footnotesize{$\pm$ 0.05} \\
48 & - & 75.16 & 91.56 \footnotesize{$\pm$ 0.24} & 76.58 \footnotesize{$\pm$ 0.38} & 87.01 \footnotesize{$\pm$ 0.74} & 87.18 \footnotesize{$\pm$ 0.65} & 80.57 \footnotesize{$\pm$ 0.32} & 11.86 \footnotesize{$\pm$ 0.13} & 81.88 \footnotesize{$\pm$ 0.13} & 84.62 \footnotesize{$\pm$ 0.05} \\
64 & - & 68.37 & 86.63 \footnotesize{$\pm$ 0.35} & 39.30 \footnotesize{$\pm$ 0.26} & 87.47 \footnotesize{$\pm$ 0.61} & 86.09 \footnotesize{$\pm$ 0.44} & 80.50 \footnotesize{$\pm$ 0.33} & 01.15 \footnotesize{$\pm$ 0.00} & 81.44 \footnotesize{$\pm$ 0.14} & 84.41 \footnotesize{$\pm$ 0.05} \\
80 & - & - & - & - & - & - & - & - & 80.89 \footnotesize{$\pm$ 0.13} & 83.91 \footnotesize{$\pm$ 0.06} \\
96 & - & - & - & - & - & - & - & - & 80.07 \footnotesize{$\pm$ 0.17} & 83.33 \footnotesize{$\pm$ 0.05} \\
112 & - & - & - & - & - & - & - & - & 79.17 \footnotesize{$\pm$ 0.17} & 82.67 \footnotesize{$\pm$ 0.06} \\
128 & - & - & - & - & 65.61 \footnotesize{$\pm$ 0.46} & 46.33 \footnotesize{$\pm$ 0.53} & 55.99 \footnotesize{$\pm$ 0.42} & - & 78.06 \footnotesize{$\pm$ 0.11} & 82.14 \footnotesize{$\pm$ 0.05} \\

\midrule
\multicolumn{11}{c}{\textit{\textbf{\paper (Ours) - change fusion ratio $r$}}} \\
- & 0.0 & 79.54 & 86.87 \footnotesize{$\pm$ 0.38} & 59.56 \footnotesize{$\pm$ 0.72} & 80.62 \footnotesize{$\pm$ 0.60} & 84.25 \footnotesize{$\pm$ 0.78} & 70.44 \footnotesize{$\pm$ 0.41} & 89.18 \footnotesize{$\pm$ 0.13} & 81.21 \footnotesize{$\pm$ 0.14} & 84.15 \footnotesize{$\pm$ 0.05} \\
- & 0.1 & 79.54 & 86.87 \footnotesize{$\pm$ 0.38} & 59.56 \footnotesize{$\pm$ 0.72} & 80.62 \footnotesize{$\pm$ 0.60} & 84.25 \footnotesize{$\pm$ 0.78} & 70.44 \footnotesize{$\pm$ 0.41} & 89.18 \footnotesize{$\pm$ 0.13} & 81.21 \footnotesize{$\pm$ 0.14} & 84.15 \footnotesize{$\pm$ 0.05} \\
- & 0.2 & 79.54 & 86.87 \footnotesize{$\pm$ 0.38} & 59.56 \footnotesize{$\pm$ 0.72} & 80.62 \footnotesize{$\pm$ 0.60} & 84.25 \footnotesize{$\pm$ 0.78} & 70.44 \footnotesize{$\pm$ 0.41} & 89.18 \footnotesize{$\pm$ 0.13} & 81.21 \footnotesize{$\pm$ 0.14} & 84.15 \footnotesize{$\pm$ 0.05} \\
- & 0.3 & 79.53 & 86.87 \footnotesize{$\pm$ 0.38} & 59.56 \footnotesize{$\pm$ 0.74} & 80.67 \footnotesize{$\pm$ 0.67} & 84.18 \footnotesize{$\pm$ 0.77} & 70.44 \footnotesize{$\pm$ 0.41} & 89.18 \footnotesize{$\pm$ 0.13} & 81.21 \footnotesize{$\pm$ 0.14} & 84.15 \footnotesize{$\pm$ 0.05} \\
- & 0.4 & 80.18 & 87.01 \footnotesize{$\pm$ 0.34} & 64.61 \footnotesize{$\pm$ 0.59} & 80.60 \footnotesize{$\pm$ 0.70} & 83.99 \footnotesize{$\pm$ 0.65} & 70.63 \footnotesize{$\pm$ 0.34} & 89.18 \footnotesize{$\pm$ 0.13} & 81.27 \footnotesize{$\pm$ 0.13} & 84.15 \footnotesize{$\pm$ 0.05} \\
- & 0.5 & 84.86 & 91.94 \footnotesize{$\pm$ 0.24} & 76.83 \footnotesize{$\pm$ 0.59} & 86.08 \footnotesize{$\pm$ 0.94} & 86.78 \footnotesize{$\pm$ 0.51} & 81.10 \footnotesize{$\pm$ 0.31} & 89.18 \footnotesize{$\pm$ 0.13} & 82.12 \footnotesize{$\pm$ 0.13} & 84.81 \footnotesize{$\pm$ 0.05} \\
- & 0.6 & 66.04 & 89.15 \footnotesize{$\pm$ 0.27} & 64.27 \footnotesize{$\pm$ 0.71} & 76.49 \footnotesize{$\pm$ 0.53} & 73.17 \footnotesize{$\pm$ 1.14} & 63.77 \footnotesize{$\pm$ 0.35} & 01.21 \footnotesize{$\pm$ 0.05} & 78.05 \footnotesize{$\pm$ 0.09} & 82.24 \footnotesize{$\pm$ 0.06} \\
- & 0.7 & 57.20 & 86.84 \footnotesize{$\pm$ 0.21} & 39.42 \footnotesize{$\pm$ 0.44} & 66.54 \footnotesize{$\pm$ 0.64} & 46.88 \footnotesize{$\pm$ 0.38} & 56.51 \footnotesize{$\pm$ 0.30} & 01.18 \footnotesize{$\pm$ 0.01} & 78.05 \footnotesize{$\pm$ 0.09} & 82.17 \footnotesize{$\pm$ 0.06} \\
- & 0.8 & 56.91 & 86.79 \footnotesize{$\pm$ 0.24} & 39.28 \footnotesize{$\pm$ 0.43} & 65.35 \footnotesize{$\pm$ 0.61} & 46.42 \footnotesize{$\pm$ 0.40} & 56.10 \footnotesize{$\pm$ 0.37} & 01.15 \footnotesize{$\pm$ 0.00} & 78.03 \footnotesize{$\pm$ 0.09} & 82.16 \footnotesize{$\pm$ 0.06} \\
- & 0.9 & 56.89 & 86.64 \footnotesize{$\pm$ 0.22} & 39.28 \footnotesize{$\pm$ 0.43} & 65.35 \footnotesize{$\pm$ 0.61} & 46.42 \footnotesize{$\pm$ 0.40} & 56.07 \footnotesize{$\pm$ 0.37} & 01.15 \footnotesize{$\pm$ 0.00} & 78.03 \footnotesize{$\pm$ 0.09} & 82.16 \footnotesize{$\pm$ 0.06} \\
- & 1.0 & 56.88 & 86.64 \footnotesize{$\pm$ 0.22} & 39.28 \footnotesize{$\pm$ 0.43} & 65.35 \footnotesize{$\pm$ 0.61} & 46.42 \footnotesize{$\pm$ 0.40} & 56.01 \footnotesize{$\pm$ 0.37} & 01.15 \footnotesize{$\pm$ 0.00} & 78.03 \footnotesize{$\pm$ 0.09} & 82.16 \footnotesize{$\pm$ 0.06} \\

\bottomrule
\end{tabular}
}
\caption{\textbf{Changing ratio at inference-time and token-budgeted inference for \paper-Gate.} We use the best backbone on each dataset with $\lambda_{\text{gate}}=10^{-11}$ and without weight decay. The token-budget represents the maximum number of tokens we accept as input in the core encoder. $r$ represents the fusion ratio threshold.}
\label{tab:inference-v3-token-budget-ratio}
\end{table*}

%% file: tables/computational-cost-empiric.tex
\begin{table}[ht!]
\resizebox{\linewidth}{!}{
\begin{tabular}{l|cc|cccccc}
\toprule
Method & \# supertokens & Token budget & Params (M) & MACs (G) & FLOPs (G) & GPU Mem (MiB) & Avg runtime (s) & Throughput (obj/s) \\
\midrule
baseline & - & - & 21.84 & 238.624 & 478.128 & 1143.0 & 0.09/0.93 & 747.11/68.71 \\
ToMe \cite{DBLP:conf/iclr/BolyaFDZFH23} ($r_{\text{tome}}=16$) & - & - & 21.84 & 115.412 & 231.319 & 1143.0 & 0.06/0.63 & 1091.42/100.96 \\
ToMe \cite{DBLP:conf/iclr/BolyaFDZFH23} ($r_{\text{tome}}=8$) & - & - & 21.84 & 97.8966 & 196.233 & 1143.0 & 0.05/0.60 & 1174.07/107.56 \\
ToMe \cite{DBLP:conf/iclr/BolyaFDZFH23} ($r_{\text{tome}}=4$) & - & - & 21.84 & 90.0449 & 180.505 & 1143.0 & 0.05/0.58 & 1213.10/109.90 \\
ToMe \cite{DBLP:conf/iclr/BolyaFDZFH23} ($r_{\text{tome}}=2$) & - & - & 21.84 & 86.572 & 173.548 & 1143.0 & 0.05/0.58 & 1219.81/110.11 \\
ToMe \cite{DBLP:conf/iclr/BolyaFDZFH23} ($r_{\text{tome}}=1$) & - & - & 21.84 & 85.0621 & 170.523 & 1143.0 & 0.05/0.58 & 1217.40/109.93 \\

\midrule
\multicolumn{9}{c}{\textit{\textbf{\paper (Ours)}}} \\
\paper & 16 & - & 22.73 & 88.9903 & 178.394 & 1135.5 & 0.06/0.61 & 1160.97/104.32 \\
\paper (ViT-T) & 16 & - & 6.33 & 71.6618 & 143.698 & 1073.0 & 0.05/0.58 & 1335.26/110.52 \\
\paper & 8 & - & 22.73 & 78.042 & 156.464 & 1135.5 & 0.05/0.58 & 1271.58/110.28 \\
\paper & 4 & - & 22.73 & 72.5678 & 145.498 & 1135.5 & 0.05/0.59 & 1309.76/109.08 \\
\paper & 2 & - & 22.73 & 69.8307 & 140.016 & 1135.5 & 0.05/0.56 & 1349.58/113.30 \\
\paper & 1 & - & 22.73 & 68.4622 & 137.274 & 1135.5 & 0.05/0.56 & 1365.92/115.18 \\
\paper-Gate & 16 & - & 22.78 & 157.593 & 315.812 & 1135.7 & 0.11/0.97 & 591.60/65.88 \\

\midrule
\multicolumn{9}{c}{\textit{\textbf{\paper (Ours) - Inference token-budgeted}}} \\
\paper-Gate & 16 & 16 & 22.78 & 87.8332/89.1921 & 176.076/178.798 & 1135.7 & 0.07/0.64 & 864.02/100.57 \\
\paper-Gate & 16 & 32 & 22.78 & 107.651/107.651 & 215.774/215.774 & 1135.7 & 0.08/0.72 & 756.92/88.91 \\
\paper-Gate & 16 & 48 & 22.78 & 126.073/127.432 & 252.674/255.396 & 1135.7 & 0.09/0.82 & 679.59/78.27 \\

\midrule
\multicolumn{9}{c}{\textit{\textbf{\paper (Ours) - change fusion ratio $r$}}} \\
\paper-Gate (r=0.1) & 16 & - & 22.78 & 87.8332/168.012 & 176.076/336.684 & 1135.7 & 0.07/0.66 & 881.45/97.50 \\
\paper-Gate (r=0.2) & 16 & - & 22.78 & 89.1921/163.293 & 178.798/327.23 & 1135.7 & 0.07/0.66 & 875.99/97.52 \\
\paper-Gate (r=0.3) & 16 & - & 22.78 & 159.254/159.141 & 319.139/318.912 & 1135.7 & 0.09/0.72 & 751.83/88.35 \\
\paper-Gate (r=0.4) & 16 & - & 22.78 & 161.028/161.783 & 322.694/324.205 & 1135.7 & 0.11/0.96 & 588.67/66.36 \\
\paper-Gate (r=0.5) & 16 & - & 22.78 & 157.593/156.196 & 315.812/313.014 & 1135.7 & 0.11/0.96 & 593.70/66.77 \\
\paper-Gate (r=0.6) & 16 & - & 22.78 & 152.006/152.006 & 304.621/304.621 & 1135.7 & 0.10/0.94 & 623.29/67.88 \\
\paper-Gate (r=0.7) & 16 & - & 22.78 & 152.006/152.006 & 304.621/304.621 & 1135.7 & 0.10/0.96 & 623.66/66.88 \\
\paper-Gate (r=0.8) & 16 & - & 22.78 & 152.006/153.403 & 304.621/307.419 & 1135.7 & 0.10/0.96 & 620.73/66.51 \\
\paper-Gate (r=0.9) & 16 & - & 22.78 & 152.006/152.006 & 304.621/304.621 & 1135.7 & 0.10/0.95 & 623.58/67.11 \\

\bottomrule
\end{tabular}
}
\caption{\textbf{Empirical computational cost at inference-time.} We use random input processed only by the pre-trained considered method. The input contains 64 point clouds of $2^{10}$ 3D points. We group them into 64 tokens of 32 points each. Runtime and throughput are averaged over 40 runs after 10 warmup runs on an NVIDIA RTX A3000 6GB Laptop GPU/Intel Core i9-11950H @ 2.60GHz (tokenization and encoding are included). All methods take N tokens and output N tokens (regardless of the specific task). ToMe \cite{DBLP:conf/iclr/BolyaFDZFH23} halves the number of tokens at each layer up to the specified $r_{\text{tome}}$. We use unfrozen backbones for \paper. For \paper-Gate, we use the model with $\lambda_{\text{gate}}=10^{-11}$ without weight decay. MACs and FLOPs are for the forward pass only and were computed with the \texttt{calflops} library. The Token-budget represents the maximum number of tokens we accept as input in the core encoder.}
\label{tab:empirical-computational-cost-complete}
\end{table}